\documentclass[double]{article}
\usepackage{times}
\usepackage{algorithm}
\usepackage{algorithmic}
\usepackage{wrapfig}
\usepackage{verbatim}
\usepackage{amsmath}
\usepackage{graphicx}
\usepackage{subcaption}
\usepackage[active]{srcltx}
\usepackage{booktabs}
\usepackage{color}
\usepackage{multirow}
\usepackage{makecell}
\usepackage{lineno}
\usepackage{dirtytalk}
\usepackage{amsthm}
\usepackage{authblk}
\usepackage{listings}
\usepackage{tikz}
\usepackage{longtable}
\usetikzlibrary{shapes.geometric, arrows}

\usepackage{epsfig,graphicx,amsfonts}
\usepackage{amsmath,amsthm,amssymb}
\usepackage{xcolor}

\usepackage{adjustbox}
\usepackage{multirow}
\usepackage{setspace}
\usepackage{hyperref}
\usepackage{comment}

\long\def\comment #1\commentend{}

\begin{document}

\title{\Large Disturbance-Aware Aerial Robotics for Ethical Wildlife Monitoring}

\author{Mahmut Osmanovic$^{1,x}$, Isac Paulsson$^{1,x}$, Teddy Lazebnik$^{1,2,*}$\\
\(^1\) Department of Computing, Jonkoping University, Jonkoping, Sweden \\
\(^2\) Department of Information Systems, University of Haifa, Haifa, Israel \\
\(^x\) These authors contributed equally. \\
\(^*\) Corresponding author: teddy.lazebnik@ju.se 
}

\date{ }

\maketitle 

\begin{abstract}
\noindent
Reliable wildlife monitoring is essential for ecology and conservation, yet many existing methods, such as tagging, capture, and close-range observation, can alter the very behaviors they aim to measure. Aerial robots offer a scalable alternative, which has shown promising performance in multiple studies. Nonetheless, existing approaches typically lack behavioral awareness, rely on fixed heuristics, or require real-world training data that are costly, impractical, and ethically difficult to obtain. As a result, there remains no general framework for adaptive drone-based monitoring that can both preserve ecological validity and scale across species, behaviors, and robotic platforms. In this study, we introduce a disturbance-aware reinforcement-learning-based framework for heterogeneous aerial robotic fleets that enables autonomous wildlife tracking while explicitly minimizing behavioral disruption. We couple a zoologically grounded simulation environment with fitted animal movement models derived from real trajectory statistics, and train control policies using a reward formulation that captures the trade-off between observation quality and disturbance risk. Across three species (pigeon, jackal, and spur-winged lapwing) with distinct ecologies and motion patterns and four increasingly strategic behavior models common in nature, the learned policies consistently surpassed currently used rule-based baselines and generalized across monitoring tasks, animal dynamics, and drone types. These results establish disturbance-aware learning as a viable foundation for non-invasive autonomous wildlife observation, opening a path towards scalable, ethically responsible, and scientifically reliable robotic monitoring in ecology and conservation.  \\

\end{abstract}

\maketitle \thispagestyle{empty}
\pagestyle{myheadings} \markboth{Draft:  \today}{Draft:  \today}
\setcounter{page}{1}

\section{Introduction}
\label{sec:introduction}
Wildlife monitoring is central to ecology and conservation since robust estimates of animal presence, movement, and behaviour are needed to understand population dynamics and guide management decisions \cite{allen2016linking,wall2014novel,elzinga2001monitoring}. In practice, monitoring still relies heavily on direct observation, camera traps, and animal-borne telemetry, which remain among the most widely used approaches in field ecology \cite{Marion2020,LahozMonfort2021,lazebnik2026empirically}. These tools have generated major advances, but each comes with important drawbacks. Direct observation and manual tracking are labour-intensive and difficult to sustain across large areas, long time horizons, or remote habitats \cite{LahozMonfort2021,gahm2026selection}. Camera traps provide a less invasive alternative, yet their outputs are shaped by imperfect detection, camera placement bias, and even the possibility that the devices themselves influence behaviour through light, odour, or mechanical noise \cite{Burton2015,Kolowski2017,Caravaggi2020}. Similarly, biologging technologies such as GPS collars can reveal fine-scale movement patterns, but the required capture, handling, and tagging procedures can themselves induce short-term behavioural changes and thereby bias subsequent observations \cite{Stiegler2024,handcock2009monitoring,horback2012effects}. As a result, current wildlife-monitoring pipelines often involve a persistent trade-off between coverage, fidelity, and invasiveness, motivating the development of more adaptive and less disruptive sensing frameworks. 

Aerial robotic platforms are an attractive solution for wildlife monitoring since they combine rapid deployment, broad field of view, and fine spatiotemporal sensing with the ability to operate over terrains that are difficult, dangerous, or costly to survey from the ground or with manned aircraft \cite{Aliane2025,Pedrazzi2025}. Recent studies illustrate this promise across diverse ecological settings \cite{cliff2018robotic,chabot2015wildlife,preethichandra2024review}. For instance, Hvala et al. integrated fixed-wing drone surveys with GPS-collared feral pigs and showed that detection probability varies strongly with survey timing and local environmental conditions, enabling more accurate correction of abundance estimates from aerial imagery \cite{Hvala2023}. In a similar manner, Carey-Douglas et al. used a quadcopter to observe African savannah elephants and found evidence of both short-term and repeated-exposure habituation when flights followed disturbance-minimizing protocols, supporting drones as a minimally invasive tool for behavioural observation \cite{CareyDouglas2025}. Alike, Thapa et al. deployed a VTOL fixed-wing drone along a 73-km river stretch in Chitwan National Park and produced high-resolution estimates of abundance, occupancy, and co-occurrence for gharials and muggers at substantially lower cost than traditional ground surveys \cite{Thapa2025}. 

Nevertheless, while current aerial wildlife-monitoring solutions have established the utility of drones, they remain limited in the level of autonomy and behavioural awareness they provide \cite{shekhawat2025challenges,Aliane2025}. Much of the literature still frames disturbance mitigation as a matter of selecting conservative flight parameters (such as altitude, speed, and approach angle) even though these thresholds vary substantially across species, habitats, and observational contexts \cite{Afridi2025,Pedrazzi2025}. In parallel, much of the recent Artificial Intelligence (AI) work has focused on animal detection, localisation, and behaviour extraction from drone imagery, rather than on closing the loop between perception and control \cite{Axford2024}. Existing autonomous systems only partially address this gap. For instance, Li et al.\ proposed a motion-camouflage navigation strategy to reduce visual disturbance during animal observation, but the method relies on a handcrafted guidance law rather than learned adaptation to heterogeneous behavioural responses \cite{Li2022}. A different approach is WildWing, which improves repeatability by autonomously maintaining useful viewing geometry for behavioural video collection, yet its objective is standardised footage acquisition rather than explicit optimisation of the trade-off between monitoring performance and animal disturbance \cite{Kline2025}. Similarly, the recent WildDrone programme highlights the promise of autonomy and multi-drone coordination, while also underscoring that wildlife monitoring in practice still depends heavily on manually flown missions, limited field-tested autonomy, and protocol-driven mitigation of disturbance \cite{Lundquist2025}. 

In parallel, Reinforcement Learning (RL) has increasingly been used to solve robotics problems that require continuous adaptation to moving targets, partial observability, and competing operational constraints \cite{zhu2025deep,nadour2025adaptive,singh2022reinforcement}. In the aerial domain, Xu et al.\ showed that deep RL (DRL) can learn continuous-control policies for joint obstacle avoidance and target tracking, allowing an Unmanned Aerial Vehicle (UAV) to maintain pursuit while reacting online to dynamic surroundings \cite{Xu2022}. Extending this idea to coordinated teams, Kong et al.\ formulated multi-UAV target assignment and path planning as a partially observable Markov decision process (POMDP) and used DRL to learn collision-free policies that jointly allocate targets and generate paths in dynamic 3D environments \cite{Kong2024}. Beyond aerial pursuit, RL has also been applied to socially aware robot navigation, where the agent must complete its task without unduly disturbing nearby humans \cite{kabir2026socially,guillen2023evolution}. For example, Cheng et al.\ proposed a multi-objective DRL framework that explicitly balances goal reaching, safety, collision avoidance, and path smoothness in crowded environments \cite{Cheng2024}, while Xue et al.\ designed a socially compliant RL navigation policy that incorporates pedestrian dynamics and discomfort-aware rewards to improve both safety and comfort \cite{Xue2024}. 

Despite these advances, there is no general framework for wildlife monitoring that jointly models animal behaviour, learns adaptive drone-control policies, and explicitly optimizes the trade-off between observation quality and behavioural disturbance. To this end, in this study, we propose a stealth-aware DRL-based framework for heterogeneous aerial robotic fleets that enables autonomous wildlife monitoring while explicitly accounting for the behavioural sensitivity of the observed animals. Our approach combines three components. First, we construct a zoologically grounded simulation environment that captures animal movement, perception, and disturbance responses, allowing the monitoring problem to be formulated without relying on extensive real-world drone--animal interaction data. Second, we fit species-specific movement models to empirical trajectories so that training occurs in environments that reflect realistic behavioural patterns rather than abstract motion assumptions. Third, we train DRL controllers with a reward formulation that explicitly balances monitoring quality against disturbance, encouraging the drone to maintain informative views of the target while avoiding aggressive pursuit, intrusive proximity, or trajectories likely to alter animal behaviour. In contrast to existing drone-monitoring systems that rely on fixed protocols, handcrafted control rules, or perception-only pipelines, the proposed framework learns adaptive closed-loop policies that generalize across species, behavioural regimes, and drone types. In this way, our study bridges wildlife monitoring, animal behaviour modelling, and autonomous robotic control, and provides a practical route towards scalable and non-invasive data collection in ecology and conservation.

The remainder of this paper is organized as follows. Section~\ref{sec:rw} reviews related work on wildlife monitoring, aerial robotic observation, and learning-based control for adaptive tracking. Section~\ref{sec:model} presents the proposed framework, including the zoologically grounded simulation environment, the animal behaviour and disturbance models, and the RL formulation. Section~\ref{sec:exp} describes the experimental design, datasets, evaluated species, and baseline methods. Section~\ref{sec:results} reports the empirical results across behavioural regimes, species, and robot morphologies. Finally, Section~\ref{sec:discussion} discusses the implications of the findings, limitations of the current approach, and directions for future research.

\section{Related Work}
\label{sec:rw}
In this section, we first examine wildlife monitoring and its relation to data bias and animal welfare. Then, we review the application of drones to wildlife monitoring, behavioral responses to disturbance from drones, movement ecology and behavior modeling, and finally, we quickly cover RL for adaptive control.

\subsection{Wildlife monitoring}
Wildlife monitoring provides an important empirical basis for studying animal behavior, spatial use, habitat selection, and population dynamics over time \cite{Yoccoz2001, Kays2015}. However, the conditions of monitoring change based on the monitoring method, which risks the introduction of bias to the data. Direct observation remains valuable for contextual interpretation of behavior, but it is labor intensive, difficult to scale, and susceptible to observer effects \cite{beale2004human}. Camera traps provide an alternative method for studying wildlife occurrence, behavior, and activity patterns, yet they are restricted to fixed viewpoints and, as such, depend heavily on placement \cite{Burton2015}. Bio-logging and telemetry can provide fine-grained data on movement and behavior, but the attachment of devices may itself influence animal behavior, energy expenditure, reproduction, survival, or welfare \cite{barron2010meta,Bodey2018}. These considerations make disturbance a central issue in monitoring methodology, beyond the ethical concerns. Monitoring systems do not passively record behavior, they may also alter the behavior being measured. Consequently, the quality of a monitoring method must be considered not only in terms of data quantity or resolution, but also in relation to how faithfully the recorded behavior reflects the animal's natural state \cite{beale2004human,barron2010meta,Bodey2018}. This perspective is particularly important for mobile sensing platforms, where the position and motion of the observer can change continuously during monitoring.

\subsection{Aerial robots for wildlife monitoring}
Aerial robots as a tool for wildlife monitoring are increasingly prevalent \cite{Schad2023}. They provide mobile platforms that can access remote or hazardous environments to capture video or images. Recent reviews suggest a diverse range of applications, including population estimation, habitat mapping, morphometric assessment, behavioral observation, anti-poaching support, and tracking of individual and group movement \cite{christie2016unmanned,Schad2023,Afridi2025,Aliane2025}. In contrast to fixed ground-based sensors, drones can reposition relative to animals and terrain. This mobility is valuable when the aim is to monitor dynamic behavior or relate behavior to spatial context, such as habitat structure, spatial organization, and group movement \cite{Schad2023}. At the same time, drones provide a flexible platform for monitoring through different drone types, sensing configurations, and operating procedures. Fixed-wing and multi-rotor systems support different trade-offs in ground coverage, maneuverability, hovering capability, flight duration, and efficiency \cite{Schad2023,Vas2015,Duporge2021,BrissonCuradeau2025}. The usefulness of any platform depends on sensing geometry, environmental conditions, and the reliability of downstream data processing \cite{Schad2023}. Furthermore, in this context, automated image analysis, detection, and tracking pipelines extend the value of aerial data collection beyond manual review and increase the feasibility of high-throughput monitoring \cite{Schad2023,Aliane2025}. Closely related to the disturbance in our problem is the field of human-aware robot navigation. Here robots must perform tasks while considering the comfort, safety, and personal space of nearby humans \cite{kruse2013human}.
    
    Recently, progress has been made towards increasing the autonomy of aerial robotic platforms for wildlife monitoring. Several methods using drones have been developed to track animals using VHF tags. For example, Nguyen et al. \cite{Nguyen2019} demonstrated a system using RSSI for localization of multiple animals with drone-based radio-telemetry. This work was extended by Chen et al. \cite{chen2023}, adding considerations for complex terrain. A method of minimizing visual disturbance has also been introduced with the goal of tracking multiple animals while reducing perceived optical flow, the method is inspired by stalking techniques found in some predators \cite{Li2022}. Additionally, motivated by the observation that manual coordination of multiple UAVs does not scale, recent work has proposed a method for decentralized multi-drone coordination for wildlife video acquisition \cite{Grushchak2024}. These systems show the value of introducing robotics to wildlife monitoring and demonstrate that aerial platforms can support increasingly autonomous sensing, tracking, and coordination in the field. However, the same mobility that makes aerial robots useful also creates an issue from a disturbance perspective. Reviews of drone-wildlife interactions show that responses depend on operational factors such as altitude, speed, and proximity, as well as on drone noise, flight patterns, and visual profile \cite{Afridi2025,Schad2023}. Lower flight altitudes and shorter distances are among the most consistent predictors of disturbance across studies. Recent work has proposed species-informed methods for selecting flight altitudes that preserve observation quality and reduce acoustic impact  \cite{Duporge2021,Schad2023}. In summary, existing work has addressed drone-based wildlife monitoring, autonomous tracking, and disturbance assessment, but relatively few studies have examined autonomous wildlife tracking while explicitly accounting for disturbance during the tracking process.
        

    Disturbance is commonly understood as the disruption of normal behavior in response to external stimuli, often interpreted as a form of perceived predation risk \cite{frid2002}. Behavioral responses to disturbance are typically dynamic and non-binary, generally varying with context, species, and perceived threat \cite{Lima1999, mulero2017, Schad2023}. Depending on the disturbance, animals may show anything from weaker responses like vigilance and avoidance, to stronger responses like immediate flight \cite{Vas2015,bennitt2019,BrissonCuradeau2025}. In the case of drones, responses have been shown to depend strongly on characteristics of the aircraft, positioning, and trajectory relative to the animal. Across species, reactions vary with flight pattern, engine type, aircraft size, altitude, speed, and approach distance \cite{Afridi2025}. Additionally, target-oriented flights tend to provoke stronger responses than more regular survey-style patterns \cite{mulero2017}. Oblique approaches appear to result in smaller responses, likely due to being perceived as less threatening \cite{Vas2015, frid2002}. These findings suggest that commonly used operational guidelines, such as conservative stand-off distances or minimum altitudes, should be understood as practical heuristics rather than universally valid thresholds.
    
    Approach geometry appears to be particularly important. Birds were found to responded more strongly to vertical than horizontal approaches, suggesting that the geometric character of the approach can alter perceived threat \cite{Vas2015}. This is consistent with the broader finding that more direct target-oriented flights tend to increase disturbance \cite{mulero2017}. Distance and speed have also emerged as important determinants of behavioral response. In a recent meta-analysis of avian drone studies, flushing responses were strongly influenced by drone distance and speed, with closer and faster approaches increasing the likelihood of adverse reactions \cite{BrissonCuradeau2025}. More recent reviews likewise identify altitude, speed, and proximity as principal operational factors shaping wildlife responses to drones \cite{Afridi2025}. Collectively, these studies suggest that disturbance is best understood as a continuous outcome of relative position and motion rather than as a fixed threshold condition. For modeling purposes, the most relevant implication is that drone disturbance can be represented by vertical and horizontal distance as well as threatening flight geometry. Furthermore, as monitoring is scaled to multi-agent systems, the accumulation of disturbance from multiple concurrent sources must be considered. Due to constraints in physiology and the need for balance between anti-predator behavior and essential activities, animal responses to multiple threats tend to be saturating \cite{Sih1998, Lima1999}. The approach of a drone toward an animal gives rise to a behavioral interaction. Consequently, disturbance-aware monitoring must account for the effects of drone motion and positioning on animal behavior over time.


    Central to the study of animal behaviors is movement ecology, the study of how, why, when and where animals (or organisms) move\cite{nathan2008movement, Fagan2014}. By understanding animal movements, we can analyze and better understand, among other things, behaviors, predator-prey interactions and population dynamics \cite{Fagan2014}. To aid in this endeavor, movement models have been introduced. A prime example is the correlated random walk (CRW), which describes a random but directionally correlated movement \cite{Kareiva1983,Codling2008,Fagan2014}. Due to its relative simplicity and interpretability (modeling movement as step lengths and turning angles), CRWs can serve as a baseline against which more structured movement models can be compared \cite{Kareiva1983, Codling2008}. Area-restricted search (ARS) is a more structured movement model relative to CRW. It introduces information from the environment to the decision-making process using triggers in the form of resource encounters, sensory cues or local expectations. Upon encountering a trigger in ARS, the animal switches from a global exploratory state to a locally exploitative one. The exploitative state is characterized by slower and more tortuous movement \cite{Dorfman2022}. This leads to a more thorough search in the area where a resource was encountered, which can be more profitable if resources are clustered. Compared to CRW, the resulting movement is dynamic and shaped by local opportunity. Additional structure can be introduced by giving spatial awareness to a movement model. By giving the model access to high-value locations or value maps in heterogeneous environments, movement can also be governed by the relative attractiveness and accessibility of regions in the environment \cite{Matthiopoulos2003}. This is important to reflect how resources and environmental variation shape movement. In simulation, monitoring strategies may differ in performance based on the type of movement the model produces. The utilized movement models serve a dual purpose. First, they provide training data for reinforcement learning. Second, they evaluate and fit empirical movement trajectories. 

\subsection{Reinforcement learning for adaptive tracking and navigation}
RL addresses sequential decision-making problems where an agent selects an action based on the current state and receives feedback through rewards. The agent is able to learn and improve its behavior through interaction with the environment \cite{Sutton2018}. This is well suited to tasks where actions have delayed consequences and in cases where multiple, possibly conflicting objectives need to be considered over time \cite{Felten2024, Tang2025}. In the context of disturbance-aware monitoring, the central challenge is maintaining informative and stable monitoring while limiting the induced animal disturbance. Since the objectives of monitoring quality and disturbance minimization are inherently conflicting, the solution must balance these criteria within a unified control formulation \cite{roijers2013survey}. Problems in RL are generally formulated as Markov Decision Processes (MDPs), which define state, actions, transition dynamics, and rewards for sequential decision-making \cite{Sutton2018}. When an agent's observation does not contain the full state, the problem is commonly formulated as a POMDP \cite{lauri_2023, Kaelbling1998}. This is relevant when the agent's observations are limited by, for instance, field of view (FoV) or occlusion. In these problems, RL learns a policy mapping states or observations to actions. Considering that an animal may be disturbed or stressed due to erroneous actions, precautions must be taken before deployment to ensure safe operation \cite{mulero2017,Schad2023}. This is especially true in the case of RL, since policies are learned through interaction with the environment. During learning, the agent must explore the environment to estimate which actions lead to favorable or unfavorable outcomes in different states \cite{Garcia2015}. Under favorable conditions, RL policies can be trained in simulation and transferred to more realistic settings. The process relies on generalization, and if there is a mismatch between the two environments, transfer may not be feasible \cite{Zhao2020}.

    Reward design is a central part of achieving a desired solution as the reward directly governs the learned behavior. Shaping terms can be introduced to encourage intermediate behaviors and improve learning efficiency. However, considerations must be made to ensure that shaping terms do not distort the intended task and learning goals \cite{ng1999policy}. For monitoring and tracking, a common approach is combining a primary objective related to information gain, localization accuracy, target visibility or observation quality with auxiliary terms and constraints concerning resource efficiency, maneuverability, robustness, safety or low-disturbance interaction \cite{Nguyen2019, chen2023, Li2022}. RL is widely used in the application of tracking with both passive and active camera control for single and multiple targets. Prominent approaches include deep q-network (DQN), proximal policy optimization (PPO), soft actor-critic (SAC), and various other methods \cite{Barrientos2024}. Regarding drones, RL-based navigation and tracking have found success, with methods considering localization constraints, obstacles and unstructured outdoor navigation \cite{Cetin2019, Zhang2025, zhu2025deep}. However, to our knowledge, RL remains relatively unexplored in the application of wildlife monitoring, especially when considering disturbance.

\section{Model Definition}
    \label{sec:model}
Let us consider disturbance-aware wildlife monitoring with a fleet of $N_d$ aerial robots observing $N_a$ animals over a finite horizon $T<\infty$. At each discrete time step $t$, drone $d$ receives a local observation, selects a high-level control action, and attempts to maintain visual contact with its assigned target animal while minimizing behavioral disturbance. In this context, let $\mathbf{x}_{a,t}\in\mathbb{R}^{3}$ and $\mathbf{x}_{d,t}\in\mathbb{R}^{3}$ denote the positions of animal $a$ and drone $d$, respectively. Each drone is assigned a target using the round-robin rule $a_d^\star = d \bmod N_a.$ The policy observes only local camera-frame information rather than the full simulator state. The task is therefore formulated as a POMDP. The learning objective is $J(\pi)=\mathbb{E}_{\pi}\left[\sum_{t=0}^{T-1} r_t \right]$, where \(r_t\) rewards monitoring quality and penalizes disturbance and unnecessary motion. The key trade-off is that closer flight generally improves visibility and alignment, but also increases the likelihood of altering the animal's behavior.

In order to solve this optimizaiton problem, we proposed a framework that consists of six coupled components: a spatial environment, animal behavior models, drone sensing and motion models, a disturbance model, a reward function, and a reinforcement-learning controller. At each step, drones first update their positions and camera headings according to the selected actions. The simulator then evaluates disturbance, updates animal behavior and motion, generates new observations, computes reward, and checks termination conditions. This produces a closed-loop monitoring task in which drone motion affects disturbance, disturbance affects animal response, and animal response changes future observations. The same simulator is used for synthetic policy training and empirical trajectory replay. We desribe below each component seperatly and than the intergration between them.

\subsection{Data}
    The study uses two categories of animal movement data. Synthetic data provide controllable trajectories for training and evaluation in simulation, while empirical GPS data provide recorded animal movement patterns for fitting movement priors and evaluating transfer to real trajectories. The synthetic trajectories are generated online within the simulation environment by stochastic behavioral models. The trajectories are produced by four behavioral models: CRW, Explore-Exploit (EE), Point of Interest (POI), and Learned Point of Interest (LPOI). These models define how animal position, velocity, and behavioral state evolve over time. They generate trajectories at the simulator time step of 0.1 seconds and are used during RL training to provide continuous interaction between drones and animals. The behavioral models interact with a simulated environment that provides spatial structure and resource signals, as described in subsequent sections.
    
    In addition to synthetic trajectories, this study makes use of empirical GPS animal tracking data freely available online from \cite{lazebnik2026empirically,lapwing1,lazebnik2025individual}. The data consist of raw positional recordings (i.e., time-stamped latitude and longitude coordinates) from three animal groups: Pigeons, Spur-winged lapwings, and Jackals. Table \ref{tab:data_summary} details statistics from the raw data, Pigeons was the most regular dataset in terms of time gap frequency with no segments containing gaps over 20 seconds. The data for Jackals and Spur-winged lapwings had a considerable amount of gaps, 233,650 and 1,388,404 respectively.

    \begin{table}[!ht]
    \centering
    \setlength{\tabcolsep}{5pt}
    \renewcommand{\arraystretch}{1.15}
    \caption{Summary of the movement datasets.}
    \label{tab:data_summary}
    \begin{tabular}{p{0.42\linewidth}ccc}
    \hline \hline
    \textbf{Property} & \textbf{Jackals} & \textbf{Pigeons} & \textbf{Spur-winged lapwings} \\
    \hline \hline
    Number of animals or files & 50 & 17 & 16 \\
    Mean tracking time per animal (hours) & 2,140.9 & 9.8 & 6,364.5 \\
    Mean number of samples per animal & 139,377 & 7,073 & 1,343,750 \\
    Total number of samples & 6,968,844 & 120,242 & 21,499,999 \\
    \hline \hline
    \end{tabular}
    \end{table}

    \subsection{Environment representation}
    The environment is a continuous three-dimensional coordinate system. Animals move primarily in the horizontal plane, while drones move in three dimensions subject to speed and altitude constraints. The simulation uses a fixed time step $\Delta t=0.1$ s and a maximum episode length of 2048 steps. Spatial resource heterogeneity is represented by a procedurally generated OpenSimplex-noise probability (i.e., an n-dimensional gradient noise function) field \cite{OpenSimplex_Paper}. Resource encounters are sampled stochastically from the probability field at the animals current position. Consequently, when an animal is in a more resource-rich region, encounters are more likely. To extract points of interests (POIs) from the resource map, the probability field is sampled on a regular grid and local maxima are extracted using a max-filter kernel. To filter out detected peaks in low-resource regions exceeding the kernel size, POIs are required to have a minimum probability. The POI extraction is parametrized by sampling resolution, kernel size and a minimum POI probability.

\begin{figure}[!ht]
        \centering
        \includegraphics[width=0.80\linewidth]{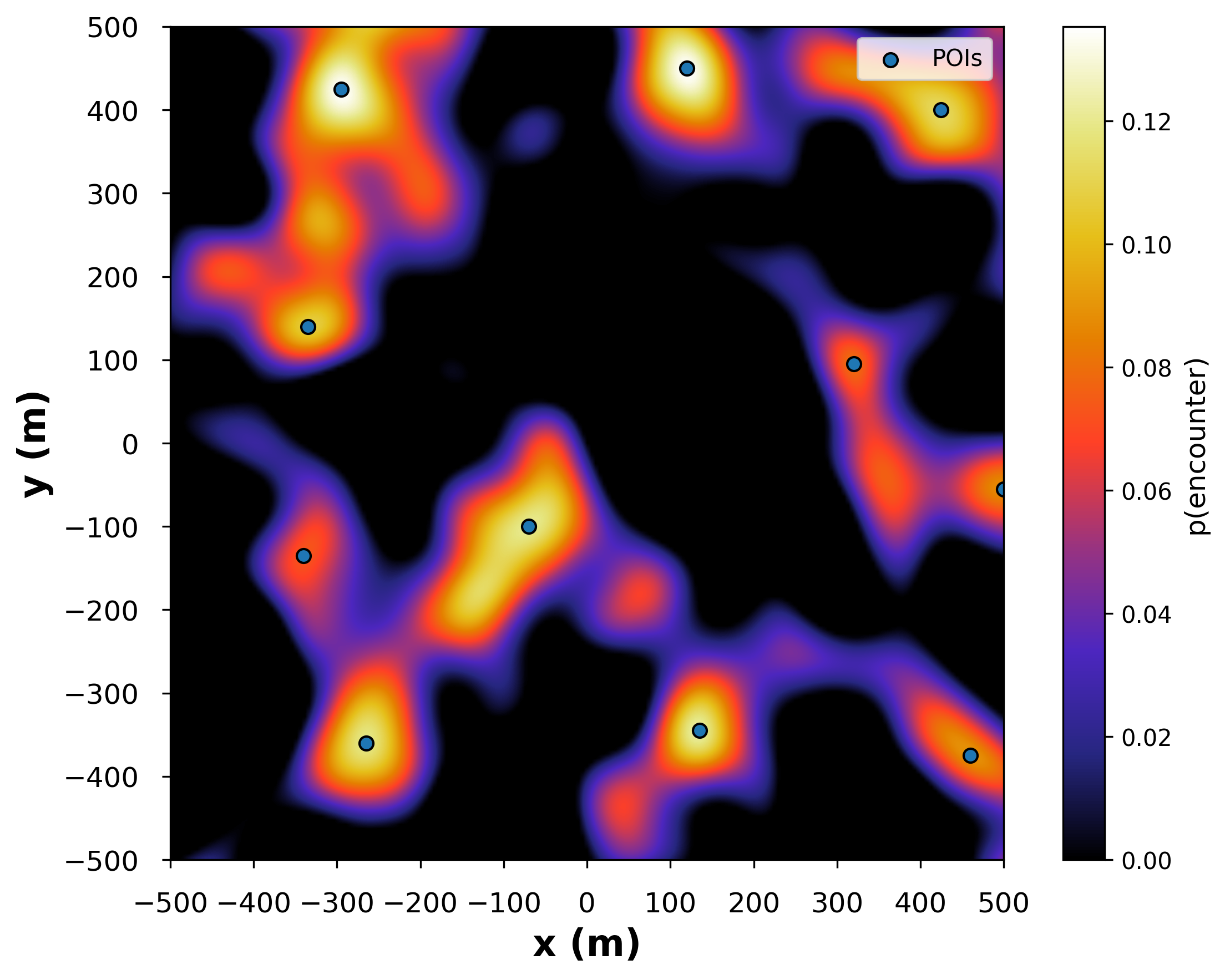}
        \caption{Example procedurally generated resource map. The heatmap shows spatially varying encounter probability, while extracted points of interest mark candidate resource locations.}
        \label{fig:resource_map}
    \end{figure}

    This representation provides a simple but flexible way to generate heterogeneous environments with continuous spatial structure, while keeping the encounter process stochastic and the POIs tied to the same resource landscape. Additionally, the representation is generally interpretable, wavelength typically corresponds to the distance between resource peaks in meters, sampling resolution is defined in meters and the assigning kernel size is defined in meters, later converted using the sampling resolution. Figure \ref{fig:resource_map} presents an example environment generated from such process.
    
    \subsection{Animal behavior model} 
    \label{subsec:behaviour_modeling}
    We introduce four behavioral models of motion which to generate synthetic animal movement data for training. In order of structural and spatial complexity they are: CRW, EE, POI and LPOI. 
    
    CRW is the most simple movement model implemented, it also serves as a movement kernel for the more structured behaviors. The kernel was implemented with support for an additional bias used in POI and LPOI. Within the kernel, speed is updated using a mean-reverting AR(1) process towards a target speed. With \(\mathbf{v}_t\) denoting the current velocity direction, the direction update is defined as $\mathbf{v}_{t+1} = \operatorname{unit}\left(\lambda \mathbf{v}_t + \boldsymbol{\epsilon}_t + \gamma \operatorname{unit}(\mathbf{b}_t)\right)$, $\qquad \boldsymbol{\epsilon}_t \sim \mathcal{N}(\mathbf{0}, \sigma_\theta^2 \mathbf{I}).$ Here, $\lambda$ controls directional persistence, $\boldsymbol{\epsilon}_t$ represents turning noise, $\mathbf{b}_t$ is the bias direction, and $\gamma$ controls the strength of the directional bias. While CRW consists of a single kernel configuration, the more structured behaviors all include two, one for the explore state, and one for the exploit state. EE is an implementation of ARS \cite{Dorfman2022}. Here, the animal slows down and increases its turning rate after encountering a resource. After a fixed dwell time without encounters, it returns to the explore state. Since the state switching is governed by resource encounters and resource encounters are not uniformly distributed in the environment, the process results in spatially dependent stateful movement trajectories. The POI behavior further extends the movement hierarchy by introducing more spatial information to the decision making process. It is given explicit knowledge about high-value locations in the environment denoted as POIs. In exploration, the animal moves with bias in the direction of a selected POI, if no POI is selected the animal chooses a new POI uniformly at random from the nearest 10 in the environment. Upon arriving within a predetermined distance from a POI, the animal switches to the exploit state. It stays in the exploit state for a predetermined amount of time before returning to the explore state. Compared to CRW and EE, the process generates directed movement towards spatial locations with exploitative behavior around points of interest. The final movement model, LPOI, directly extends the POI behavior by incorporating preference values for each POI. The next target selection in the explore state is modified to use an \(\epsilon\)-greedy strategy with selection among the 10 nearest POIs. In simulation, preferences are updated dynamically based on the animals experience. The preference value for a location is increased when encountering a resource, and decreased when disturbance is induced. All movement models are modulated by disturbance. 
    
    Let \(D_{a,t}\in[0,1]\) denote the total disturbance acting on animal \(a\) at time \(t\), and let \(\hat{\mathbf{e}}_{a,t}\) denote the corresponding escape direction, as defined by the disturbance model. Disturbance induces one of three response states: calm, avoid, or flee. If
    \(
    D_{a,t}\leq \tau_{\mathrm{avoid}},
    \)
    the animal remains calm and follows the movement selected by the movement model. If
    \(\tau_{\mathrm{avoid}} < D_{a,t}\leq \tau_{\mathrm{flee}},\)
    the animal enters the avoid state. Here, the direction \(\mathbf{u}^{\mathrm{nom}}_{a,t+1}\) is blended with the escape direction,
    $$\mathbf{u}_{a,t+1}=\operatorname{unit}\!\left((1-\eta)\mathbf{u}^{\mathrm{nom}}_{a,t+1} + \eta \hat{\mathbf{e}}_{a,t}\right),$$
    with \(I_{a,t}^{\mathrm{avoid}}=1\) and \(I_{a,t}^{\mathrm{flee}}=0\). If
    \(D_{a,t} > \tau_{\mathrm{flee}},\)
    the animal enters the flee state, the step update is overridden, the direction is set to
    \(\mathbf{u}_{a,t+1}=\hat{\mathbf{e}}_{a,t},\)
    and speed is set to its maximum, with \(I_{a,t}^{\mathrm{avoid}}=0\) and \(I_{a,t}^{\mathrm{flee}}=1\). 
    
    These four behaviors collectively define a hierarchy of increasing structural and decision-making complexity. CRW represents unstructured movement with directional persistence \cite{Codling2008, Fagan2014}, EE introduces state-dependent search behavior (ARS) \cite{Dorfman2022}. Finally, POI and LPOI introduce spatially anchored, value-driven navigation \cite{Matthiopoulos2003}. The shared disturbance-response layer allows all four movement models to generate calm, avoidance, and flight behavior under drone disturbance. In the framework, the hierarchy enables controlled variation in movement structure, allowing the study to evaluate how different behavioral assumptions influence learned monitoring policies.

    The speed parameters for CRW configurations were estimated using an autoregressive formulation fitted to successive observed speeds and then rescaled to the simulator time step of 0.1 s. Directional persistence and turning noise were fitted separately by simulating candidate parameters and comparing distributional scores to determine suitable parameters. In the case of two state movement models (EE, POI and LPOI), one crw model was fitted for each state, producing one set of parameters for each state (explore and exploit). 

    For movement models that require states (EE, POI and LPOI), a distinction was made in the empirical data between the two possible states. The state was inferred using \(k\)-means clustering with two clusters, applied to standardized values of speed, absolute turning angle, and tortuosity. To reduce rapid state switching, inferred states were smoothed using a rolling median filter with a centred window of 5 steps. The cluster with higher mean absolute turning angle was assigned to the exploit state, the other was assigned to the explore state. Dwell time for EE, POI and LPOI was estimated as the median duration of time spent in every contiguous exploit sequence. POIs for evaluation were inferred using DBSCAN clustering for POI and LPOI models. For LPOI evaluation, the preference for each location was estimated as the revisitation rate to the POI. Final fit performance was evaluated using the normalized Wasserstein distance for speed and tortuosity, normalized binned circular Wasserstein distance with 72 bins for turning angle, and finally, revisitation score defined as $d_{\mathrm{revisit}} =
    \left|
    \frac{N^{\mathrm{sim}}_{\mathrm{repeat}}}
         {N^{\mathrm{sim}}_{\mathrm{entry}}}
    -
    \frac{N^{\mathrm{emp}}_{\mathrm{repeat}}}
         {N^{\mathrm{emp}}_{\mathrm{entry}}}
    \right|.$ Simulation was conducted over 20 seeds with 50000 steps in each seed. Simulated runs were evaluated against full empirical trajectories yielding the final distributional scores for speed, turn, tortuosity and revisit of which the mean was taken as the fit score.

\subsection{Drone model}
The simulated drones are represented as simplified aerial sensing agents. The purpose is to study the higher-level decision-making problem of how an aerial robot should position itself relative to animals in order to obtain useful observations whilst minimizing induced disturbance. Therefore, low-level flight control, motor dynamics or attitude stabilization are considered out of scope. Each drone $d$ has position $\mathbf{x}_{d,t} = (x_{d,t},y_{d,t},z_{d,t}) \in \mathbb{R}^3$, a unit velocity direction $\hat{\mathbf{u}}_{d,t}$, speed $v_{d,t}$, and viewing direction $\hat{\mathbf{c}}_{d,t}$. The policy outputs a continuous action $\mathbf{a}_{d,t} = (a_x,a_y,a_z,a_v,a_\theta)$, consisting of a movement direction, a normalized speed command, and a camera yaw command. Table \ref{tab:drone_state_action} summarizes the drone state and action quantities used in the simulation.

    \begin{table}[!ht]
    \centering
    \setlength{\tabcolsep}{5pt}
    \renewcommand{\arraystretch}{1.15}
    \caption{Drone state and action quantities used in the simulation.}
    \label{tab:drone_state_action}
    \begin{tabular}{lll}
    \hline \hline
    \textbf{Quantity} & \textbf{Symbol} & \textbf{Description} \\
    \hline \hline
    Position & \(\mathbf{x}_{d,t}\) & Drone position in three-dimensional space \\
    Velocity direction & \(\hat{\mathbf{u}}_{d,t}\) & Unit direction of translational motion \\
    Speed & \(v_{d,t}\) & Translational speed, bounded by drone limits \\
    Viewing direction & \(\hat{\mathbf{c}}_{d,t}\) & Unit vector describing camera direction \\
    Action & \(\mathbf{a}_{d,t}\) & Direction, speed, and camera yaw command \\
    Camera yaw & \(\theta_{d,t}\) & Bounded camera rotation command \\
    \hline \hline
    \end{tabular}
    \end{table}
    
    Consequently, the drone action space consists of five continuous high-level control dimensions, three for normalized translational movement direction, one for a normalized speed command mapped to \([v_{\min}, v_{\max}]\), and one for a camera yaw command mapped to \([-\theta_{\max}, \theta_{\max}]\).
    
    The drone movement dynamics are constrained for purposes of an increased simulation fidelity. Specifically, the speed is constrained by a drone-specific speed interval $[v_{\min},v_{\max}]$. The camera command is constrained by a bounded yaw rotation $[-\theta_{\max},\theta_{\max}]$. Consequently, the drone position is updated according to the holonomic kinematic model
    \(
    \mathbf{x}_{d,t+1} = \mathbf{x}_{d,t} + v_{d,t}\hat{\mathbf{u}}_{d,t}\Delta t,
    \)
    subject to additional altitude and speed constraints. The model dynamics correspond to a high-level velocity-control abstraction where low-level stabilization is assumed to be handled by an onboard flight controller.
    
    The viewing direction is updated by rotating the camera direction in yaw. Specifically, the horizontal camera heading is updated by interpolation toward the commanded direction, producing a smoothed camera motion rather than an instantaneous discontinuous rotation. Accordingly, the specified drone model enables the policy to control both drone movement and viewpoint direction whilst keeping the action space compact.
    
    Each drone is equipped with a directional camera model defined by a horizontal For \(\phi_d\), vertical For \(\psi_d\), sensing range \(R_d\), and viewing direction \(\hat{\mathbf{c}}_{d,t}\). For an animal \(a\), let \(\mathbf{r}_{d,a,t} = \mathbf{x}_{a,t} - \mathbf{x}_{d,t}\) denote the vector from the drone to the animal, with distance \(\rho_{d,a,t} = \|\mathbf{r}_{d,a,t}\|\). The animal is considered visible if it satisfies \(\rho_{d,a,t} \leq R_d\), \(|h_{d,a,t}| \leq \phi_d/2\), and \(|v_{d,a,t}| \leq \psi_d/2\), where \(h_{d,a,t}\) and \(v_{d,a,t}\) are the horizontal and vertical angular offsets in the camera frame. Thus, visibility depends on both distance and viewing geometry. 
    
    In order to aid the drone in target tracking in a partially observable environment, the observation space is extended with last-seen features for each animal target. This memory stores the most recent normalized drone to animal distance, angular offsets at which the animal was visible, and normalized time-since-seen variable. In addition, targets position and velocity components relative to the drone camera frame is also included upon animal visibility. The extended features provide the policy with short-term tracking data. Furthermore, they enable the drone to exploit practically available sensor information (e.g., depth cameras) when orienting itself in the partially observable environment whilst executing its task objective.

\subsubsection{Animal disturbance model}

    The disturbance calculation for an animal is conducted using a two stage process. First, all individual disturbances caused by each drone are calculated for each animal. Second, individual disturbances are aggregated into a saturating per animal disturbance total. 
    
    For each drone animal pair, let \(\mathbf{r} = (d_x,d_y,d_z)\) be the relative position from the animal to the drone, then define the horizontal and vertical separations as \(\rho = \sqrt{d_x^2 + d_y^2}\) and \(z = |d_z|\). Using the separations, we introduce a normalized stand-off measure \(s = \frac{\rho}{S_{xy}} + \frac{z}{S_z}\), where \(S_{xy}\) and \(S_z\) are horizontal and vertical distance scales. The normalized stand-off measure defines a distance gate \(g_{\mathrm{dist}} = e^{-s}\) and a base utility \(u_0 = 1 - e^{-s}\). As a result, geometric terms have the strongest effect at short range. Additionally, larger stand-off distances increase the base utility.
    
    The model includes three bounded geometric disturbance terms \(g_{\angle}(\mathbf{r})\), \(g_{\mathrm{head}}(\mathbf{r},\mathbf{v}_d)\), and \(g_{\mathrm{axis}}(\mathbf{r},\mathbf{v}_a)\), representing, the effect of vertical angle, drone heading relative to the animal, and the drone's position relative to the animal's movement axis. The terms are distance-gated as \(\tilde g_i = g_i\, g_{\mathrm{dist}}\), where \(i \in \{\angle,\mathrm{head},\mathrm{axis}\}\).
    
    Let \(w_{\angle}, w_{\mathrm{head}}, w_{\mathrm{axis}} \ge 0\) be component weights. The single-drone utility is then
    \[
    u
    =
    u_0
    \prod_{i \in \{\angle,\mathrm{head},\mathrm{axis}\}}
    \operatorname{clip}_{[0,1]}\!\left(1 - w_i \tilde g_i\right),
    \]
    where \(\operatorname{clip}_{[0,1]}(x)=\min\{1,\max\{0,x\}\}\). The resulting pairwise disturbance is \(\delta = 1 - \operatorname{clip}_{[0,1]}(u)\). For multiple drones, let \(\delta_{ad}\) denote the disturbance induced by drone \(d\) on animal \(a\), including a drone-specific multiplier \(m_d\), such that \(\delta_{ad} = m_d\,\delta(\mathbf{r}_{ad},\mathbf{v}_d,\mathbf{v}_a)\). 
    
    In addition, let the corresponding escape direction away from drone \(d\), and the diminishing-return accumulation variables, be defined as
    \[
    \begin{aligned}
    \mathbf{e}_{ad}
    &=
    \begin{cases}
    -\dfrac{\mathbf{r}_{ad}}{\|\mathbf{r}_{ad}\|}, & \|\mathbf{r}_{ad}\| > 0,\\[1.5ex]
    \mathbf{0}, & \text{otherwise},
    \end{cases}
    \qquad
    \left\{
    \begin{aligned}
    I_a^{(k)} &= \bigl(1 - D_a^{(k-1)}\bigr)\,\delta_{a(k)}, \\
    D_a^{(k)} &= D_a^{(k-1)} + I_a^{(k)}, \\
    \mathbf{E}_a^{(k)} &= \mathbf{E}_a^{(k-1)} + I_a^{(k)}\mathbf{e}_{a(k)}.
    \end{aligned}
    \right.
    \end{aligned}
    \]
    Here, \(D_a^{(0)} = 0\) and \(\mathbf{E}_a^{(0)} = \mathbf{0}\), and the pairwise disturbances are sorted in descending order before accumulation. The final disturbance is
    \(D_a = \operatorname{clip}_{[0,1]}\!\bigl(D_a^{(N)}\bigr)\). The final escape direction is
    \[
    \hat{\mathbf{e}}_a
    =
    \begin{cases}
    \dfrac{\mathbf{E}_a^{(N)}}{\|\mathbf{E}_a^{(N)}\|}, & \|\mathbf{E}_a^{(N)}\| > 0,\\[2ex]
    \dfrac{\mathbf{v}_a}{\|\mathbf{v}_a\|}, & \text{otherwise}.
    \end{cases}
    \]
        
    With this definition, disturbance increases when drones are closer, use more targeted approaches, or approach animals in larger numbers. Specifically, the base distance gate \(g_{\mathrm{dist}}\) reflects the role of proximity and altitude in shaping wildlife disturbance \cite{BrissonCuradeau2025, Afridi2025}. Geometric modifiers further capture specific threatening patterns, the vertical angle term \(g_{\angle}\) accounts for the elevated disturbance caused by overhead approaches \cite{Vas2015}. Additionally, the heading (\(g_{\mathrm{head}}\)) and axis (\(g_{\mathrm{axis}}\)) terms penalize direct, target-oriented flights, which are known to provoke stronger avoidance than predictable, tangential paths \cite{mulero2017}. 
    
    For multi-drone scenarios, the aggregation is saturating, meaning that once an animal is already highly disturbed, additional drones have a progressively smaller effect. The aggregation mirrors the Multiplicative Risk Model used in ecology to quantify the combined risk from multiple simultaneous predators \cite{Soluk1988, Sih1998}. Additionally, it is consistent with the risk allocation hypothesis \cite{Lima1999}.

\begin{figure}[ht]
        \centering
        \includegraphics[width=1\linewidth]{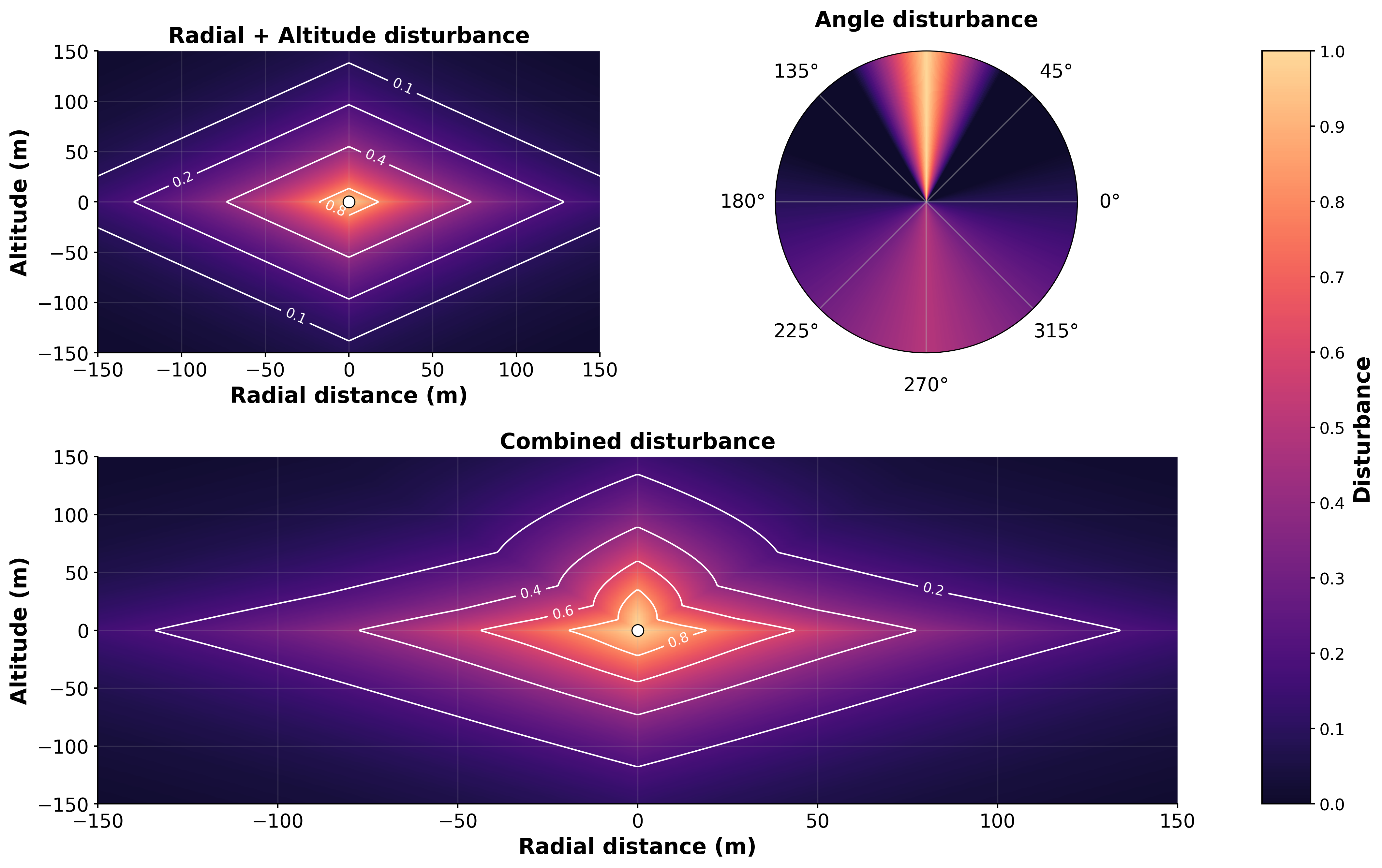}
        \caption{Visual representation of static disturbance function. Top left: Combined radial and altitude gain functions. Top right: angular gain penalizing unfavorable vertical angles. Bottom: combined angle-shaped distance gain illustrating the spatial region where disturbance sensitivity is highest.}
        \label{fig:static_disturbance}
    \end{figure}

    Figure \ref{fig:static_disturbance} visualizes the partial disturbances \(g_{\mathrm{dist}}\) and \(g_{\angle}\) as well as the combined disturbance \(D_a\).
    
\subsection{Reward Function}

    The objective of the reward design is to encourage disturbance-aware monitoring, not naive target pursuit. The policy is at each time step rewarded for maintaining maximally informative observations of the assigned target whilst simultaneously avoiding inducing excessive disturbance, exhibit unnecessary motion, monitor the target from unsafe proximity, or lose the animal for a prolonged amount of steps. 
    
    For a specific drone $d$, let $y_{d,t}^{+}\in\{0,1\}$ signify whether its assigned target animal is in view. Moreover, let $d_{d,t}^{+}\in[0,1]$ designate the normalized distance to the target given its visibility. Then define the distance reward as $r_{d,t}^{\mathrm{dist}} = -d_{d,t}^{+}$ if \(y_{d,t}^{\star}=1\) and \(r_{d,t}^{\mathrm{dist}} = -1\) otherwise. The mean distance reward across drones is then $\bar r_t^{\mathrm{dist}}=\frac{1}{N_d}\sum_{d=1}^{N_d} r_{d,t}^{\mathrm{dist}}$. Moreover, let $D_{a,t}\in[0,1]$ represent the disturbance induced on animal $a$ at time step $t$. Similarly, the mean disturbance penalty is then defined as  $p_t^{\mathrm{disturb}}=\frac{1}{N_a}\sum_{a=1}^{N_a}D_{a,t}$. The main monitoring-disturbance trade-off is computed as
    $$
    r_t^{\mathrm{trade}} =
    \operatorname{clip}_{[0,1]}
    \left(
    \frac{
    \alpha(1-p_t^{\mathrm{disturb}}) + (1-\alpha)\bar r_t^{\mathrm{dist}}
    }{s_r}
    \right),
    $$
    where $\alpha\in[0,1]$ controls the trade-off and $s_r$ is a normalization constant. Furthermore, in order to encourage stable visual tracking, an alignment reward is added. Specifically, if the target is visible then $v_{d,t}^{+}$ and $h_{d,t}^{+}$ define the normalized vertical and horizontal angular offsets. Consequently, the alignment score is defined as
    $$
    r_{d,t}^{\mathrm{align}} =
    \operatorname{clip}_{[0,1]}
    \left(
    1 -
    \frac{
    \max(0,|v_{d,t}^{\star}|-\delta) + \max(0,|h_{d,t}^{\star}|-\delta)
    }{2}
    \right),
    $$
    and only contributes when the target is visible. The average alignment reward is $\bar r_t^{\mathrm{align}}=\frac{1}{N_d}\sum_{d=1}^{N_d} y_{d,t}^{+} r_{d,t}^{\mathrm{align}}$. The final monitoring reward is uniquely defined in the single and multi-drone cases. The monitoring reward in the single drone case is defined as $r_t^{\mathrm{monitor}} = w_{\mathrm{trade}} r_t^{\mathrm{trade}} + w_{\mathrm{align}} \bar r_t^{\mathrm{align}}$. In the multi-drone case, this is extended to
    $
    r_t^{\mathrm{monitor}} =
    w_{\mathrm{trade}} r_t^{\mathrm{trade}} +
    w_{\mathrm{align}} \bar r_t^{\mathrm{align}} +
    w_{\mathrm{sep}} r_t^{\mathrm{sep}} +
    r_t^{\mathrm{bonus}}.
    $ Given that all drones observe their targets then an all-visible bonus reward is defined as $r_t^{\mathrm{bonus}} = b_{\mathrm{all}}$. All drones are defined to observe their targets upon the satisfaction of the following equality $\frac{1}{N_d}\sum_{d=1}^{N_d} y_{d,t}^{+}=1$, and otherwise set to $0$. For purposes of simulation fidelity, drones are encouraged to observe the target from diverse viewing angles. For two unit direction vectors $\mathbf{q}_i$ and $\mathbf{q}_j$ (pointing from the animal to each observant drone), the pairwise score is $s_{ij} = \frac{1-\mathbf{q}_i^\top \mathbf{q}_j}{2}$. Subsequently the overall separation reward is
    $$
    r_t^{\mathrm{sep}} =
    \frac{1}{|\mathcal{A}_{\mathrm{sep}}|}
    \sum_{a\in\mathcal{A}_{\mathrm{sep}}}
    \frac{1}{|\mathcal{P}_a|}
    \sum_{(i,j)\in\mathcal{P}_a}
    \frac{1-\mathbf{q}_{i,a}^{\top}\mathbf{q}_{j,a}}{2},
    $$ where $ \frac{1}{|\mathcal{A}_{\mathrm{sep}}|}$ defines the number of animals considered, $\mathcal{P}_a = \binom{n}{2}=\frac{n(n-1)}{2}$ signifies all unique drone pairs around the animal, $\mathbf{q}_{i,a}^{\top}\mathbf{q}_{j,a}$ denotes two unit vectors that point from the animal to two unique drones $i$ and $j$. The formula encourages each pair of drones to have a diametrically opposed view direction, encouraging diversification of visual data retrieval while also serving as an antidote against the formation of drone clusters. Noteworthy is that as drones are added, the 3D space constrains the quantity of drone pairs that can be nearly opposite. Moreover, several auxiliary penalties are subtracted. The first three penalties are designed to smooth policy actions. Larger linear or rotational movements incur higher penalties. Specifically, 
    \[
    \begin{aligned}
    p_t^{\mathrm{speed}}
    &=
    \frac{1}{N_d}\sum_{d=1}^{N_d}
    \left(
    \beta_v \frac{v_{d,t}}{v_{d,\max}}
    \right)^{q_v},
    \qquad
    p_t^{\mathrm{yaw}}
    =
    \frac{1}{N_d}\sum_{d=1}^{N_d}
    \left(
    \beta_\theta \frac{|\theta_{d,t}|}{\theta_{d,\max}}
    \right)^{q_\theta},
    \\[8pt]
    p_t^{\mathrm{turn}}
    &=
    \frac{1}{N_d}\sum_{d=1}^{N_d}
    \left[
    \beta_{\Delta}
    \max\!\left(0,-\langle \hat{\mathbf{u}}_{d,t},\hat{\mathbf{u}}_{d,t-1}\rangle\right)
    \frac{v_{d,t}}{v_{d,\max}}
    \right]^{q_\Delta}.
    \end{aligned}
    \]
    where $N_d$ is the number of drones, $v_{d,t}$ and $\theta_{d,t}$ are the linear speed and yaw rotation of drone $d$ at time $t$, $\hat{\mathbf{u}}_{d,t}$ is the unit velocity vector, $v_{d,\max}$ and $\theta_{d,\max}$ are the respective maximum speed and yaw, and $\beta_\cdot, q_\cdot$ are scaling and exponent parameters controlling penalty magnitude and sharpness. A track-loss penalty discourages losing the target: $p_t^{\mathrm{lost}} = \frac{1}{N_d}\sum_{d=1}^{N_d} \eta_{\mathrm{lost}} c_{d,t}.$ Additionally, discouraged animal behavioural responses are penalized via $p_t^{\mathrm{state}} = \frac{1}{N_a}\sum_{a=1}^{N_a}(I_a^{\mathrm{avoid}} + I_a^{\mathrm{flee}})$.
    
    Finally, the hard safety penalty is defined as
    $$
    p_t^{\mathrm{hard}} = c \;\; \mathbf{1}\Bigg(
    \min_{i \neq j} \|\mathbf{x}_{i,t} - \mathbf{x}_{j,t}\| < R_\mathrm{drone} 
    \;\;\text{or}\;\;
    \min_{i,a} \|\mathbf{x}_{i,t} - \mathbf{x}_{a,t}\| < R_\mathrm{animal}
    \Bigg),
    $$
    where $p_t^{\mathrm{hard}}$ denotes the fixed penalty which is applied when a hard safety violation occurs, $c$ is the penalty magnitude, $\mathbf{x}_{i,t}$ and $\mathbf{x}_{j,t}$ are the positions of drone $i$ and $j$ at time $t$, $\mathbf{x}_{a,t}$ is the position of animal $a$, and the radii $R_\mathrm{drone}$ and $R_\mathrm{animal}$ is the hard safety radius, in practice, the radii are often chosen to be a few meters.  
    
    Lastly, an episode is terminated if the maximum episode length is reached, target loss exceeds a grace period, or if a hard safety violation occurs.

    The final single-drone reward is
    $$
    r_t := r_t^{\mathrm{monitor}} - p_t^{\mathrm{speed}} - p_t^{\mathrm{yaw}} - p_t^{\mathrm{turn}} - p_t^{\mathrm{lost}},
    $$
    while the multi-drone reward is
    $$
    r_t := r_t^{\mathrm{monitor}} - p_t^{\mathrm{state}} - p_t^{\mathrm{lost}} - p_t^{\mathrm{speed}} - p_t^{\mathrm{yaw}} - p_t^{\mathrm{hard}}.
    $$

\subsection{Reinforcement learning formulation}

The animal-monitoring task is formulated as a RL problem in which one or more drones interact with a simulated environment. More specifically, the environment is modeled as a POMDP \cite{lauri_2023, Kaelbling1998}, implying that each drone policy only receives access to local observations rather than the global environment state. At each time step, every drone receives an observation and selects a continuous control action.

A rule-based controller was developed to serve as a benchmark and to motivate the introduction of RL-based controllers. The controller design is further described in the subsequent section. The learned methods evaluated in this study are DQN, PPO, and SAC, representing value-based off-policy, on-policy actor-critic, and off-policy actor-critic paradigms, respectively. Previous work has found PPO and SAC particularly suitable for continuous control problems \cite{Schulman2017, Haarnoja2018}.

The DQN implementation uses a branching dueling Q-network \cite{zhu2024effective}, where separate action branches represent movement direction, speed, and camera yaw. Since DQN operates over discrete actions, the continuous drone action space is discretized into fixed sets of direction vectors, speed bins, and yaw bins. PPO and SAC instead operate directly in the continuous action space. PPO employs a Gaussian actor-critic policy with clipped policy updates, whereas SAC uses entropy regularization together with twin Q-functions to improve exploration and stability. Further implementation details, including network architectures are provided in the Appendix.

The RL agents are trained entirely in simulation using synthetic animal trajectories generated by the four previously introduced movement models (CRW, EE, POI, and LPOI). Each training run uses a single movement model as the source of animal behavior, enabling the study to evaluate under what circumstances different structural movement priors generalize to empirical GPS trajectories.

\section{Experiments}
\label{sec:exp}
The experiments evaluate whether stealth-aware RL can produce monitoring policies that generalize across animal behaviours, empirical species trajectories, sensing configurations, and multi-agent settings. The evaluation is divided into five blocks: synthetic behaviour generalization, empirical trajectory transfer, movement-model fit, sensing-parameterization sensitivity, and multi-agent scaling.

\subsection{Simulated environment and scenario configuration}

All training and evaluation were performed in the proposed simulation environment. Animals move in a continuous horizontal plane, while drones move in three dimensions. The simulator uses a fixed time step of \(\Delta t=0.1\) s and a maximum episode length of 2048 steps, corresponding to 204.8 s per episode. The environment includes a procedurally generated spatial resource map based on OpenSimplex noise. The map defines spatially varying encounter probabilities and extracted points of interest. These structures are used by the EE, POI, and LPOI movement models to generate increasingly structured animal trajectories. At the start of each episode, animals and drones are initialized in safe configurations. Drones are spawned at drone-type-specific stand-off distances from their assigned targets, with randomized yaw perturbations. Episodes terminate when the time horizon is reached, when the assigned target is lost beyond a grace period, or when a hard safety constraint is violated. In order to ensure comparability across experiments, all training runs were conducted using a fixed random seed (seed 42). This ensures consistent initialization and reproducible stochasticity in the environment. Consequently, observed differences in training dynamics can primarily be attributed to the learning algorithm or underlying animal movement model, rather than variations in data generation. All training results were generated using an RTX 5070 Ti and a NVIDIA RTX 5090 GPU. The utilized operating system was Windows 10, nonetheless, the development and training occurred in Windows Subsystem for Linux 2. The simulation environment and learning framework were primarily implemented using Python modules PyTorch, NumPy and Matplotlib. Table \ref{tab:simulation_parameters} summarize the default paramter values used for the simulations. 

\begin{table}[!ht]
\centering
\setlength{\tabcolsep}{5pt}
\renewcommand{\arraystretch}{1.15}
\caption{Main simulation parameters used in the experiments.}
\label{tab:simulation_parameters}
\begin{tabular}{lll}
\hline \hline
\textbf{Parameter} & \textbf{Symbol} & \textbf{Value / Description} \\
\hline \hline
Time step & \(\Delta t\) & \(0.1\) s \\
Episode length & \(T\) & 2048 steps \\
Avoidance threshold & \(\tau_{\mathrm{avoid}}\) & Disturbance threshold for avoidance \\
Flee threshold & \(\tau_{\mathrm{flee}}\) & Disturbance threshold for fleeing \\
Avoidance blend & \(\eta\) & Escape-direction blending factor \\
Horizontal disturbance scale & \(S_{xy}\) & Horizontal stand-off scale \\
Vertical disturbance scale & \(S_z\) & Vertical stand-off scale \\
Evaluation repeats & \(N\) & 100 unless otherwise stated \\
\hline \hline
\end{tabular}
\end{table}

\subsection{Animal behavior settings}

Four synthetic movement models are evaluated: CRW, EE, POI, and LPOI. These models form a hierarchy of increasing behavioral structure. CRW represents unstructured movement with directional persistence. EE adds behavioral switching between exploration and local exploitation. POI introduces persistent spatial targets extracted from the resource map. LPOI further adds learned preferences over POIs, allowing the animal to revisit high-value locations and avoid locations associated with disturbance. This hierarchy allows policies to be tested across movement regimes ranging from weakly structured stochastic motion to memory-based spatial behavior. It also enables investigation of whether policies trained on more realistic movement priors transfer better to empirical trajectories.


\subsection{Robot configurations}

Three drone sensing parametrizations were introduced to investigate how sensing capabilities influence the monitoring-disturbance trade-off: D1, D2, and D3. The parametrizations share the same underlying kinematic drone model and should not be interpreted as separate physical drone morphologies. Instead they represent different sensing capabilities of the same aerial robot model. The parametrization differ in camera FoV, sensing range, maximum speed and disturbance footprint , as detailed in Table \ref{tab:drone_parameterizations}. A narrower and shorter-range sensor requires the drone to remain closer or more precisely aligned to maintain visibility. A wider and longer-range sensor allows the drone to monitor from larger stand-off distances and with greater tolerance to angular misalignment. This renders the parameterizations useful for purposes of evaluating how variations in sensing capabilities may affect the trade-off between observation quality and disturbance. Each drone observes its own relative sensing range and FoV enabling RL agents to distinguish between the various drone types it outputs actions for. The relevance of differing capabilities is especially pronounced in the heterogeneous multi-drone experiments, where drones with different sensing capabilities may learn complementary monitoring roles. In the multi-agent experiments, both homogeneous and heterogeneous fleets are evaluated. Homogeneous fleets contain drones with identical parameterizations, while heterogeneous fleets contain mixtures of D1, D2, and D3. This allows testing of whether mixed sensing capabilities produce complementary monitoring roles or whether weaker drones constrain team performance.
    
\begin{table}[!ht]
\centering
\setlength{\tabcolsep}{5pt}
\renewcommand{\arraystretch}{1.15}
\caption{Drone parametrizations used to evaluate sensing capability and mobility.}
\label{tab:drone_parameterizations}
\begin{tabular}{llll}
\hline \hline
\textbf{Parameter} & \textbf{D1} & \textbf{D2} & \textbf{D3} \\
\hline \hline
Vertical FOV & \(70^\circ\) & \(90^\circ\) & \(110^\circ\) \\
Horizontal FOV & \(100^\circ\) & \(140^\circ\) & \(170^\circ\) \\
View range & 120 m & 200 m & 280 m \\
Maximum camera rotation & \(90^\circ\) & \(90^\circ\) & \(90^\circ\) \\
Maximum altitude & 150 m & 150 m & 150 m \\
Maximum speed & 15 m/s & 20 m/s & 25 m/s \\
Disturbance multiplier & 1.0 & 1.1 & 1.2 \\
Spawn distance & 70--90 m & 90--110 m & 110--130 m \\
\hline \hline
\end{tabular}
\end{table}

\subsection{Training protocol}

The learned controllers are compared against a tuned rule-based baseline. The baseline is a centroid stand-off controller that attempts to keep the target visible at a desired normalized distance and altitude. When the target is visible, proportional feedback controls forward motion, vertical motion, and camera yaw using the observed target distance and angular offsets. When the target is not visible, the controller switches to a search mode and applies a constant yaw rotation while maintaining a target altitude. Baseline parameters are selected by grid search. Three reinforcement-learning algorithms were evaluated in the single-agent experiments: DQN, PPO, and SAC. DQN uses a branching dueling architecture over a discretized action space. PPO and SAC operate directly in the continuous action space. MAPPO is used for the multi-agent experiments because PPO demonstrated stable convergence during training and is less sensitive to the instability observed in SAC under larger multi-agent settings. Hyperparameters are provided in the Appendix. All training runs use synthetic animal trajectories generated from one of the four movement models. Evaluation is performed without exploration noise. Results are reported using total reward and decomposed reward components to distinguish between strategies that achieve similar reward through different monitoring--disturbance trade-offs.

\subsection{Evaluation metrics}

The results were analyzed both qualitatively and quantitatively. The main performance metric is normalized total reward. Examples of high and low reward episodes are provided in Appendix~\ref{app:reward_interpreataion} to aid interpretation. To further interpret policy behavior, total reward is decomposed into monitoring reward and disturbance penalty. Monitoring reward measures the quality of target observation, including visibility, distance, and camera alignment. Disturbance penalty measures the mean modeled behavioral impact induced in the animal. Drone movement patterns relative to each animal were inspected visually; see Appendix. Two types of drone-to-animal visualizations were used. The first is an XY-to-Z view, showing disturbance contour lines together with the time-aggregated density of drone positions relative to the animal. The second is a top-down view, where drone positions are represented relative to the animal’s forward heading, highlighting how the drone moved around the animal over time in terms of viewing angles and relative positioning. Policy robustness was also assessed by measuring how far the drone could progress without losing track of the animal. A successful policy should maintain continuous visual contact. Unless otherwise stated, each configuration is evaluated over \(N=100\) episodes. Additional metrics are used in specific experiments. For movement-model fit, empirical and simulated trajectories are compared using normalized discrepancies in speed, turning angle, tortuosity, and revisitation behavior. For multi-agent experiments, coordination is evaluated using stand-off distance, viewpoint-separation score, and performance degradation under scaling. Behavioral impact is further summarized by the fraction of time animals spend in calm, avoidance, and fleeing states.

\subsection{Experimental scenarios}

The first experiment evaluates synthetic behavior generalization. Each algorithm is trained and evaluated on CRW, EE, POI, and LPOI movement. The second experiment evaluates transfer to empirical GPS trajectories. Policies trained on synthetic movement are evaluated while animals replay processed empirical trajectories from jackals, pigeons, and spur-winged lapwings. The third experiment compares movement-model fit with transfer performance. This tests whether synthetic movement priors that better match empirical statistics also produce better monitoring policies. The fourth experiment evaluates sensitivity to drone sensing parameterization. Policies are compared across D1, D2, and D3 to measure how FoV and sensing range affect the observation--disturbance trade-off. The fifth experiment evaluates multi-agent scaling and heterogeneous fleets. We test single-target monitoring with multiple drones, proportional scaling of drones and animals, and mixed teams with different sensing parameterizations.

\section{Results}
\label{sec:results}
\begin{figure}[!hb]
        \centering
        \includegraphics[width=.99\linewidth]{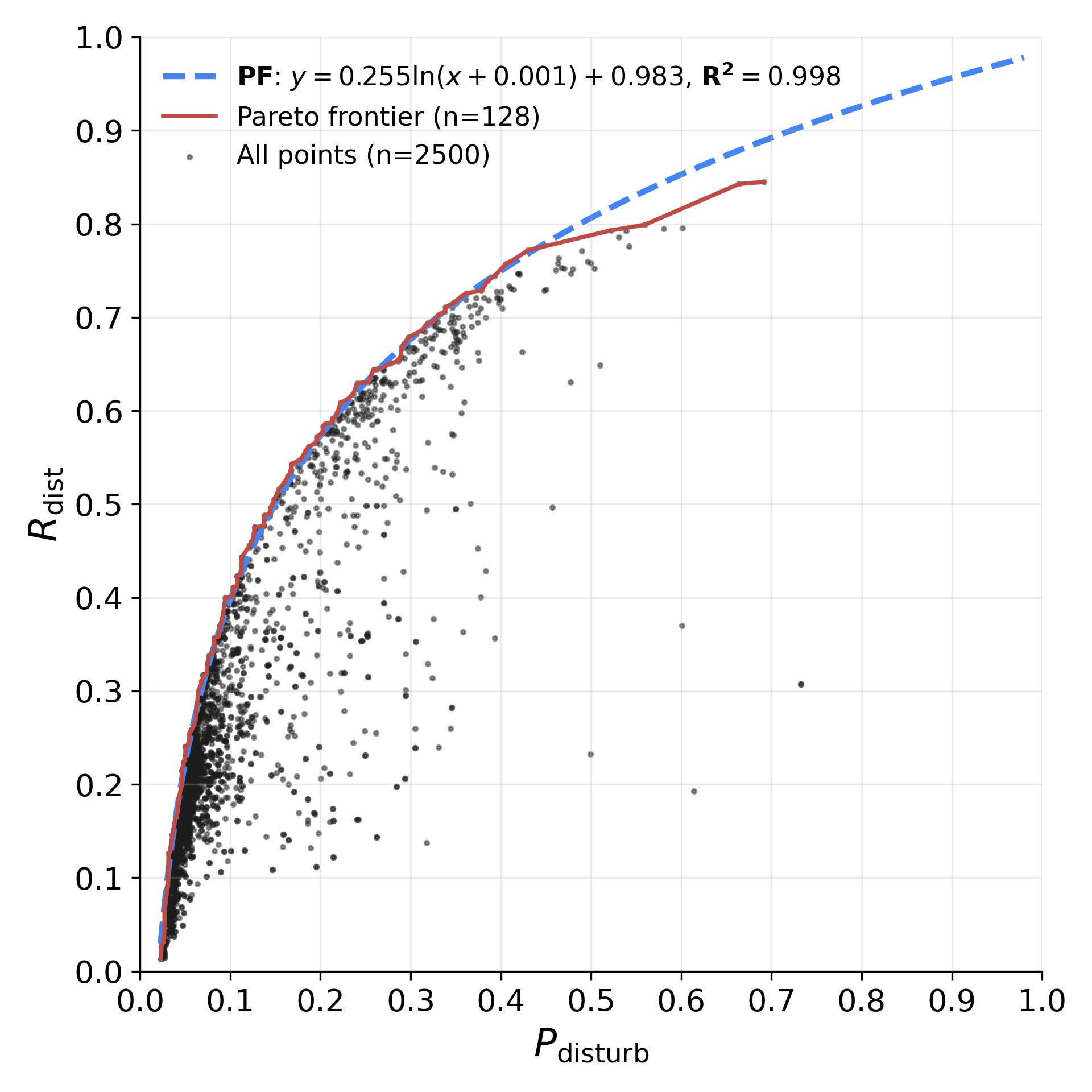}
        \caption{Pareto frontier between monitoring reward and disturbance penalty across all evaluated policy configurations. Each point represents an evaluated simulation. The dashed curve shows a fitted Pareto frontier, approximating the upper envelope of achievable trade-offs.}
        \label{fig:pareto}
    \end{figure}

Figure \ref{fig:pareto} presents the Pareto frontier \cite{teich2001pareto} (i.e., the set of optimal solutions in a multi-objective problem where no single metric can be improved without sacrificing another) between the probability for disturbance and the distance of the drown from the animal. The figure highlights a clear trade-off between monitoring reward and disturbance. At very low disturbance levels, monitoring reward is also limited, indicating that policies which remain highly conservative tend to avoid disturbing the animals but also collect less useful monitoring information. As disturbance increases, monitoring reward rises rapidly at first, suggesting that a moderate reduction in stand-off distance can substantially improve visibility and tracking quality. However, the curve gradually flattens, indicating diminishing returns beyond a certain disturbance level. Further increases in disturbance yield only small improvements in monitoring reward. For the remaining results, we used the Pareto optimal configuration. 

    Table \ref{tab:behavior_results_full} compares the evaluated models across the four synthetic behaviours used for training. All of the four learned models (DQN, PPO and SAC) consistently outperform the rule-based controller baseline in total reward. This is indicative that the learned models are better at adapting to the dynamic disturbance geometry compared to the rule-based baseline. Among the learned models, SAC demonstrates both the highest and most consistent performance. It achieves the highest total reward across all movement types while at the same time attaining the lowest standard deviation. DQN performs well for CRW and POI while PPO performs well for CRW and EE. In addition, the reward decomposition reveals a clear difference in strategy between the learned models. DQN often achieves the highest distance reward at the cost of a higher disturbance penalty. PPO consistently yields the lowest disturbance penalty and the lowest distance reward showing a more conservative policy. SAC generally finds an intermediate policy between DQN and PPO.

        \begin{table}[!ht]
        \centering
        \setlength{\tabcolsep}{4pt}
        \caption{Performance comparison across behaviors. Evaluated on unseen synthetically generated episodes. Results are reported as mean $\pm$ standard deviation across \(n=100\) evaluation episodes. }
        \label{tab:behavior_results_full}
        \begin{tabular}{llcccc}
        \hline \hline
        \textbf{Model} & \textbf{Metric} & \textbf{CRW} & \textbf{EE} & \textbf{POI} & \textbf{LPOI} \\
        \hline \hline
        Rule-based & Total reward & $0.772 \pm 0.024$ & $0.825 \pm 0.022$ & $0.771 \pm 0.022$ & $0.774 \pm 0.018$ \\
        & Distance reward & $0.485 \pm 0.006$ & $0.497 \pm 0.004$ & $0.483 \pm 0.007$ & $0.474 \pm 0.007$ \\
        & Disturbance penalty & $0.217 \pm 0.008$ & $0.198 \pm 0.008$ & $0.215 \pm 0.007$ & $0.207 \pm 0.004$ \\
        \midrule
        DQN & Total reward & $0.909 \pm 0.006$ & $0.886 \pm 0.009$ & $0.916 \pm 0.010$ & $0.855 \pm 0.048$ \\
        & Distance reward & $0.585 \pm 0.018$ & $0.546 \pm 0.013$ & $0.554 \pm 0.010$ & $0.517 \pm 0.009$ \\
        & Disturbance penalty & $0.216 \pm 0.016$ & $0.193 \pm 0.009$ & $0.182 \pm 0.007$ & $0.176 \pm 0.009$ \\
        \midrule
        PPO & Total reward & $0.914 \pm 0.013$ & $0.911 \pm 0.054$ & $0.867 \pm 0.028$ & $0.898 \pm 0.015$ \\
        & Distance reward & $0.532 \pm 0.029$ & $0.550 \pm 0.020$ & $0.467 \pm 0.051$ & $0.522 \pm 0.025$ \\
        & Disturbance penalty & $0.170 \pm 0.019$ & $0.184 \pm 0.015$ & $0.142 \pm 0.037$ & $0.169 \pm 0.020$ \\
        \midrule
        SAC & Total reward & $0.936 \pm 0.004$ & $0.919 \pm 0.005$ & $0.924 \pm 0.010$ & $0.937 \pm 0.011$ \\
        & Distance reward & $0.565 \pm 0.011$ & $0.569 \pm 0.010$ & $0.527 \pm 0.020$ & $0.573 \pm 0.012$ \\
        & Disturbance penalty & $0.184 \pm 0.009$ & $0.194 \pm 0.007$ & $0.159 \pm 0.012$ & $0.189 \pm 0.008$ \\
        \hline \hline
        \end{tabular}
        \end{table}
                
\begin{figure}[h]
            \centering
            \includegraphics[width=0.99\linewidth]{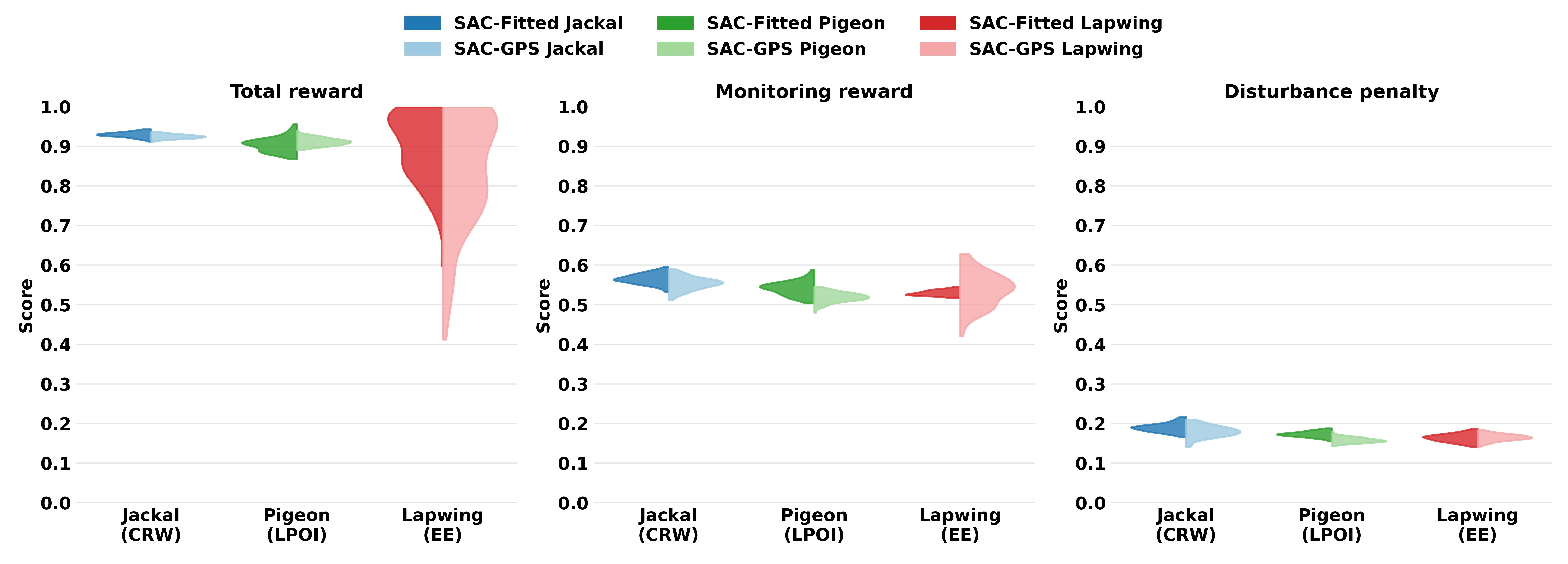}
            \caption{
            Comparison between SAC policies trained on fitted simulated movement priors and SAC policies trained directly on empirical GPS trajectories. Both policy variants were evaluated on 100 real GPS replay episodes using D2 drone sensing capabilities and no wind. The fitted condition uses policies trained on simulated animal movement models fitted to each species, while the GPS condition uses policies trained directly on the corresponding empirical GPS movement data. Split violins show the distribution of total reward, monitoring reward, and disturbance penalty across evaluation episodes for Jackal, Pigeon, and Spur-winged lapwing.
            }
            \label{fig:sac_vs_gps}
        \end{figure}

        Overall, SAC consistently obtains the best transfer performance, $0.929 \pm 0.007$ for Jackals when trained with EE, $0.925\pm0.015$ for Pigeons and $0.933\pm0.018$ for Spur-winged lapwings, both trained with LPOI. Notably, PPO underperformed on all animals, in particular on Pigeons. Nevertheless, it similarly managed to obtain the lowest disturbance penalties across all animals, opting for a conservative policy that prioritizes safe monitoring. DQN managed to achieve its best performance on all animals through training on the Correlated-Random-Walk animal movement-model. In this context, both PPO and SAC obtained their highest test score on Jackals utilizing their policies that were trained on the EE model. Performance in terms of a low standard deviation remains consistent across all RL-models and animal pairs. Similarly, all of the models manage to converge to policies that do not trigger disturbing animal behaviors (avoidance is triggered at disturbance 0.40 and fleeing behavior at 0.70). Figure \ref{fig:sac_vs_gps} further highlights and provides a detailed overview of the transferability of performance of the SAC algorithm. The figure compares the performance to the out-of-sample and real-valued animal GPS trajectories. Not only are the monitoring and disturbance penalties highly similar, but also the total reward. The similarity in total reward signifies that the drone managed to maintain smoothness in its movements over time. Similar observations are present across all of the three tested animals. 
    
        Next, in order to evalaute the framework's robustness, we conducted several tests. We first analyse potential performance increase under the scaling of drones around a target animal. The results demonstrate that performance improves slightly when increasing from one to two drones as indicated in Table \ref{tab:single_target_coordination}. This implies that limited multi-agent cooperation can enhance monitoring through complementary viewpoints whilst simultaneously maintaining low disturbance. However, further scaling to three drones results in a significant performance drop. This degradation is accompanied by an increase in stand-off distance without a corresponding increase in angular separation. Taken together, this suggests that naively increasing the amount of drone agents results in a coordination overhead, which in turn implies more conservative policies. These results indicate the presence of an optimal team size for single-target monitoring, beyond which adding agents yields diminishing returns and negatively impacts performance.

        \begin{table}[!ht]
        \centering
        \caption{Single-target scaling with homogeneous drones. Distance and angular separation statistics are reported as mean $\pm$ std.}
        \label{tab:single_target_coordination}
        
        \resizebox{0.80\textwidth}{!}{
        \begin{tabular}{cccccc}
        \toprule
        \toprule
        Case & Drone & \# Drones & Reward & Mean distance (m) & Mean angular sep. \\
        \midrule
        \midrule
        A & D2 & 1 & 0.851 $\pm$ 0.012 & 96.83 $\pm$ 6.727 & Undefined \\
        B & D2 & 2 & 0.860 $\pm$ 0.003 & 136.0 $\pm$ 8.928 & 0.844 $\pm$ 0.086 \\
        C & D2 & 3 & 0.700 $\pm$ 0.006 & 145.5 $\pm$ 17.27 & 0.827 $\pm$ 0.077 \\
        \bottomrule
        \bottomrule
        \end{tabular}
        }
        \end{table}

        Upon proportionally scaling both the number of drones and animals, performance degrades substantially despite maintenance of a constant drone-to-animal ratio as presented in Table \ref{tab:scaling}. This is indicative of that the accomplishment of performance does not scale linearly. Additional drones and animals introduce coordination and allocation challenges. In particular, the problem transitions from independent tracking to a coupled multi-agent assignment problem, where inefficient division of labor and increased interference between agents lead to reduced monitoring performance. Furthermore, we can observe an increase in the stand-off distance, suggesting a more conservative behavior, likely driven by overlapping disturbance constraints. Additionally, the rise in variance highlights increased instability as system size grows.

        \begin{table}[ht]
        \centering
        \caption{Joint scaling with homogeneous morphology at a constant drone-to-animal ratio. Both the number of drones and animals increase proportionally. Reward, distance, reward, and angular separation statistics are reported as mean $\pm$ std.}
        \label{tab:scaling}
        
        \resizebox{0.8\textwidth}{!}{
        \begin{tabular}{cccccc}
        \toprule
        \toprule
        Morphology & \# Drones & \# Animals & Reward & Mean distance (m) \\
        \midrule
        \midrule
        D2 & 1 & 1 & 0.851 $\pm$ 0.012 & 96.83 $\pm$ 6.727 \\
        D2 & 2 & 2 & 0.760 $\pm$ 0.232 & 116.1 $\pm$ 22.39 \\
        D2 & 3 & 3 & 0.439 $\pm$ 0.265 & 119.4 $\pm$ 26.99 \\
        \bottomrule
        \bottomrule
        \end{tabular}
        }
        \end{table}
        
        In Table \ref{tab:morphology_sensitivity}, the results demonstrate a clear dependence of performance on agent sensing capabilities. More capable drones consistently achieve superior monitoring outcomes despite operating at larger stand-off distances. While partially heterogeneous teams exhibit intermediate performance, their results closely track the average capability of their constituent agents and show no evidence of synergistic gains. Notably, the presence of lower-capability agents significantly degrades performance, indicating that team effectiveness is constrained by its weakest members. Moreover, fully heterogeneous teams are not demonstrated to outperform homogeneous configurations. The learned policies are therefore unable to exploit potential complementarities between different sensing parametrizations. Overall, these findings highlight that sensing quality dominates performance and that heterogeneity introduces coordination challenges without clear benefits.
        
        \begin{table*}[h!]
        \centering
        \renewcommand{\arraystretch}{1.25}
        \setlength{\tabcolsep}{5pt}
        
        \caption{Drone sensing sensitivity with three drones observing a single animal. For mixed teams, morphologies are randomly assigned to drones at each episode. Reported values correspond to mean $\pm$ std over these random assignments. Reward varies between -1 and 1. Angular seperation varies between 0 and 1. Mean averaged over 100 episodes.}
        \label{tab:morphology_sensitivity}
        
        \begin{tabular}{l|l|cc|c}
        \hline \hline
        & & \multicolumn{3}{c}{\textbf{Metrics}} \\
        \cline{3-5}
        \textbf{Team type} & \textbf{Morphology}
        & \textbf{Reward} & \textbf{Mean angular sep.} & \textbf{Mean distance (m)} \\
        \hline \hline
        
        \multirow{3}{*}{\shortstack{Fully\\Homogeneous}}
        & D1, D1, D1
        & $0.349 \pm 0.090$
        & $0.701 \pm 0.082$
        & $(\mathrm{D1})\ 113.3 \pm 6.997$ \\
        \cline{5-5}
        
        & D2, D2, D2
        & $0.700 \pm 0.006$
        & $0.827 \pm 0.077$
        & $(\mathrm{D2})\ 145.5 \pm 17.27$ \\
        \cline{5-5}
        
        & D3, D3, D3
        & $0.830 \pm 0.113$
        & $0.818 \pm 0.089$
        & $(\mathrm{D3})\ 185.2 \pm 22.00$ \\
        
        \hline
        \addlinespace[4pt]
        
        \multirow{12}{*}{\shortstack{Partially\\Heterogeneous}}
        & \multirow{2}{*}{D1, D1, D2}
         & \multirow{2}{*}{$0.409 \pm 0.009$}
        & \multirow{2}{*}{$0.824 \pm 0.059$}
        & $(\mathrm{D1})\ 114.1 \pm 5.696$ \\
        \cline{5-5}
        &  &  &  & $(\mathrm{D2})\ 171.2 \pm 12.97$ \\
        \cline{2-5}
        
        & \multirow{2}{*}{D1, D1, D3}
        & \multirow{2}{*}{$0.489 \pm 0.010$}
        & \multirow{2}{*}{$0.626 \pm 0.123$}
        & $(\mathrm{D1})\ 109.6 \pm 7.036$ \\
        \cline{5-5}
        &  &  &  & $(\mathrm{D3})\ 180.3 \pm 16.12$ \\
        \cline{2-5}
        
        & \multirow{2}{*}{D1, D2, D2}
        & \multirow{2}{*}{$0.574 \pm 0.005$}
        & \multirow{2}{*}{$0.871 \pm 0.059$}
        & $(\mathrm{D1})\ 110.2 \pm 7.226$ \\
        \cline{5-5}
        &  &  &  & $(\mathrm{D2})\ 145.0 \pm 16.27$ \\
        \cline{2-5}
        
        & \multirow{2}{*}{D1, D3, D3}
        & \multirow{2}{*}{$0.708 \pm 0.007$}
        & \multirow{2}{*}{$0.773 \pm 0.124$}
        & $(\mathrm{D1})\ 110.3 \pm 6.232$ \\
        \cline{5-5}
        &  &  &  & $(\mathrm{D3})\ 178.5 \pm 18.91$ \\
        \cline{2-5}
        
        & \multirow{2}{*}{D2, D2, D3}
        & \multirow{2}{*}{$0.778 \pm 0.006$}
        & \multirow{2}{*}{$0.861 \pm 0.048$}
        & $(\mathrm{D2})\ 152.0 \pm 8.594$ \\
        \cline{5-5}
        &  &  &  & $(\mathrm{D3})\ 189.2 \pm 12.86$ \\
        \cline{2-5}
        
        & \multirow{2}{*}{D2, D3, D3}
        & \multirow{2}{*}{$0.814 \pm 0.008$}
        & \multirow{2}{*}{$0.818 \pm 0.088$}
        & $(\mathrm{D2})\ 133.9 \pm 13.57$ \\
        \cline{5-5}
        &  &  &  & $(\mathrm{D3})\ 171.9 \pm 16.36$ \\
        
        \hline
        \addlinespace[4pt]
        
        \multirow{3}{*}{\shortstack{Fully\\Heterogeneous}}
        & \multirow{3}{*}{D1, D2, D3}
        & \multirow{3}{*}{$0.626 \pm 0.009$}
        & \multirow{3}{*}{$0.734 \pm 0.048$}
        & $(\mathrm{D1})\ 100.8 \pm 4.596$ \\
        \cline{5-5}
        &  &  &  & $(\mathrm{D2})\ 147.9 \pm 11.27$ \\
        \cline{5-5}
        &  &  &  & $(\mathrm{D3})\ 161.8 \pm 9.393$ \\
        
        \hline \hline
        \end{tabular}
        \end{table*}

\section{Discussion}
\label{sec:discussion}
In this study, we presented a disturbance-aware RL framework for autonomous wildlife monitoring with aerial robots (drones). The proposed framework combines synthetic animal movement models, empirical GPS trajectory replay, a geometry-based disturbance model, aerial sensing agents, and reward shaping that explicitly balances monitoring quality against induced animal disturbance. The aim was to investigate whether RL can produce effective monitoring strategies that retain useful observations while limiting behavioral disturbance.

The disturbance model in this study is intentionally designed to be general-purpose, relying on geometry- and evidence-based behavioural rules rather than species-specific calibration. While empirical evaluation was performed on Pigeons, Spur-winged lapwings, and Jackals, the objective is not to tailor predictions to individual species. Instead, these evaluations demonstrate the framework’s robustness across diverse movement behaviours, supporting its applicability as a broadly generalizable monitoring tool. Under the proposed disturbance formulation, motivated by drone-animal disturbance literature \cite{BrissonCuradeau2025,mulero2017}, there is a meaningful trade-off between disturbance and monitoring, formed by the conflicting nature of the two objectives. Closer positioning generally improves monitoring quality, while simultaneously increasing the risk of disturbance. The trade-off is apparent in the Pareto analysis where monitoring reward increases rapidly with moderate disturbance, but gains decrease as disturbance levels rise.

In synthetic experiments, learning agents show considerably better performance compared to the rule-based baseline as indicated in Table \ref{tab:behavior_results_full}. Evaluation of the learning agents across synthetic behaviours shows a clear trend. SAC achieved the overall strongest performance followed by PPO and DQN as presented in Table \ref{tab:behavior_results_full} and Figure \ref{fig:sac_vs_gps}. This pattern holds across all movement models except POI. The results are expected from SAC and PPO given previous demonstrations of strong performance in continuous spaces \cite{Schulman2017, Haarnoja2018}. Furthermore, given the limited action space of DQN, it was still able to consistently outperform the rule-based baseline. Notably, PPO exhibited highly regular training behavior and was considered the most stable learner. This observation motivated its choice as the basis for multi-agent experiments.

Interestingly, the different learning algorithms did not generally make the same trade-off. DQN tended to prioritize proximity, resulting in strategies that were relatively aggressive, PPO generally exhibited a more conservative strategy and SAC achieved the most favorable trade-off by balancing competing objectives. This pattern is reasonable given the different optimization strategies used by the agents. PPO uses clipped policy updates to limit large policy changes and improve stability \cite{Schulman2017}, whereas SAC uses a maximum-entropy objective encouraging exploration while optimizing expected return in continuous action spaces \cite{Haarnoja2018}. More broadly, the result aligns with multi-objective RL theory, where competing objectives often give rise to a set of valid policy trade-offs rather than a single universally optimal behaviors \cite{roijers2013survey}, as presented by Fig \ref{fig:pareto}.

When considering real-world data, SAC maintained its top-scoring position, with the best observed transfer performance. Interestingly, DQN achieved its strongest transfer performance for Spur-winged lapwings. This may be due to its discretized action space being more densely configured at lower speeds, coupled with the fact that they were the slowest animal with lowest amount of exhibited movement variance. The performance of PPO on synthetic behaviors did not appear to translate to better transfer performance. It achieved subpar performance on pigeons, but still managed to outperform the rule-based baseline. One possible explanation is that PPO’s stable learning on synthetic movement models resulted in less adaptable policies during transfer.

The performance of the policies that were trained on different movement models managed to transfer to empirical trajectories. A trend was observed in which the best-fit movement model obtained the best average total reward on unseen empirical trajectories, conforming to academic literature each animal's general movement patterns. Jackals best-fit was CRW, closely followed by EE, which also is highlighted in the result section. For Pigeons, the best-fitting behavior was LPOI closely followed by POI. Likewise, the best performing average model was POI. Note that the POI and LPOI behaviors express a high degree of similarity. Lastly, for Spur-winged lapwings, the best-fitting animal movement model was EE, which also produced best fitting RL-models in terms of average total reward. In summary, the results suggest that simulation fidelity is sufficiently high to ensure meaningful usability of the framework, where policy performance transfers to unseen animal trajectories. Moreover, training on matching behaviours improves transfer performance when evaluating policies trained with synthetic movement models on empirical GPS data. Since policies trained using more structured movement models still transferred relatively well to less structured ones, it may indicate a one-way relationship. Overall, the findings are consistent with research on domain randomization and robust RL indicating that exposure to more challenging or informative variations of the same problem can result in more generalizable policies \cite{Mehta2020, Greenberg2023, Muratore2022}.

In sensing experiments, wider fields of view and longer view ranges were observed to perform better compared to weaker parameterizations. This is expected since disturbance is naturally distance gated \cite{Schad2023,christie2016unmanned,BrissonCuradeau2025}. Better configurations allowed drones to maintain useful monitoring while staying at a greater distance, effectively shifting the Pareto frontier. Consequently, better hardware likely provides an alternative route to lowering disturbance. Additionally, the results display the importance of considering disturbance in more resource-constrained scenarios which necessitate closer monitoring distance. The evaluation of team composition indicates that heterogeneous teams did not provide a performance benefit compared to homogeneous ones, and that performance generally followed the average of constituent agents. These results could indicate that the learning agent may not be able to utilize any complementary sensing benefits, that there may not be a major benefit to heterogeneous teams, or that larger teams suffer lower scores because of coordination difficulties.

The final line of reasoning is also supported by the multi-target experiments. In a single-target setting there is a slight benefit in adding a second drone, but the same does not apply for a third one. This shows that the benefit of additional drones suffers from diminishing returns. This is reasonable given the additional disturbance induced when more drones are in the vicinity of the animal. It is also a likely indicator of redundancy as the system grows and increasingly shifts towards an assignment or coordination problem. The observed scaling difficulty is consistent with broader challenges in RL, where local observations and shared rewards can be insufficient in larger cooperative systems \cite{Lowe2017,Rashid2020,Oroojlooy2023}. Interpreting these results suggests that future work should consider communication, role assignment, hierarchical control, or centralized training with structured coordination objectives, in order to improve cooperation between agents which could result in better performance for multi-target settings.

Several limitations should be acknowledged in the study. First, comparisons in the framework depend on evidence-based assumptions built into disturbance and reward formulations. As such, the assumptions constrain simulation fidelity and obtained results. Second, while the results are evaluated over identically seeded runs, DQN requires discretization of the action space (a fundamental limitation of the agent). This means that the results are tied to the specific discretization and may vary with another one. Third, with regards to external validity, robustness and sensitivity experimentation reveals several limitations. In particular, obtained policies perform worse under environmental perturbations which are likely further exacerbated under real-world conditions. Forth, since there is a lack of data covering species-specific disturbance factors, the model cannot be easily calibrated. Thus, future work can extend the proposed framework using empirical drone-animal interaction data, more realistic robotic simulation, explicit perception uncertainty, terrain and occlusion modeling, and more structured multi-agent coordination mechanisms. These extensions would support stronger ecological validation and bring the approach closer to real-world deployment in conservation and behavioral research.

Taken jointly, the results show that disturbance-aware monitoring with drones can be automated using learned strategies to established stand-off guidelines. Furthermore, the study demonstrated that transfer from synthetic to GPS trajectories benefits from training on matching movement types, indicating the importance in choosing matched movement models for training policies that transfer to real scenarios. The results also suggest that employing multiple drones for wildlife monitoring may be difficult when considering the additional expected disturbance, however, for smaller teams a slight benefit was still observed.

\section*{Declarations}
\subsection*{Funding}
This study has not obtained any funding.

\subsection*{Conflicts of interest/Competing interests}
None.

\subsection*{Data availability}
The data used in this study is available from the cited sources. 

\subsection*{Code availability}
The study's code is freely available in the following Github repository: \url{https://github.com/mosmar99/Stealth-Fleet}.

\subsection*{Acknowledgments}
The authors with to thank Orr Spiegel for his guidance and for sharing the animal movement data.   

\subsection*{Author Contribution}
Mahmut Osmanovic: Conceptualization, Methodology, Software, Formal analysis, Investigation, Writing - Original Draft, Visualization
Isac Paulsson: Conceptualization, Methodology, Software, Formal analysis, Investigation, Writing - Original Draft, Visualization
Teddy Lazebnik: Methodology, Validation, Investigation, Resources, Data Curation, Writing - Review \& Editing, Supervision. 

\bibliography{biblio}
\bibliographystyle{unsrt}

\appendix

\section{Architecture and Hyperparameters}
\label{app:architecture_hyperparameters}

This appendix provides a detailed overview of the simulation environment, neural network architectures, and training configurations used for the evaluated reinforcement learning agents. While the SAC architecture is illustrated and discussed in the main text, the remaining architectural details and hyperparameter settings are presented here to improve reproducibility and provide additional implementation clarity. The appendix further summarizes the environment design, model structures, discretization schemes, and optimization settings employed across DQN, PPO, SAC, and MAPPO. 

\subsection{Environment architecture}
\label{subapp:environment_architecture}

\begin{figure}[hb]
    \centering
    \includegraphics[width=1\linewidth]{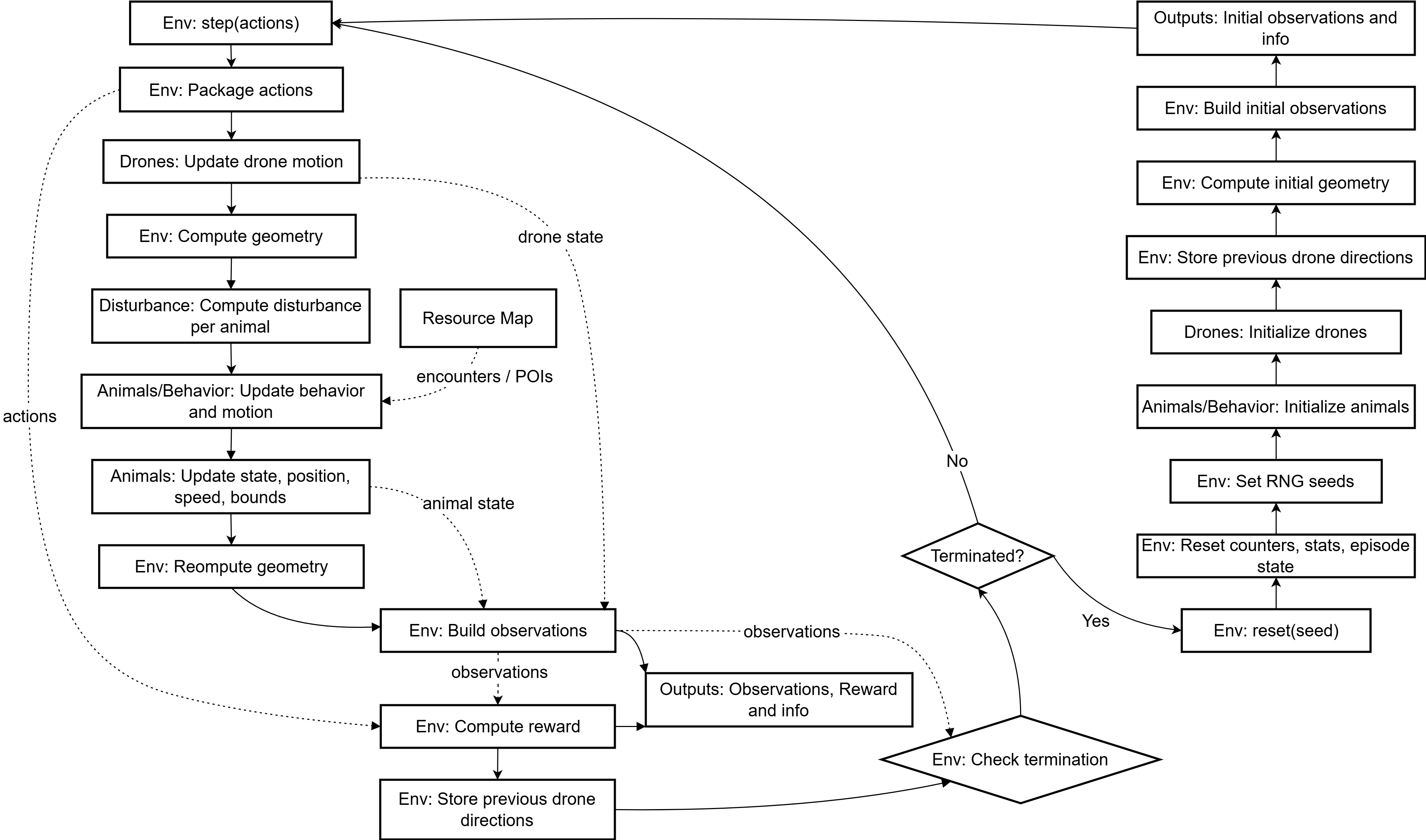}
    \caption{High level environment function flow chart}
    \label{fig:env_diagram}
\end{figure}

Figure \ref{fig:env_diagram} details the high level architecture of the simulation environment. It describes both procedures and information flow pertaining to the step and reset functions provided by the environment.

\subsection{Model architectures}
\label{subapp:model_architectures}

The evaluated RL-methods differ in how policies are represented and optimized. DQN operates on a discretized action space using a value-based formulation, while PPO and SAC learn stochastic policies directly in continuous action spaces. MAPPO extends PPO to the multi-agent setting through centralized training with decentralized execution. All models share a common backbone of fully connected networks with two hidden layers and LeakyReLU activations. Table~\ref{tab:algorithm_architectures} summarizes the main neural network architectures used by the evaluated RL-methods. The purpose of the Table is to provide a compact overview of how each method represents policies, value functions, and action selection. This complements the methodological description in the main text by specifying the structural differences between value-based, actor-critic, and multi-agent actor-critic models.

\begin{table}[ht]
\centering
\caption{Core architecture of the evaluated RL-models.}
\label{tab:algorithm_architectures}
\begin{tabular}{p{3.0cm}p{8cm}}
\hline \hline
\textbf{Model} & \textbf{Architecture} \\
\hline \hline
DQN & Branching dueling Q-network with a shared two-layer MLP trunk $(256,256)$ using LeakyReLU activations. Separate branches are used for direction, speed, and camera yaw actions. A scalar value head and branch-wise advantage heads are combined using a dueling Q formulation, enabling factorized action selection in a discretized continuous space. \\
\midrule
PPO & Actor--critic architecture with separate actor and critic MLPs. Both networks use two hidden layers $(256,256)$ with LeakyReLU activations. The actor outputs the mean of a Gaussian policy with a learned log-standard deviation, followed by tanh action squashing for bounded continuous control. \\
\midrule
SAC & Off-policy actor--critic architecture with a stochastic Gaussian actor, twin Q-functions, and corresponding target critics. Networks use two hidden layers $(256,256)$ with LeakyReLU activations. Actions are tanh-squashed, with log-probability correction applied during optimization. \\
\midrule
MAPPO & Multi-agent extension of PPO with decentralized actors and a centralized critic. Each actor receives local observations, while the critic operates on the concatenated global observation across agents. Both actor and critic use two hidden layers $(256,256)$ with LeakyReLU activations. \\
\hline \hline
\end{tabular}
\end{table}

\subsection{Hyperparameters}
\label{subapp:hyperparameters}

The hyperparameters were selected to ensure stable and comparable training across all methods. Default values from prior literature were used where applicable, with minor empirical adjustments to improve convergence stability across different environments. 

The main training hyperparameters are reported in Table~\ref{tab:rl_hyperparameters}. These settings define the optimization procedure, batch sizes, discounting, learning rates, exploration schedules, and update frequencies used during training. Reporting these values supports reproducibility and makes the comparison between learning algorithms more transparent.

\begin{table}[h]
\centering
\caption{Main hyperparameters used for DQN, PPO, SAC, and MAPPO training.}
\label{tab:rl_hyperparameters}
\renewcommand{\arraystretch}{1.32}

\begin{tabular}{lc}
\hline\hline
\multicolumn{2}{c}{\textbf{DQN-specific parameters}} \\
\hline
\textbf{Parameter} & \textbf{DQN} \\
\hline
Replay buffer & $200{,}000$ \\
Learn after & $10{,}000$ \\
Learn every & $10$ \\
Target update & $16{,}384$ \\
Soft update $\tau$ & $0.005$ \\
Exploration & $1.0 \rightarrow 0.05$ \\
Decay steps & $500{,}000$ \\
\hline\hline
\end{tabular}

\vspace{0.6em}

\begin{tabular}{lccc}
\hline\hline
\multicolumn{4}{c}{\textbf{Policy-gradient / actor--critic parameters}} \\
\hline
\textbf{Parameter} & \textbf{PPO} & \textbf{SAC} & \textbf{MAPPO} \\
\hline
Critic LR & $3\times10^{-4}$ & $3\times10^{-4}$ & $3\times10^{-4}$ \\
GAE $\lambda$ & $0.95$ & $0.95$ & $0.95$ \\
Policy clip & $0.2$ & $0.2$ & $0.2$ \\
Value loss coef. & $0.5$ & $0.5$ & $0.5$ \\
Entropy coef. & $0.02 \rightarrow 0.001$ & $0.02 \rightarrow 0.001$ & $0.02 \rightarrow 0.002$ \\
Rollout steps & $8192$ & $8192$ & $2048$ \\
Epochs/update & $4$ & $4$ & $10$ \\
\hline\hline
\end{tabular}

\vspace{0.6em}

\begin{tabular}{lcccc}
\hline\hline
\multicolumn{5}{c}{\textbf{General training parameters}} \\
\hline
\textbf{Parameter} & \textbf{DQN} & \textbf{PPO} & \textbf{SAC} & \textbf{MAPPO} \\
\hline
Total timesteps & $2{,}000{,}000$ & $2{,}000{,}000$ & $800{,}000$ & $3{,}000{,}000$ \\
Mini-batch size & $256$ & $256$ & $256$ & $256$ \\
Discount factor $\gamma$ & $0.99$ & $0.99$ & $0.99$ & $0.99$ \\
Actor / Policy LR & $3\times10^{-4}$ & $3\times10^{-4}$ & $3\times10^{-4}$ & $3\times10^{-4}$ \\
\hline\hline
\end{tabular}
\end{table}

Because DQN requires a discrete action space, the continuous drone control commands were mapped to a finite set of action branches. The direction branch used a fixed set of normalized movement vectors, while the speed and camera-yaw commands were discretized into scalar bins:
\[
v \in \{-1.00,-0.98,-0.96,-0.92,-0.88,-0.84,-0.76,-0.68,-0.56,-0.30,0.10,0.50,1.00\},
\]
\[
\theta \in \{-1,-0.5,-0.25,0,0.25,0.5,1\}.
\]

This factorized discretization enables tractable learning while retaining expressive control over the original continuous action space.

\subsection{Network Architectures}
\label{subapp:network_architectures}

\begin{figure}
    \centering
    \includegraphics[width=0.80\linewidth]{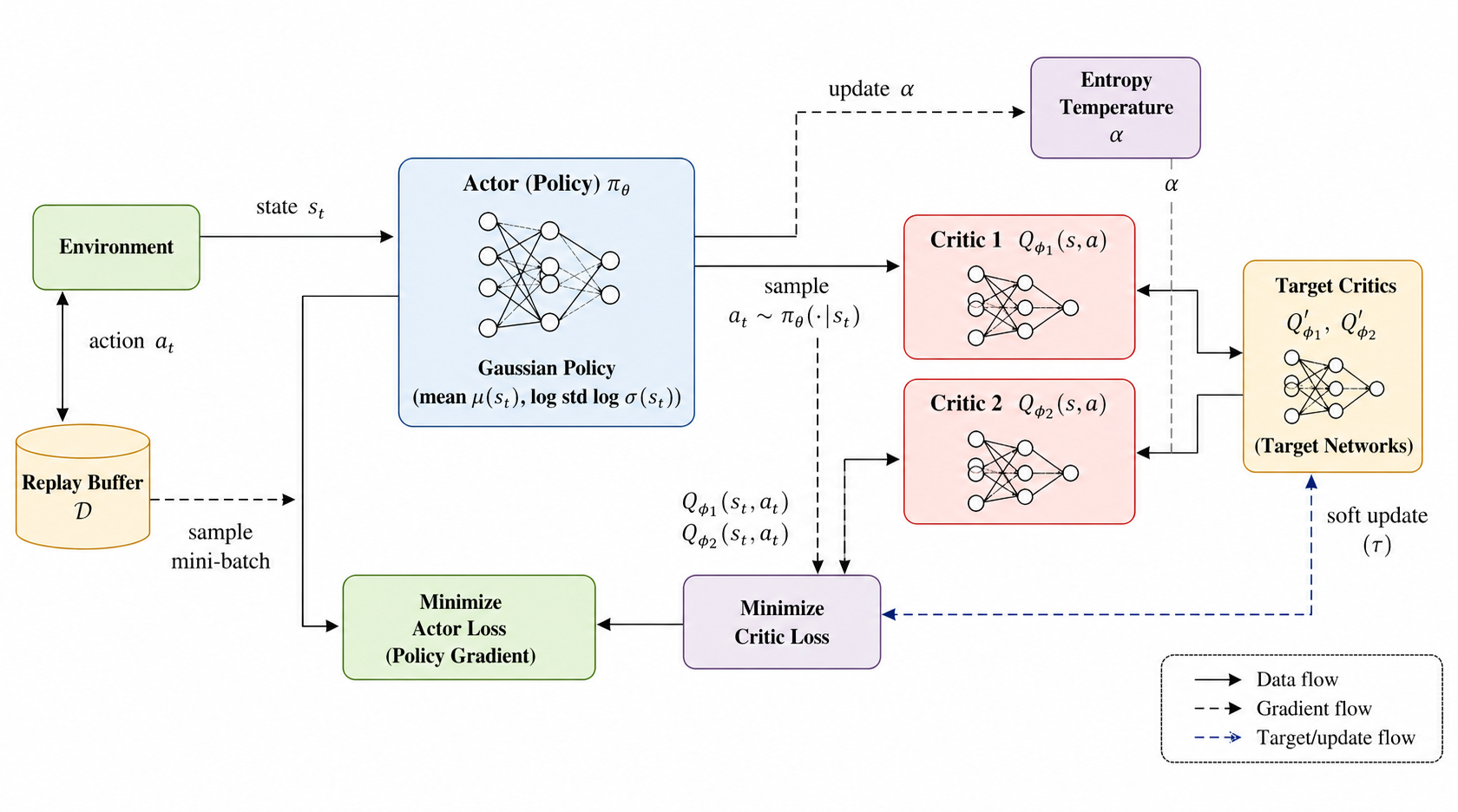}
    \caption{SAC architecture. The diagram illustrates the interaction between the stochastic actor, twin Q-function critics, and target networks. Experience is stored in a replay buffer and used for off-policy updates. The actor outputs a Gaussian policy with tanh-squashed actions, while the critics are trained using a Bellman objective with entropy regularization.}
    \label{fig:sac}
\end{figure}

\begin{figure}
    \centering
    \includegraphics[width=0.80\linewidth]{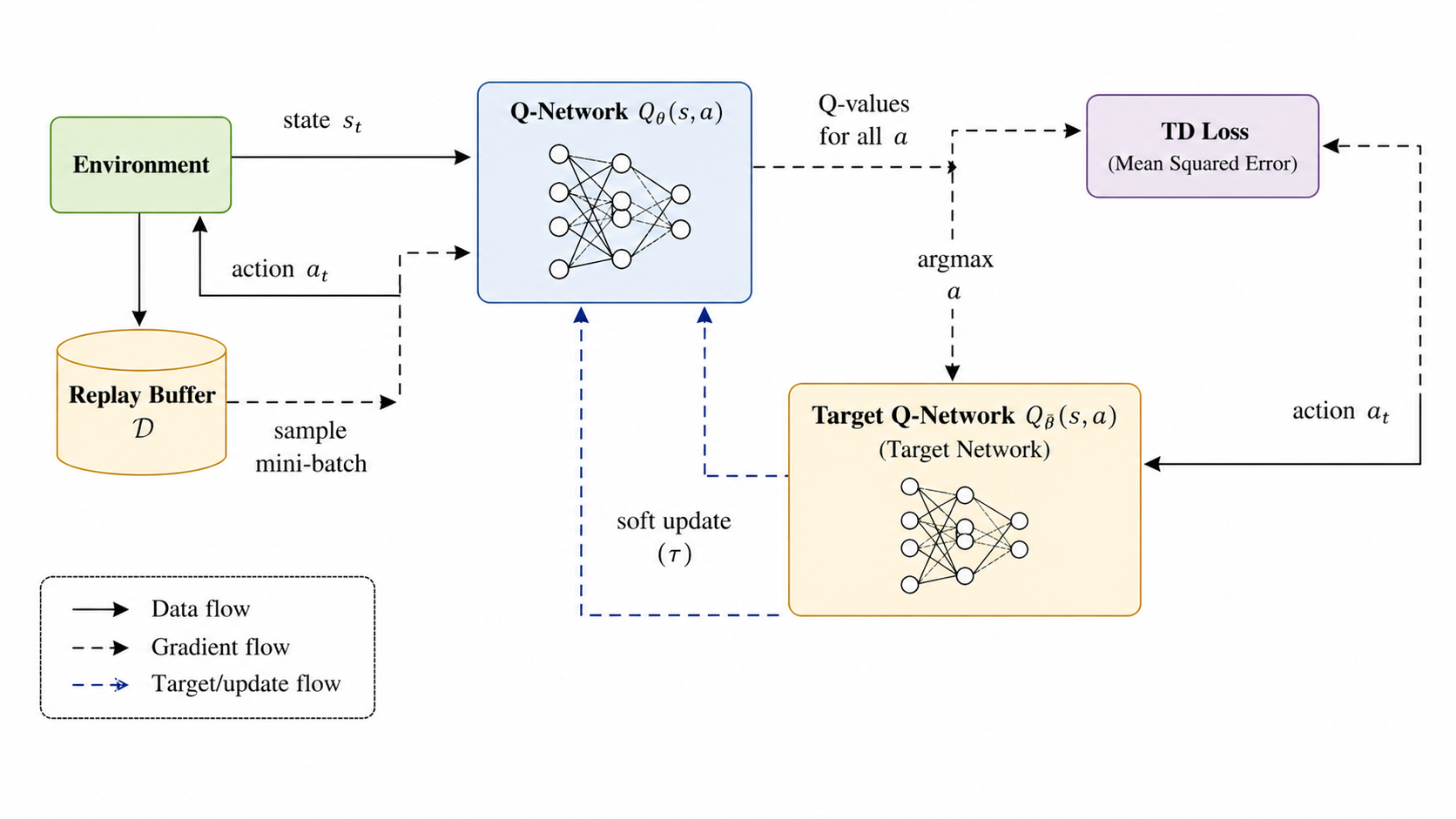}
    \caption{DQN architecture, two-layer MLPs with LeakyReLU activations. Experience is stored in a replay buffer and used for off-policy updates.}
    \label{fig:dqn}
\end{figure}

\begin{figure}
    \centering
    \includegraphics[width=0.80\linewidth]{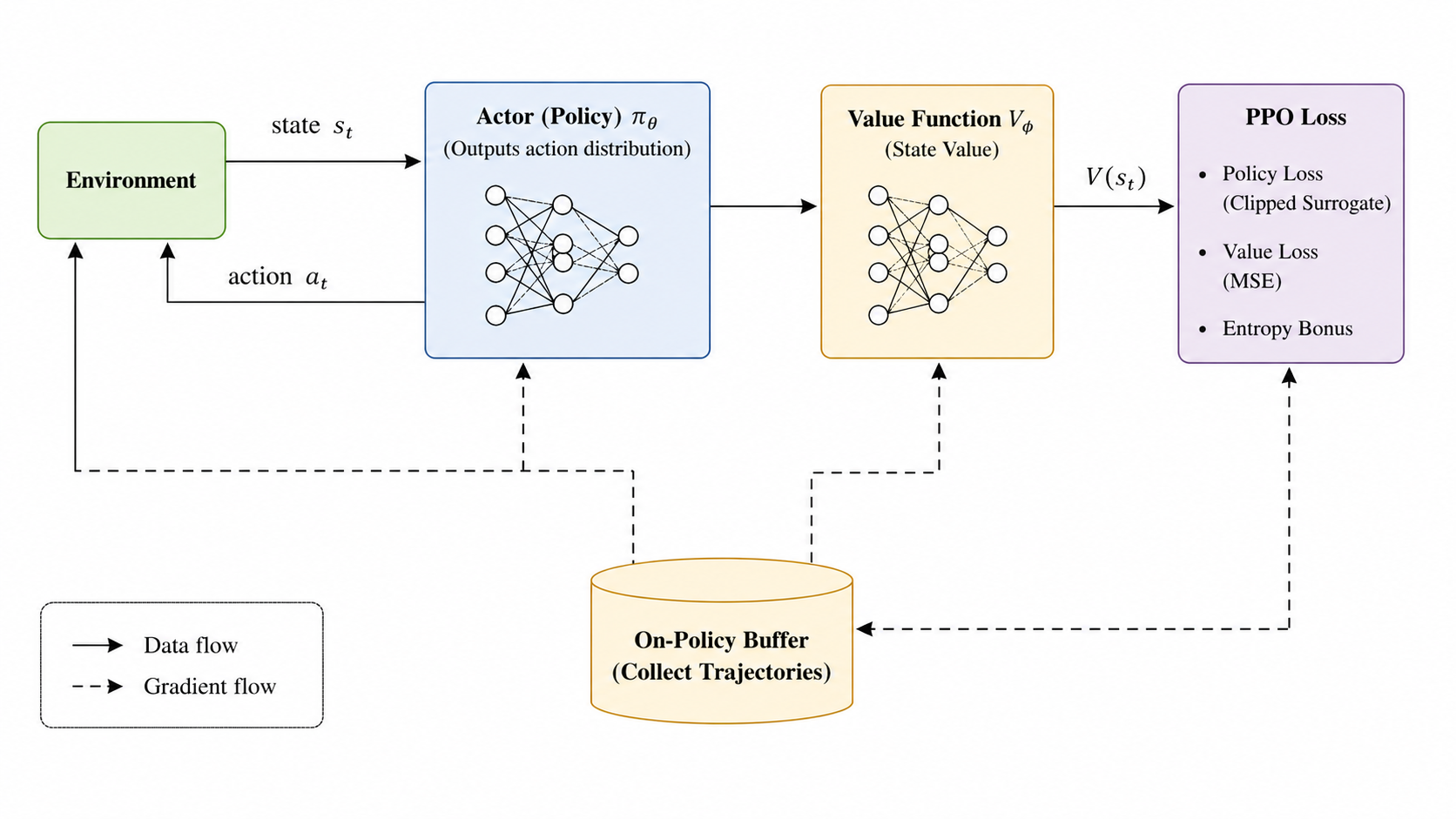}
    \caption{PPO architecture, two-layer MLPs with LeakyReLU activations.}
    \label{fig:ppo}
\end{figure}

To complement the tabular description, Figures \ref{fig:sac}, \ref{fig:dqn}, and \ref{fig:ppo} illustrate the neural network structures used by the evaluated models. The diagrams highlight the structural differences between value-based and actor--critic methods, including action branching in DQN and stochastic policy representations in PPO and SAC.

\newpage

\section{Transfer and Behavioral Alignment}
\label{app:transfer_alignment}

Table \ref{tab:behavior_results_selected_animals} reports the transfer performance for the best performing movement type, selected by the highest mean total reward. Across all three animals, the learned models outperform the rule-based strategy in total reward. 

\begin{table*}[ht]
\centering
\setlength{\tabcolsep}{4pt}
\caption{Summarized transfer results. Best movement prior per model and animal, selected among CRW, EE, POI, and LPOI (unfitted parameters) using mean total reward. Results are reported as mean $\pm$ standard deviation across 100 evaluation episodes. Drones are of D2 type sensing capability and no wind is present in the environment. Policy evaluations are made on sampled real GPS data.}
\label{tab:behavior_results_selected_animals}
\begin{tabular}{llccc}
\hline \hline
\textbf{Model} & \textbf{Metric} & \textbf{Jackals} & \textbf{Pigeons} & \textbf{Spur-winged lapwings} \\
\hline \hline
Rule-based & Total reward & $0.849 \pm 0.013$ & $0.815 \pm 0.038$ & $0.845 \pm 0.033$ \\
& Distance reward & $0.501 \pm 0.001$ & $0.483 \pm 0.019$ & $0.498 \pm 0.007$ \\
& Disturbance penalty & $0.188 \pm 0.006$ & $0.192 \pm 0.010$ & $0.189 \pm 0.014$ \\
\midrule
DQN & Total reward & $0.914 \pm 0.008$ (CRW) & $0.883 \pm 0.040$ (CRW) & $0.919 \pm 0.013$ (CRW) \\
& Distance reward & $0.572 \pm 0.015$ (CRW) & $0.575 \pm 0.037$ (CRW) & $0.534 \pm 0.015$ (CRW) \\
& Disturbance penalty & $0.202 \pm 0.013$ (CRW) & $0.219 \pm 0.040$ (CRW) & $0.173 \pm 0.015$ (CRW) \\
\midrule
PPO & Total reward & $0.910 \pm 0.072$ (EE) & $0.831 \pm 0.163$ (POI) & $0.917 \pm 0.030$ (LPOI) \\
& Distance reward & $0.542 \pm 0.026$ (EE) & $0.517 \pm 0.067$ (POI) & $0.513 \pm 0.032$ (LPOI) \\
& Disturbance penalty & $0.176 \pm 0.020$ (EE) & $0.186 \pm 0.056$ (POI) & $0.151 \pm 0.011$ (LPOI) \\
\midrule
SAC & Total reward & $0.929 \pm 0.007$ (EE) & $0.925 \pm 0.015$ (LPOI) & $0.933 \pm 0.018$ (LPOI) \\
& Distance reward & $0.585 \pm 0.009$ (EE) & $0.553 \pm 0.018$ (LPOI) & $0.534 \pm 0.009$ (LPOI) \\
& Disturbance penalty & $0.202 \pm 0.009$ (EE) & $0.179 \pm 0.017$ (LPOI) & $0.161 \pm 0.010$ (LPOI) \\
\hline \hline
\end{tabular}
\end{table*}

\newpage

\begin{table*}[!ht]
\centering
\setlength{\tabcolsep}{4pt}
\caption{Detailed transfer results. Performance comparison across animals and behavior priors (unfitted) for each animal, movement type, and agent. Results are reported as mean $\pm$ standard deviation across 100 evaluation episodes. Here, $R_{\text{tot}}$ denotes total reward, $R_{\text{dist}}$ denotes distance reward, and $P_{\text{disturb}}$ denotes disturbance penalty.}
\label{tab:behavior_results_full_animals}
\begin{tabular*}{\textwidth}{@{\extracolsep{\fill}}p{1.0cm}cp{1.3cm}cccc}
\hline \hline
\textbf{Animal} & \textbf{Model} & \textbf{Metric} & \textbf{CRW} & \textbf{EE} & \textbf{POI} & \textbf{LPOI} \\
\hline \hline

\multirow{12}{*}{Jackals}
& \multirow{3}{*}{Rule-based} & $R_{\text{tot}}$ & $0.849 \pm 0.013$ & $0.849 \pm 0.013$ & $0.849 \pm 0.013$ & $0.849 \pm 0.013$ \\
& & $R_{\text{dist}}$ & $0.501 \pm 0.001$ & $0.501 \pm 0.001$ & $0.501 \pm 0.001$ & $0.501 \pm 0.001$ \\
& & $P_{\text{disturb}}$ & $0.188 \pm 0.006$ & $0.188 \pm 0.006$ & $0.188 \pm 0.006$ & $0.188 \pm 0.006$ \\
\cmidrule(lr){2-7}
& \multirow{3}{*}{DQN} & $R_{\text{tot}}$ & $0.914 \pm 0.008$ & $0.893 \pm 0.009$ & $0.901 \pm 0.010$ & $0.888 \pm 0.009$ \\
& & $R_{\text{dist}}$ & $0.572 \pm 0.015$ & $0.588 \pm 0.020$ & $0.558 \pm 0.014$ & $0.509 \pm 0.014$ \\
& & $P_{\text{disturb}}$ & $0.202 \pm 0.013$ & $0.217 \pm 0.018$ & $0.188 \pm 0.009$ & $0.153 \pm 0.009$ \\
\cmidrule(lr){2-7}
& \multirow{3}{*}{PPO} & $R_{\text{tot}}$ & $0.908 \pm 0.007$ & $0.910 \pm 0.072$ & $0.906 \pm 0.006$ & $0.909 \pm 0.007$ \\
& & $R_{\text{dist}}$ & $0.553 \pm 0.015$ & $0.542 \pm 0.026$ & $0.544 \pm 0.020$ & $0.537 \pm 0.011$ \\
& & $P_{\text{disturb}}$ & $0.189 \pm 0.012$ & $0.176 \pm 0.020$ & $0.181 \pm 0.016$ & $0.174 \pm 0.009$ \\
\cmidrule(lr){2-7}
& \multirow{3}{*}{SAC} & $R_{\text{tot}}$ & $0.921 \pm 0.004$ & $0.929 \pm 0.007$ & $0.927 \pm 0.006$ & $0.923 \pm 0.012$ \\
& & $R_{\text{dist}}$ & $0.546 \pm 0.014$ & $0.585 \pm 0.009$ & $0.556 \pm 0.010$ & $0.549 \pm 0.017$ \\
& & $P_{\text{disturb}}$ & $0.174 \pm 0.010$ & $0.202 \pm 0.009$ & $0.179 \pm 0.008$ & $0.178 \pm 0.010$ \\

\midrule

\multirow{12}{*}{Pigeons}
& \multirow{3}{*}{Rule-based} & $R_{\text{tot}}$ & $0.815 \pm 0.038$ & $0.815 \pm 0.038$ & $0.815 \pm 0.038$ & $0.815 \pm 0.038$ \\
& & $R_{\text{dist}}$ & $0.483 \pm 0.019$ & $0.483 \pm 0.019$ & $0.483 \pm 0.019$ & $0.483 \pm 0.019$ \\
& & $P_{\text{disturb}}$ & $0.192 \pm 0.010$ & $0.192 \pm 0.010$ & $0.192 \pm 0.010$ & $0.192 \pm 0.010$ \\
\cmidrule(lr){2-7}
& \multirow{3}{*}{DQN} & $R_{\text{tot}}$ & $0.883 \pm 0.040$ & $0.673 \pm 0.321$ & $0.852 \pm 0.156$ & $0.857 \pm 0.026$ \\
& & $R_{\text{dist}}$ & $0.575 \pm 0.037$ & $0.531 \pm 0.097$ & $0.535 \pm 0.051$ & $0.507 \pm 0.040$ \\
& & $P_{\text{disturb}}$ & $0.219 \pm 0.040$ & $0.208 \pm 0.044$ & $0.180 \pm 0.026$ & $0.168 \pm 0.037$ \\
\cmidrule(lr){2-7}
& \multirow{3}{*}{PPO} & $R_{\text{tot}}$ & $0.791 \pm 0.223$ & $0.822 \pm 0.187$ & $0.831 \pm 0.163$ & $0.773 \pm 0.234$ \\
& & $R_{\text{dist}}$ & $0.503 \pm 0.062$ & $0.520 \pm 0.053$ & $0.517 \pm 0.067$ & $0.458 \pm 0.086$ \\
& & $P_{\text{disturb}}$ & $0.171 \pm 0.020$ & $0.182 \pm 0.023$ & $0.186 \pm 0.056$ & $0.143 \pm 0.026$ \\
\cmidrule(lr){2-7}
& \multirow{3}{*}{SAC} & $R_{\text{tot}}$ & $0.739 \pm 0.292$ & $0.896 \pm 0.108$ & $0.894 \pm 0.126$ & $0.925 \pm 0.015$ \\
& & $R_{\text{dist}}$ & $0.475 \pm 0.061$ & $0.567 \pm 0.021$ & $0.546 \pm 0.036$ & $0.553 \pm 0.018$ \\
& & $P_{\text{disturb}}$ & $0.142 \pm 0.018$ & $0.194 \pm 0.016$ & $0.177 \pm 0.019$ & $0.179 \pm 0.017$ \\

\midrule

\multirow{12}{*}{\shortstack{Spur-\\winged\\lapwings}}
& \multirow{3}{*}{Rule-based} & $R_{\text{tot}}$ & $0.845 \pm 0.033$ & $0.845 \pm 0.033$ & $0.845 \pm 0.033$ & $0.845 \pm 0.033$ \\
& & $R_{\text{dist}}$ & $0.498 \pm 0.007$ & $0.498 \pm 0.007$ & $0.498 \pm 0.007$ & $0.498 \pm 0.007$ \\
& & $P_{\text{disturb}}$ & $0.189 \pm 0.014$ & $0.189 \pm 0.014$ & $0.189 \pm 0.014$ & $0.189 \pm 0.014$ \\
\cmidrule(lr){2-7}
& \multirow{3}{*}{DQN} & $R_{\text{tot}}$ & $0.919 \pm 0.013$ & $0.917 \pm 0.015$ & $0.897 \pm 0.037$ & $0.877 \pm 0.017$ \\
& & $R_{\text{dist}}$ & $0.534 \pm 0.015$ & $0.542 \pm 0.021$ & $0.570 \pm 0.034$ & $0.461 \pm 0.020$ \\
& & $P_{\text{disturb}}$ & $0.173 \pm 0.015$ & $0.177 \pm 0.020$ & $0.201 \pm 0.022$ & $0.125 \pm 0.014$ \\
\cmidrule(lr){2-7}
& \multirow{3}{*}{PPO} & $R_{\text{tot}}$ & $0.916 \pm 0.010$ & $0.901 \pm 0.012$ & $0.890 \pm 0.008$ & $0.917 \pm 0.030$ \\
& & $R_{\text{dist}}$ & $0.549 \pm 0.018$ & $0.554 \pm 0.017$ & $0.513 \pm 0.024$ & $0.513 \pm 0.032$ \\
& & $P_{\text{disturb}}$ & $0.181 \pm 0.011$ & $0.194 \pm 0.012$ & $0.166 \pm 0.019$ & $0.151 \pm 0.011$ \\
\cmidrule(lr){2-7}
& \multirow{3}{*}{SAC} & $R_{\text{tot}}$ & $0.910 \pm 0.007$ & $0.932 \pm 0.007$ & $0.926 \pm 0.006$ & $0.933 \pm 0.018$ \\
& & $R_{\text{dist}}$ & $0.500 \pm 0.012$ & $0.557 \pm 0.010$ & $0.542 \pm 0.014$ & $0.534 \pm 0.009$ \\
& & $P_{\text{disturb}}$ & $0.147 \pm 0.007$ & $0.174 \pm 0.008$ & $0.166 \pm 0.009$ & $0.161 \pm 0.010$ \\
\hline \hline
\end{tabular*}
\end{table*}

\begin{table*}[ht]
\centering
\renewcommand{\arraystretch}{1.35}
\renewcommand\theadfont{\bfseries}
\renewcommand\theadalign{cc}
\setlength{\tabcolsep}{6pt}
\caption{Comparison between behavioral fit and downstream learned performance across species. The columns under \emph{Behavior fit} report how well each synthetic movement prior matches the corresponding animal trajectory statistics. The final column reports the average learned transfer performance, computed as the mean total reward across the learned agents (DQN, PPO, and SAC) trained with that movement prior.}
\label{tab:model_fit_all_species}

\begin{tabular}{l|l|ccccc|c}
\hline \hline
\rule{0pt}{4.2ex}
&
&
\multicolumn{5}{c|}{\textbf{Behavior fit}}
&
\makecell[c]{\textbf{Learned}\\[2pt]\textbf{performance}} \\
\cline{3-8}
\rule{0pt}{4.2ex}
\textbf{Species}
&
\textbf{Model}
&
\textbf{Speed}
&
\textbf{Turn}
&
\textbf{Tortuosity}
&
\textbf{Revisit}
&
\textbf{Mean}
&
\makecell[c]{\textbf{Avg. total}\\[2pt]\textbf{reward}} \\
\hline \hline

\multirow{4}{*}{Jackals}
& CRW  & 0.612 & 0.105 & 0.238 & 0.056 & \textbf{0.253} & \textbf{0.914} \\
& EE   & 0.617 & 0.167 & 0.267 & 0.223 & 0.318 & 0.911 \\
& POI  & 0.544 & 0.350 & 0.367 & 0.012 & 0.318 & 0.911 \\
& LPOI & 0.572 & 0.319 & 0.331 & 0.115 & 0.334 & 0.907 \\

\hline

\multirow{4}{*}{Pigeons}
& CRW  & 0.359 & 0.121 & 0.107 & 0.393 & 0.245 & 0.804 \\
& EE   & 0.247 & 0.075 & 0.077 & 0.660 & 0.265 & 0.797 \\
& POI  & 0.348 & 0.127 & 0.091 & 0.111 & 0.169 & \textbf{0.859} \\
& LPOI & 0.370 & 0.096 & 0.091 & 0.003 & \textbf{0.140} & 0.852 \\

\hline

\multirow{4}{*}{\makecell[c]{Spur-winged\\lapwings}}
& CRW  & 0.552 & 0.149 & 0.199 & 0.069 & 0.242 & 0.915 \\
& EE   & 0.535 & 0.156 & 0.209 & 0.015 & \textbf{0.229} & \textbf{0.917} \\
& POI  & 0.665 & 0.384 & 3.697 & 0.540 & 1.322 & 0.904 \\
& LPOI & 0.678 & 0.381 & 3.695 & 0.555 & 1.327 & 0.909 \\

\hline \hline
\end{tabular}
\end{table*}

Table \ref{tab:model_fit_all_species} presents a comparison between the four movement models (CRW, EE, POI, and LPOI). The comparisons were made utilizing distributional fit scores for speed, turning angle, tortuosity, and revisitation. Lower values indicate better agreement with the actual GPS trajectories. The mean score summarizes overall fit across metrics. This is compared against the average total reward in transfer for learned models. 

The behavioral fitting results demonstrate that Pigeons are best described by LPOI, Jackals are best modeled through the use of the CRW movement-model and Spur-winged lapwing by EE. This aligns with previous studies about these animals' movement \cite{pigeon1,pigeon2,pigeon3,pigeon4,jackal1,jackal2,jackal3,lapwing1}. Notably, the best average total reward for both Jackals and Spur-winged lapwings is achieved when using the best fitting behavior for training. For Pigeons the best average total reward is achieved using POI, however both POI and LPOI generate good transfer results.

Following the previous two tables, Table \ref{tab:behavior_results_full_animals} presents the performance comparison of the baseline and proposed models across the same three real-world animal cases, where each RL result is reported using the behavioral model selected from the synthetic-data analysis in Table \ref{tab:model_fit_all_species} and indicated in parentheses. The results show that the transition from synthetic behavioral fitting to real-animal evaluation preserves the main advantage of learning-based control, with both PPO and SAC outperforming the rule-based baselines on all three animals. 

\section{Reward Interpretation}
\label{app:reward_interpreataion}

\begin{figure}[!ht]
\centering

\includegraphics[width=1\textwidth]{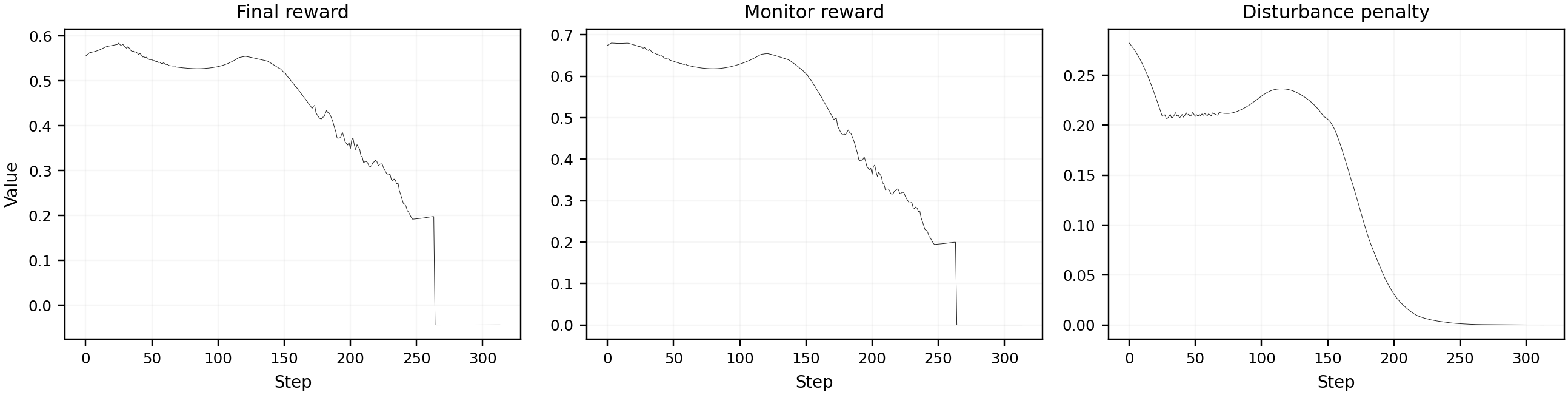}
\vspace{2mm}

\includegraphics[width=1\textwidth]{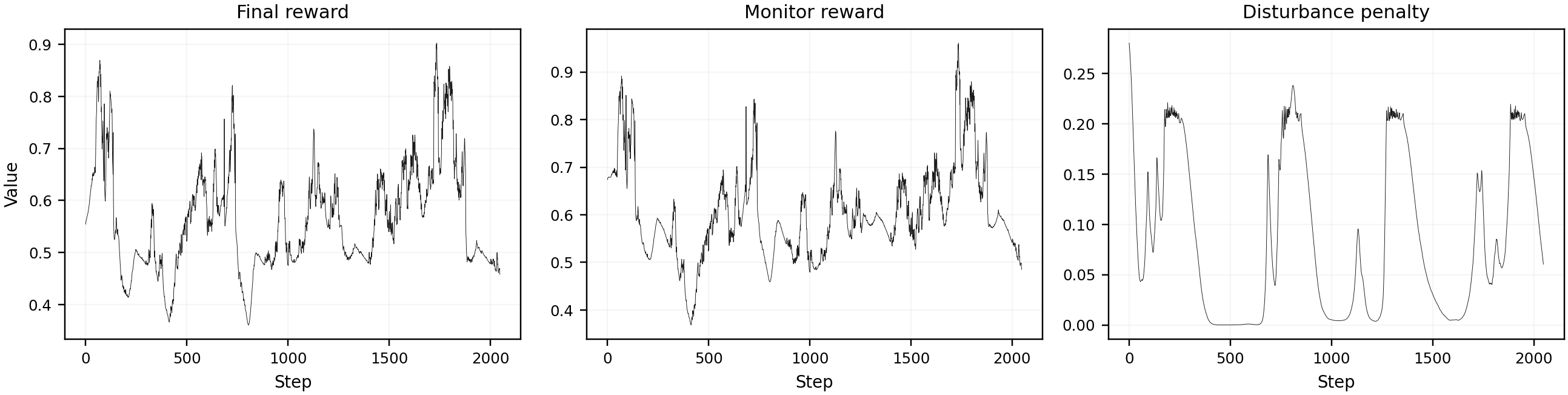}
\vspace{2mm}

\includegraphics[width=1\textwidth]{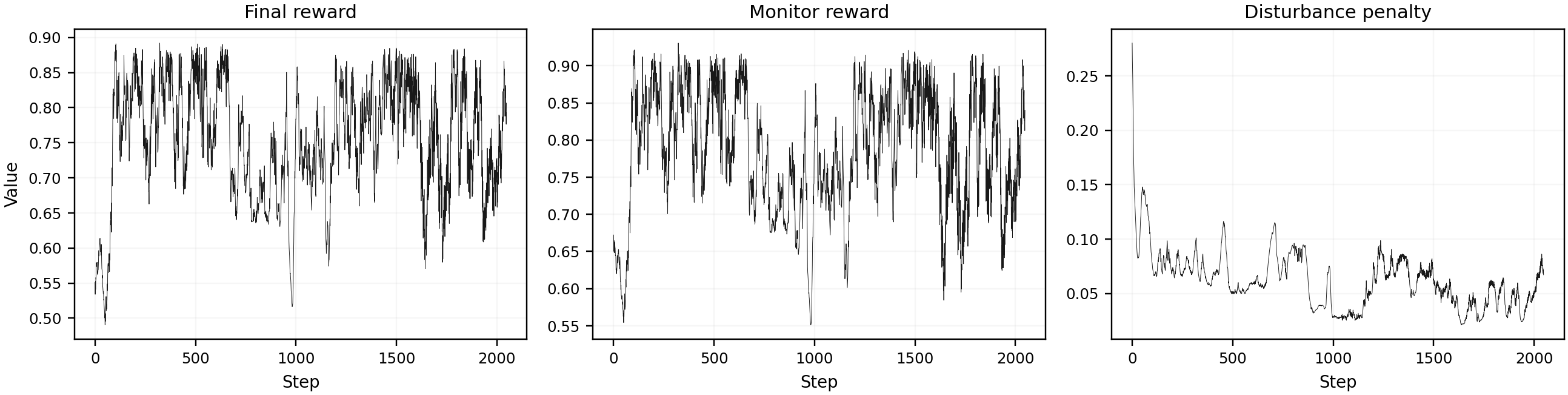}
\vspace{3mm}

\begin{minipage}{0.245\textwidth}
    \centering
    \includegraphics[width=\linewidth]{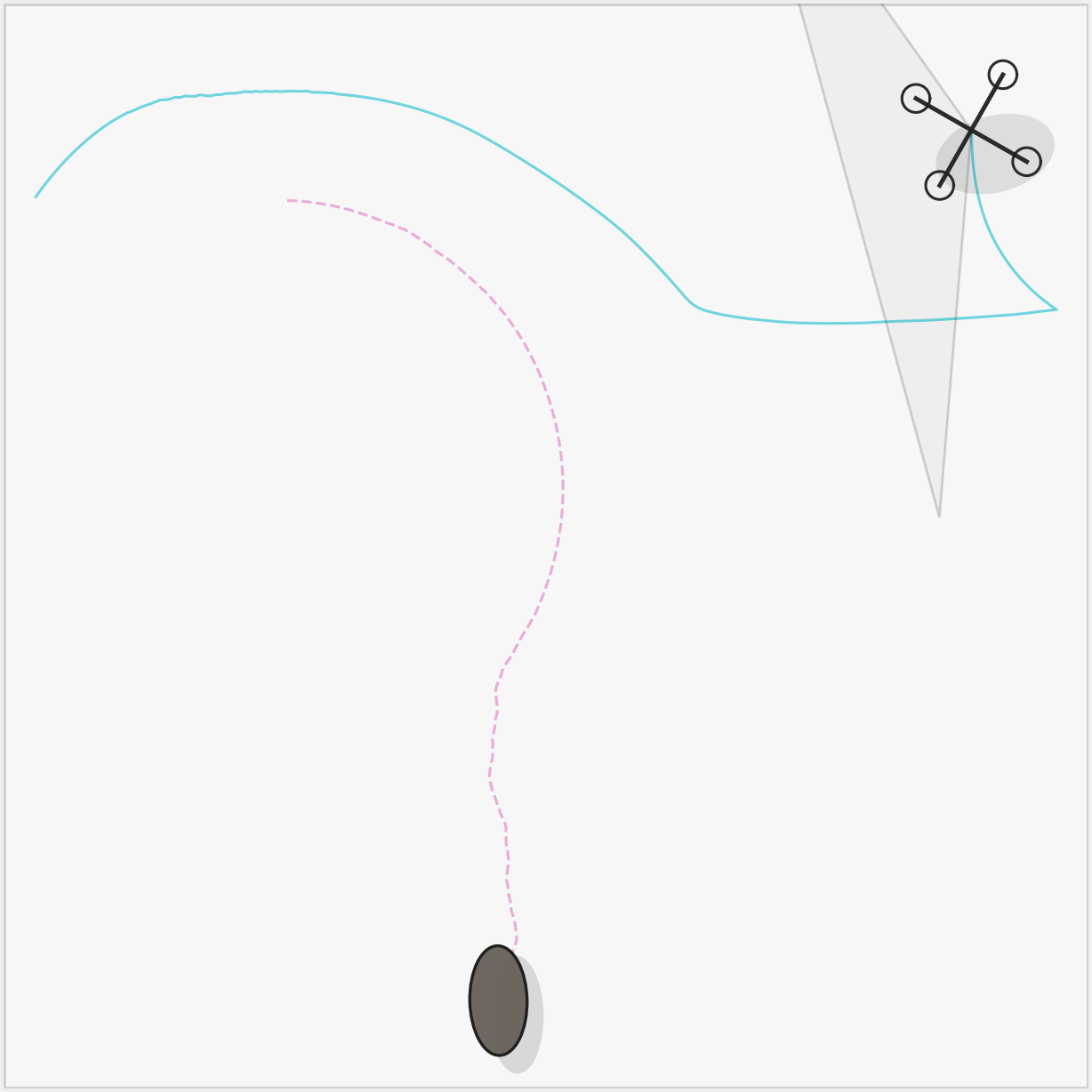}
\end{minipage}
\hfill
\begin{minipage}{0.245\textwidth}
    \centering
    \includegraphics[width=\linewidth]{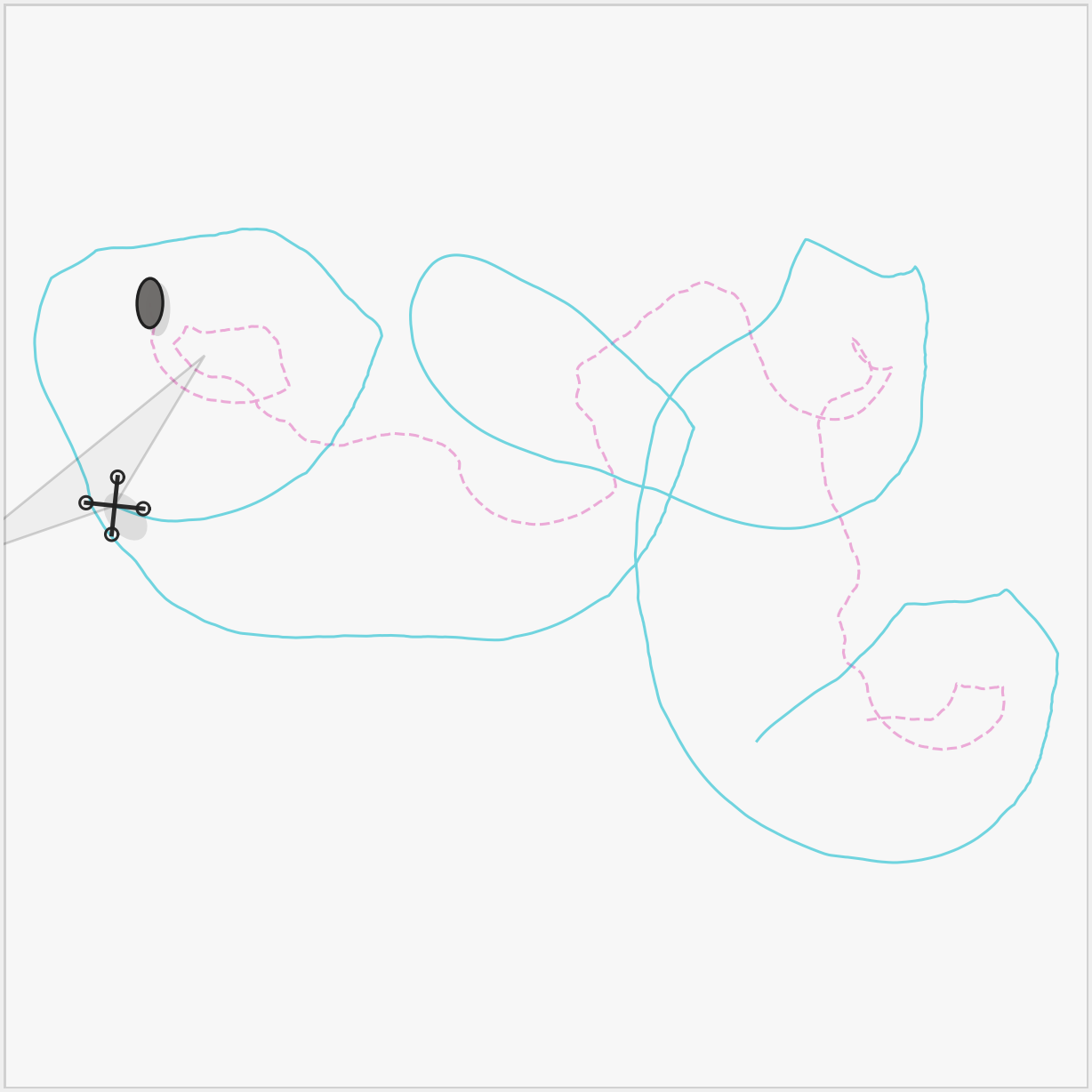}
\end{minipage}
\hfill
\begin{minipage}{0.245\textwidth}
    \centering
    \includegraphics[width=\linewidth]{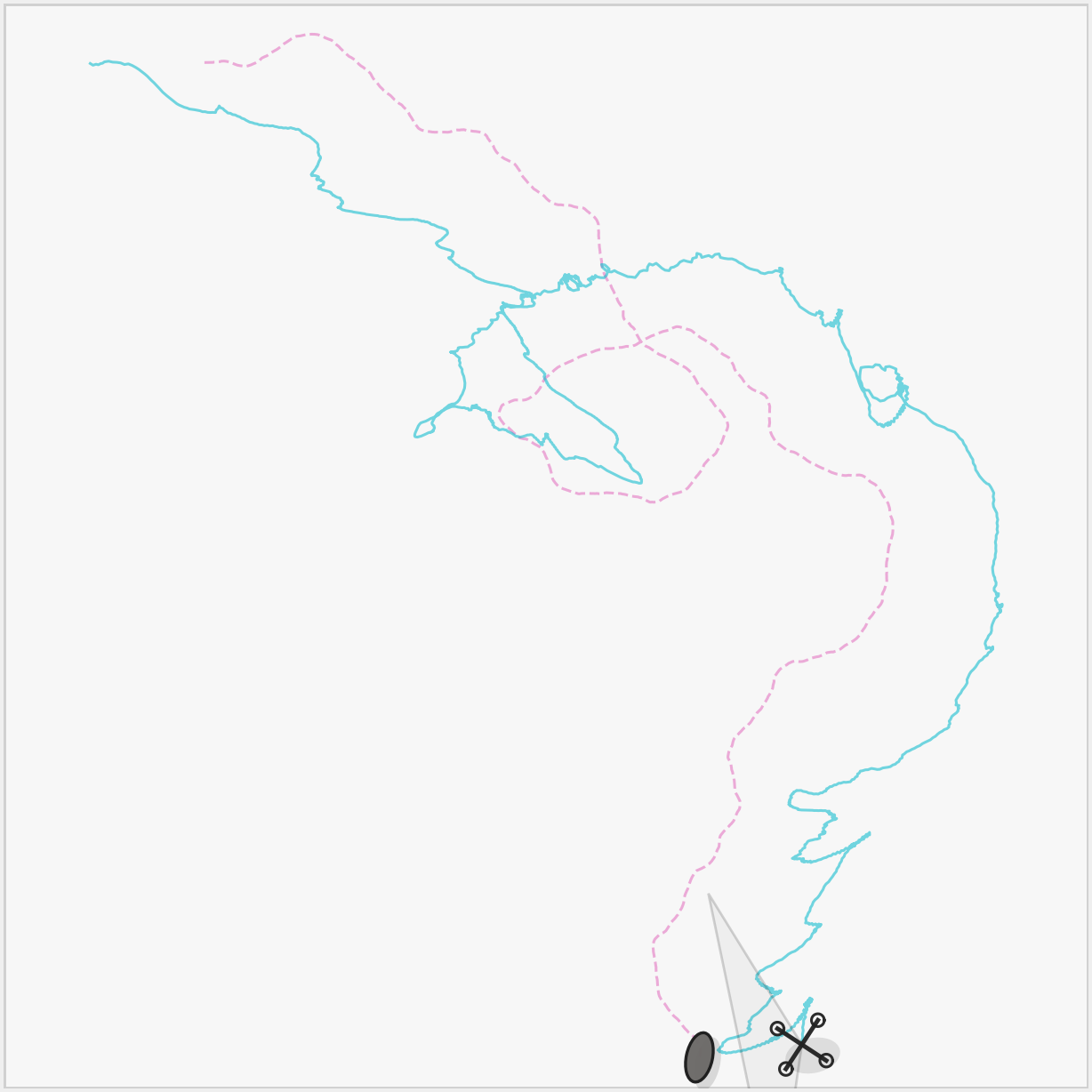}
\end{minipage}

\caption{
Final obtained reward (using PPO) in each of the cases is approximately 0.06, 0.50 and 0.75. At 0.06 in final reward, the agent quickly loses track of the animal. At 0.50, the animal is decently tracked however monitoring is often too close and disturbs the animal, altering its behavior. Finally, at around 0.75 the drone manages to accurately monitor the animal whilst causing negligible disturbance. Specifically, compare the radii of the animal circles, and quantity of them. They occur more often and are more sharp in the turn when the incurred drone disturbance to the animal is great. Contrast this with the smooth and enlarged circle in the rightmost image.
}
\label{fig:four_images_row}

\end{figure}

Figure~\ref{fig:four_images_row} illustrates how different final reward levels correspond to qualitatively different monitoring behaviors. Low reward values correspond to unsuccessful monitoring behavior, where the drone either loses track of the animal or causes excessive disturbance. Intermediate rewards indicate partial monitoring success but with noticeable behavioral impact on the animal, whereas high rewards correspond to stable long-term tracking with minimal disturbance.

\newpage

\section{Training Dynamics}
\subsection{Training dynamics}

\begin{figure}[ht]
        \centering
        
        \includegraphics[width=0.98\linewidth]{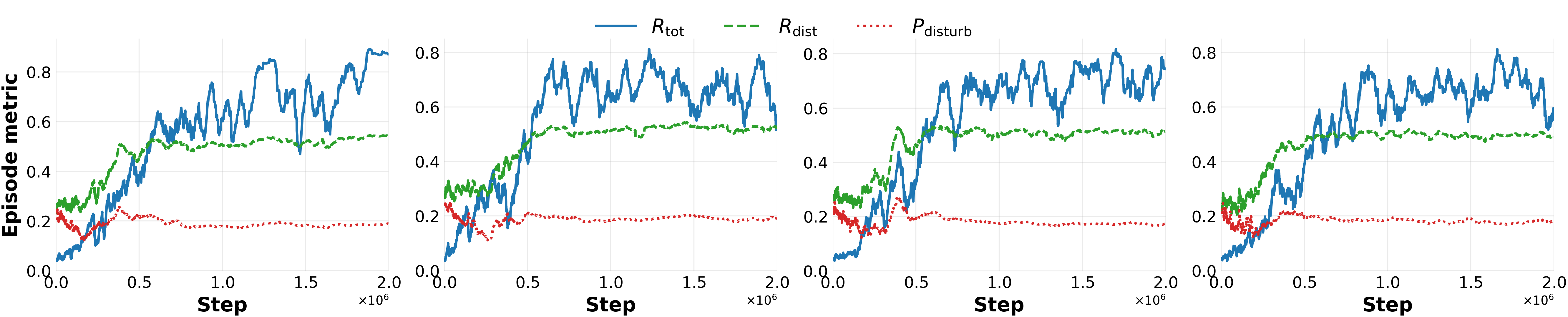}
        
        \vspace{10pt}
        
        \includegraphics[width=0.98\linewidth]{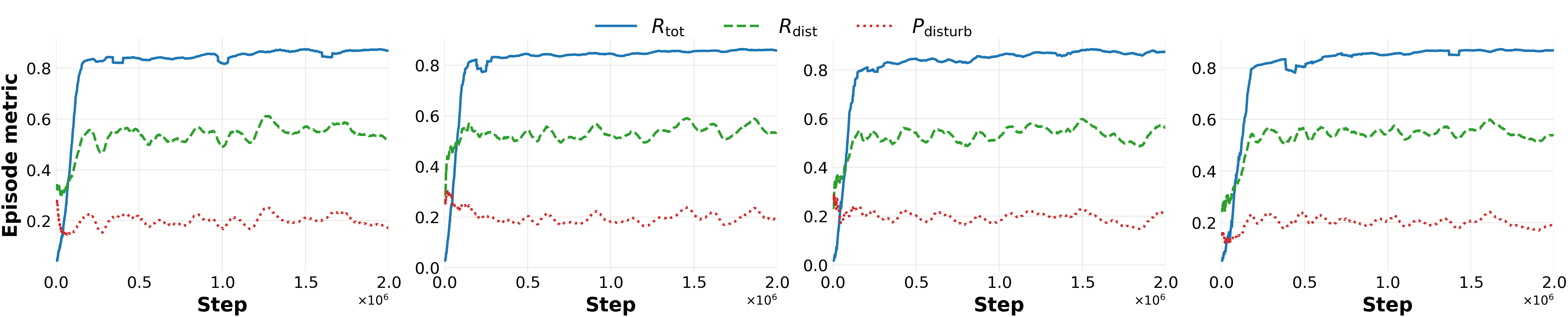}
        
        \vspace{10pt}
        \includegraphics[width=0.98\linewidth]{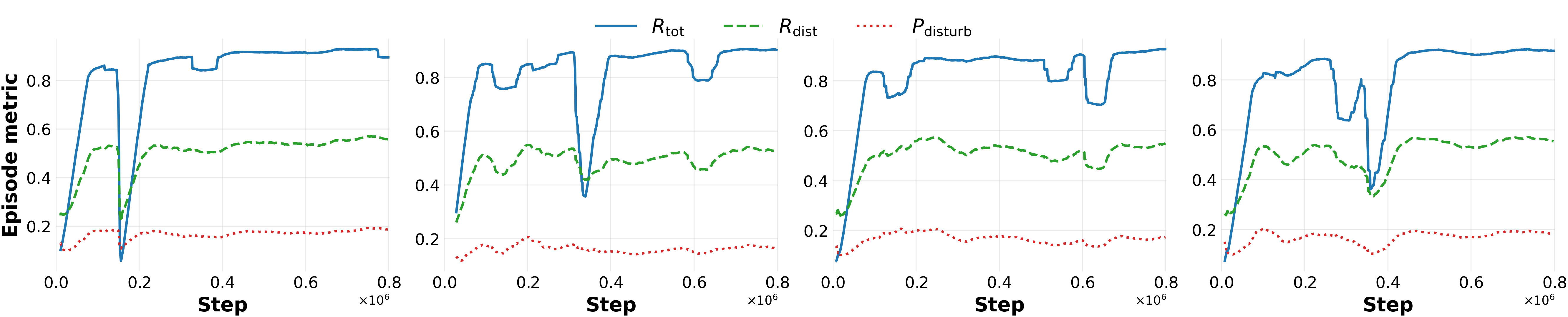}
        
    \caption{
    Training dynamics across learning algorithms. Each row shows the evolution of total reward ($R_{\text{tot}}$), distance reward ($R_{\text{dist}}$), and disturbance penalty ($P_{\text{disturb}}$) over training steps for DQN (top), PPO (middle), and SAC (bottom). Within each row, plots from left to right correspond to training on different animal movement models: CRW, EE, POI, and LPOI. DQN exhibits high variance and unstable learning with frequent oscillations. PPO shows smooth and stable convergence with low variance. SAC achieves high performance with generally stable learning but occasional transient drops, reflecting a balance between exploration and exploitation.
    }
        
    \label{fig:training_comparison}
    \end{figure}

    Figure~\ref{fig:training_comparison} specifies the training dynamics for DQN, PPO and SAC across the evaluated datasets. All models were trained for two million time steps. Clear differences in stability and convergence behaviour can be observed. DQN in particular exhibits high variance and frequent oscillations with respect to the total reward, which is indicative of unstable learning and sensitivity to the training distribution. In contrast, PPO demonstrates smooth and monotonic convergence with minimal variance, consistently reaching a stable performance regime early in training. SAC achieves the highest overall performance but displays occasional transient drops in reward, reflecting periods of instability likely associated with its more exploratory learning dynamics.
    
\newpage

\section{Policy Analysis}
\label{app:policy_analysis}

\begin{figure*}[!ht]
    \centering

    \begin{subfigure}[t]{0.24\textwidth}
        \centering
        \includegraphics[width=\linewidth]{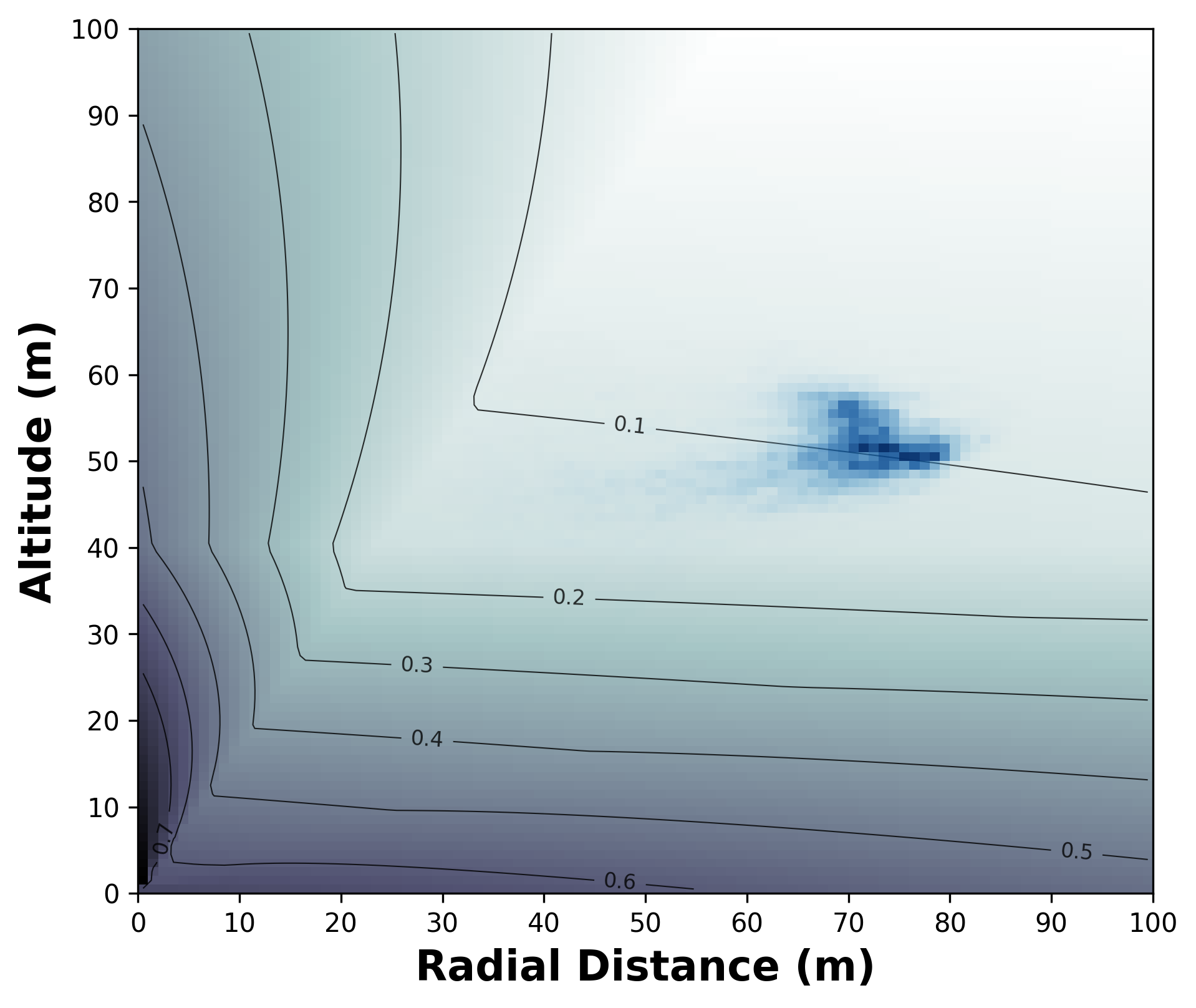}
        \label{fig:synthetic_cont_crw_dqn}
    \end{subfigure}
    \hfill
    \begin{subfigure}[t]{0.24\textwidth}
        \centering
        \includegraphics[width=\linewidth]{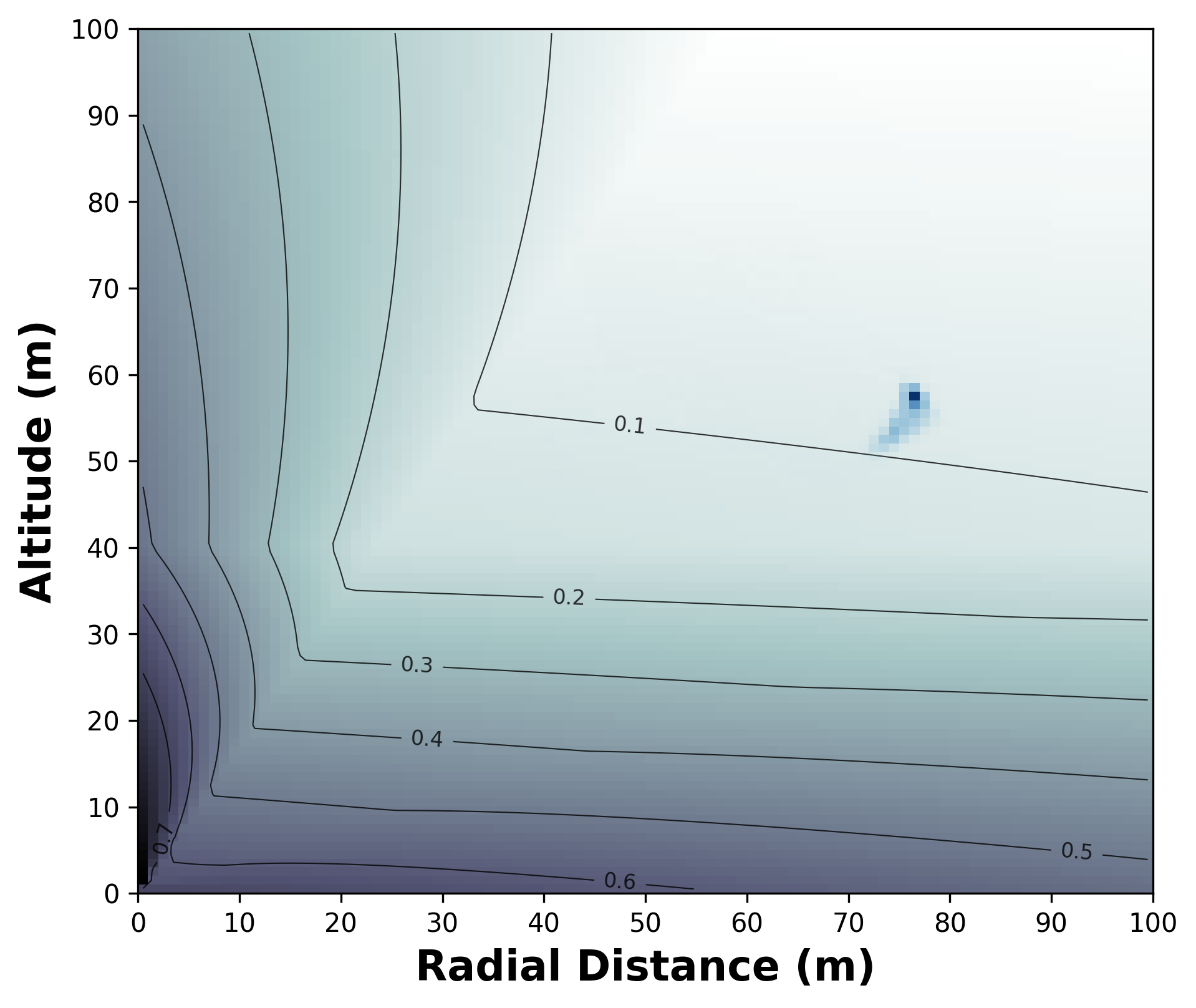}
        \label{fig:synthetic_cont_ee_dqn}
    \end{subfigure}
    \hfill
    \begin{subfigure}[t]{0.24\textwidth}
        \centering
        \includegraphics[width=\linewidth]{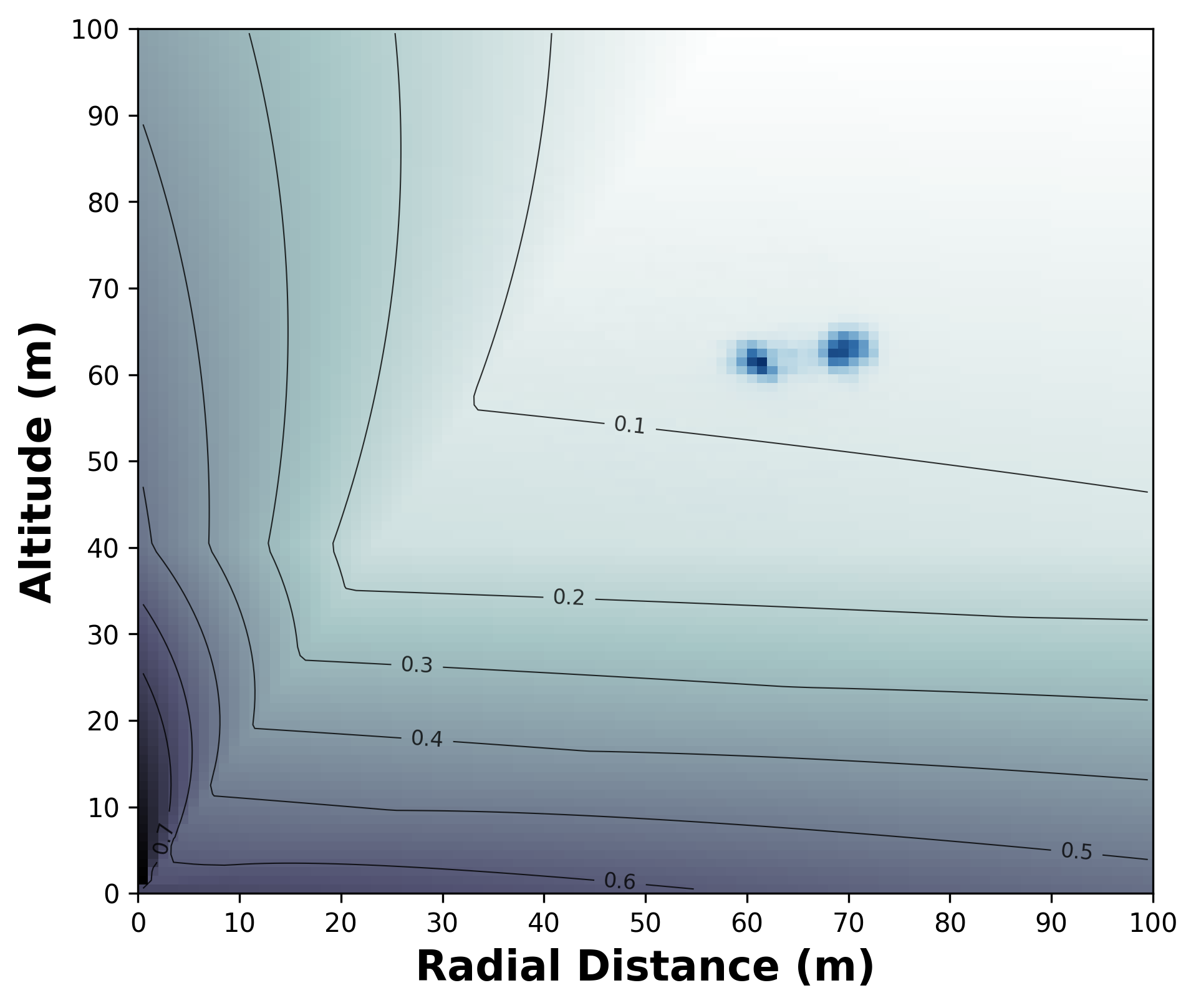}
        \label{fig:synthetic_cont_poi_dqn}
    \end{subfigure}
    \hfill
    \begin{subfigure}[t]{0.24\textwidth}
        \centering
        \includegraphics[width=\linewidth]{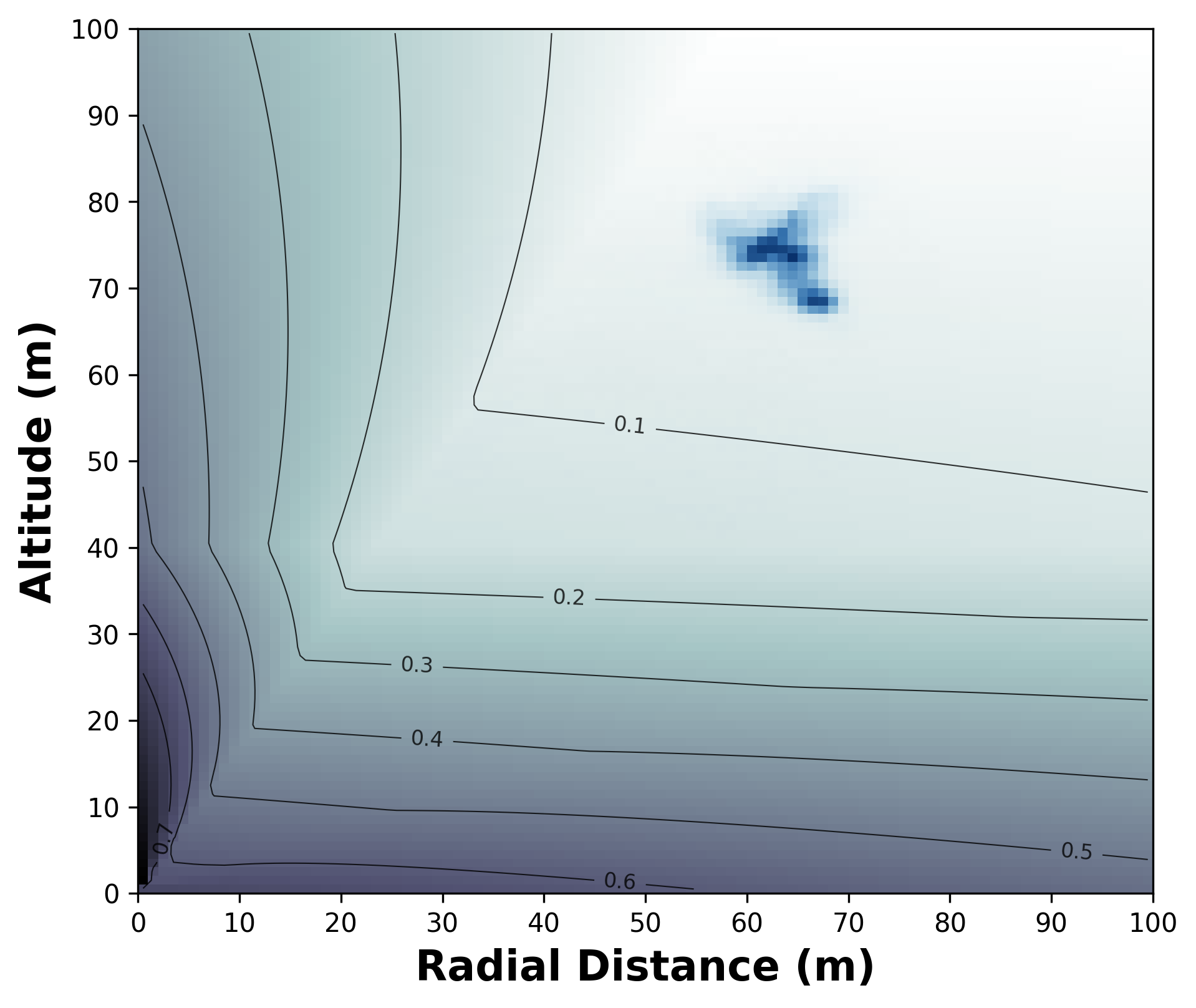}
        \label{fig:synthetic_cont_lpoi_dqn}
    \end{subfigure}

    \vspace{0.4cm}

    \begin{subfigure}[t]{0.24\textwidth}
        \centering
        \includegraphics[width=\linewidth]{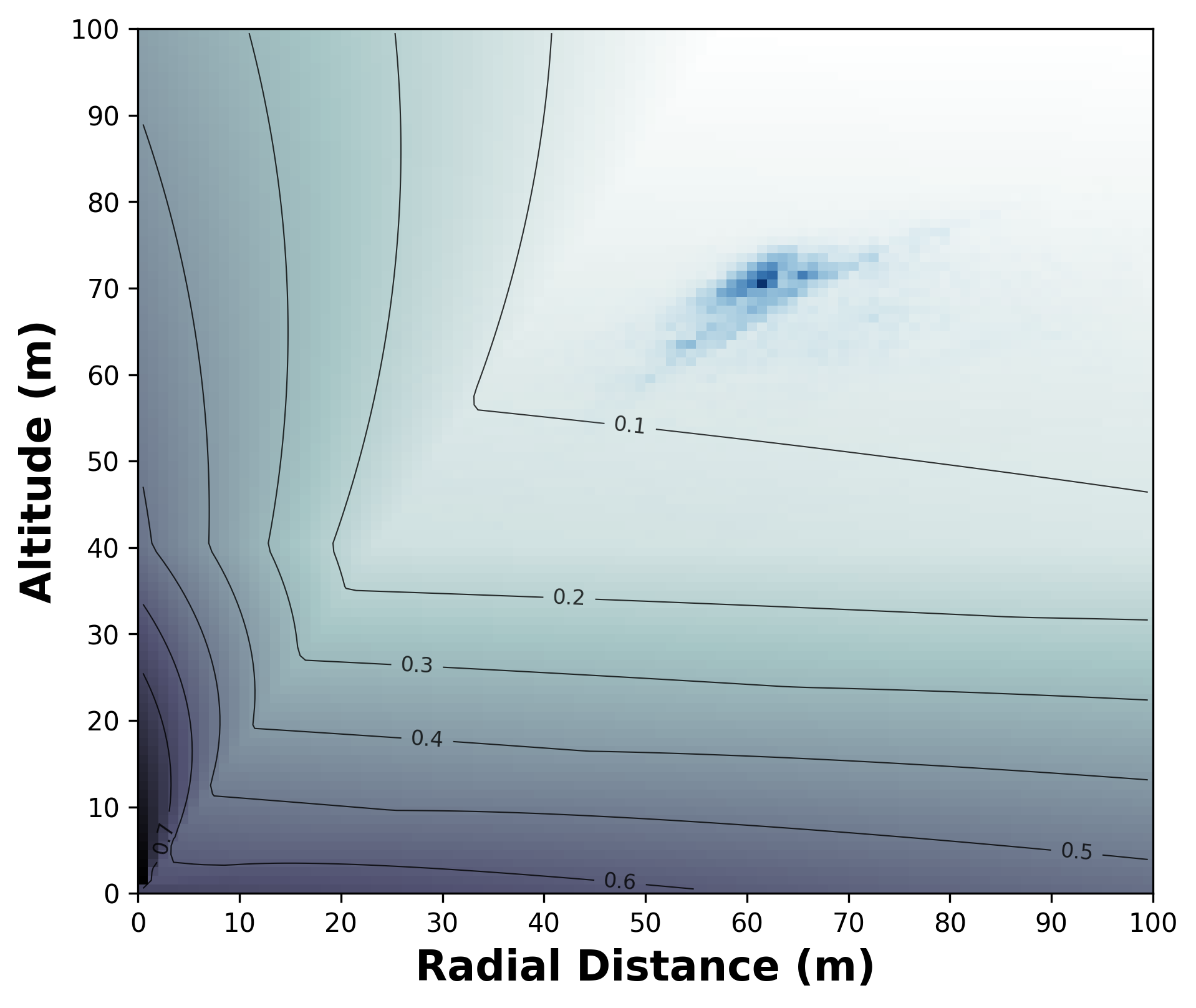}
        \label{fig:synthetic_cont_crw_ppo}
    \end{subfigure}
    \hfill
    \begin{subfigure}[t]{0.24\textwidth}
        \centering
        \includegraphics[width=\linewidth]{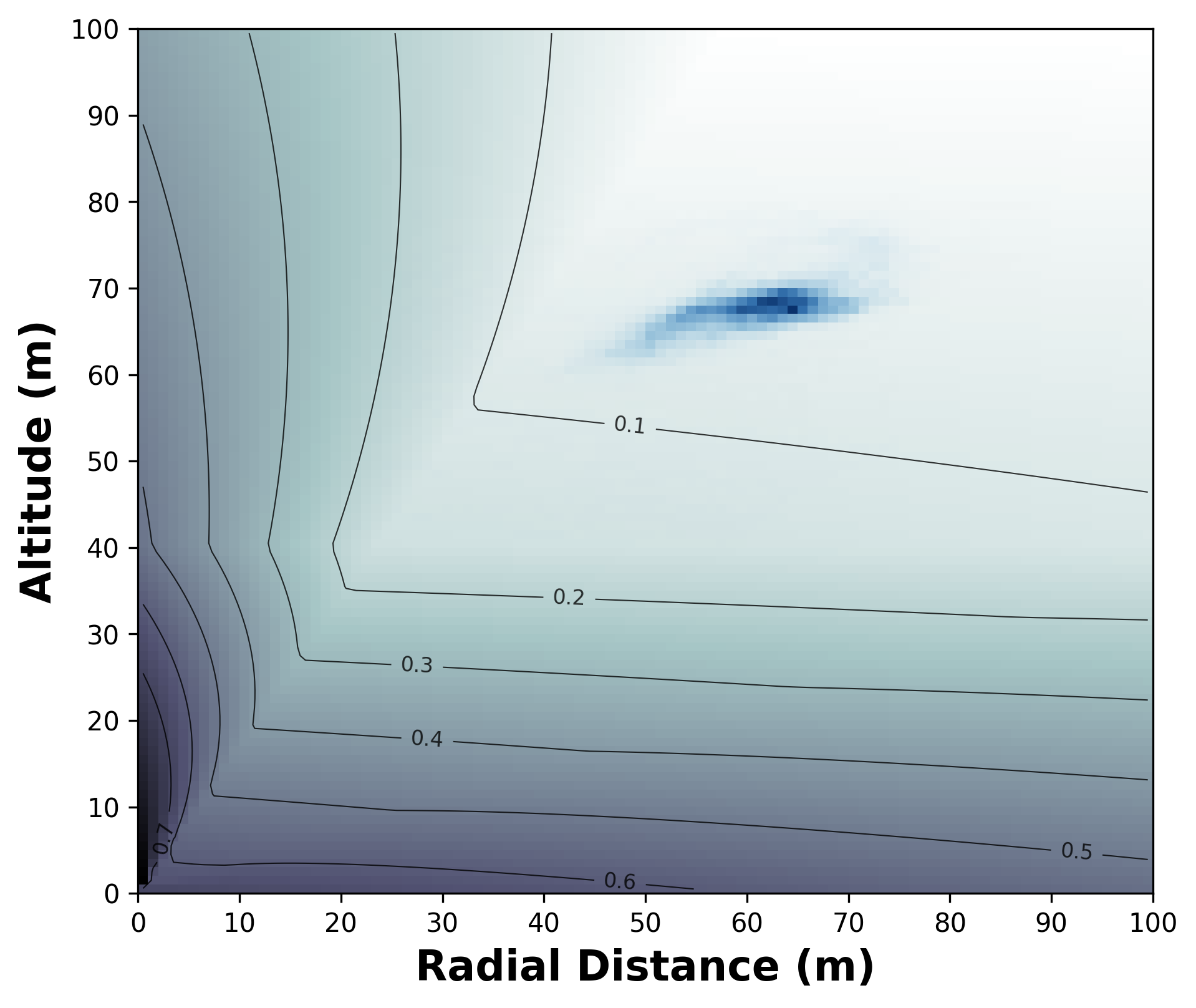}
        \label{fig:synthetic_cont_ee_ppo}
    \end{subfigure}
    \hfill
    \begin{subfigure}[t]{0.24\textwidth}
        \centering
        \includegraphics[width=\linewidth]{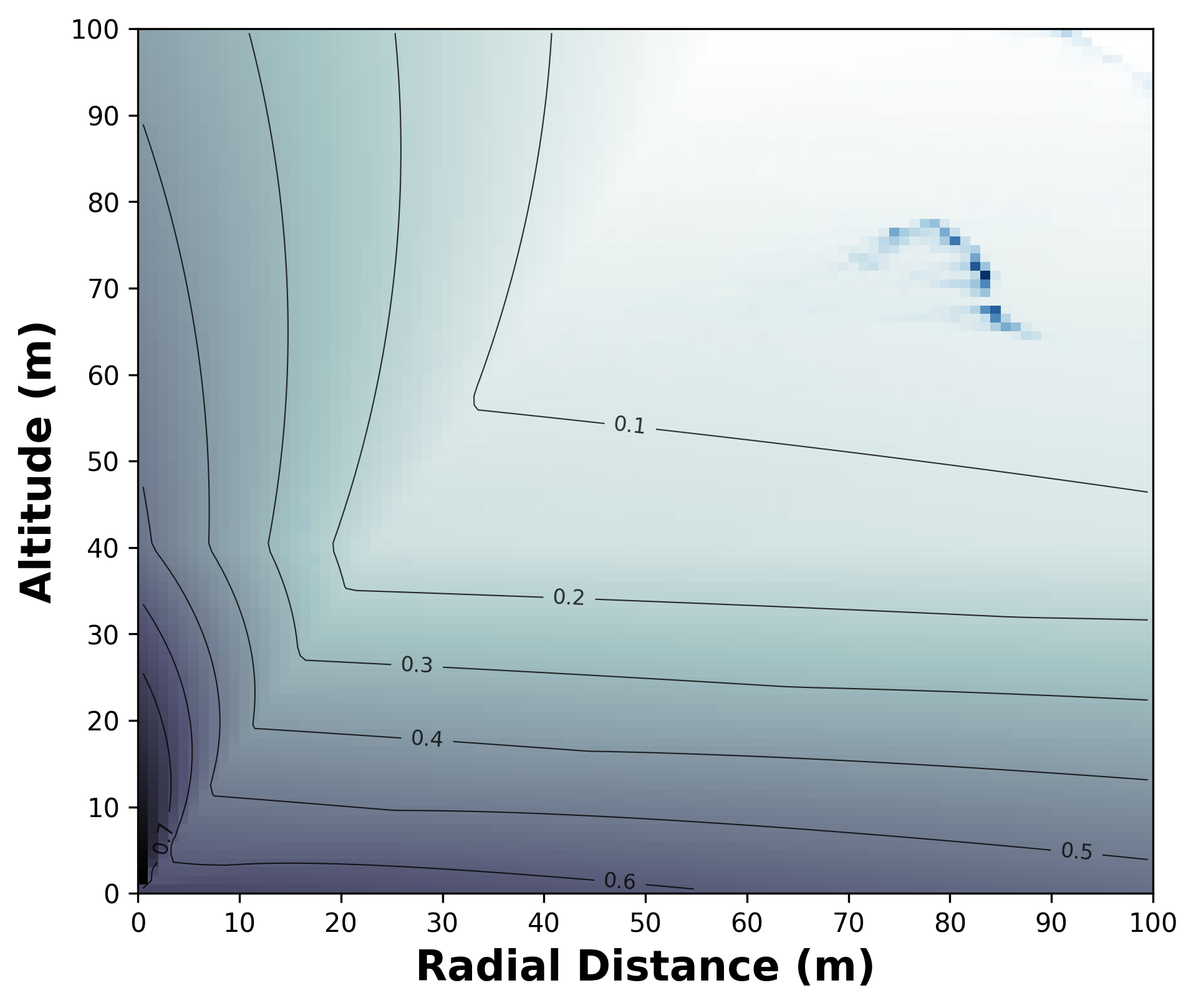}
        \label{fig:synthetic_cont_poi_ppo}
    \end{subfigure}
    \hfill
    \begin{subfigure}[t]{0.24\textwidth}
        \centering
        \includegraphics[width=\linewidth]{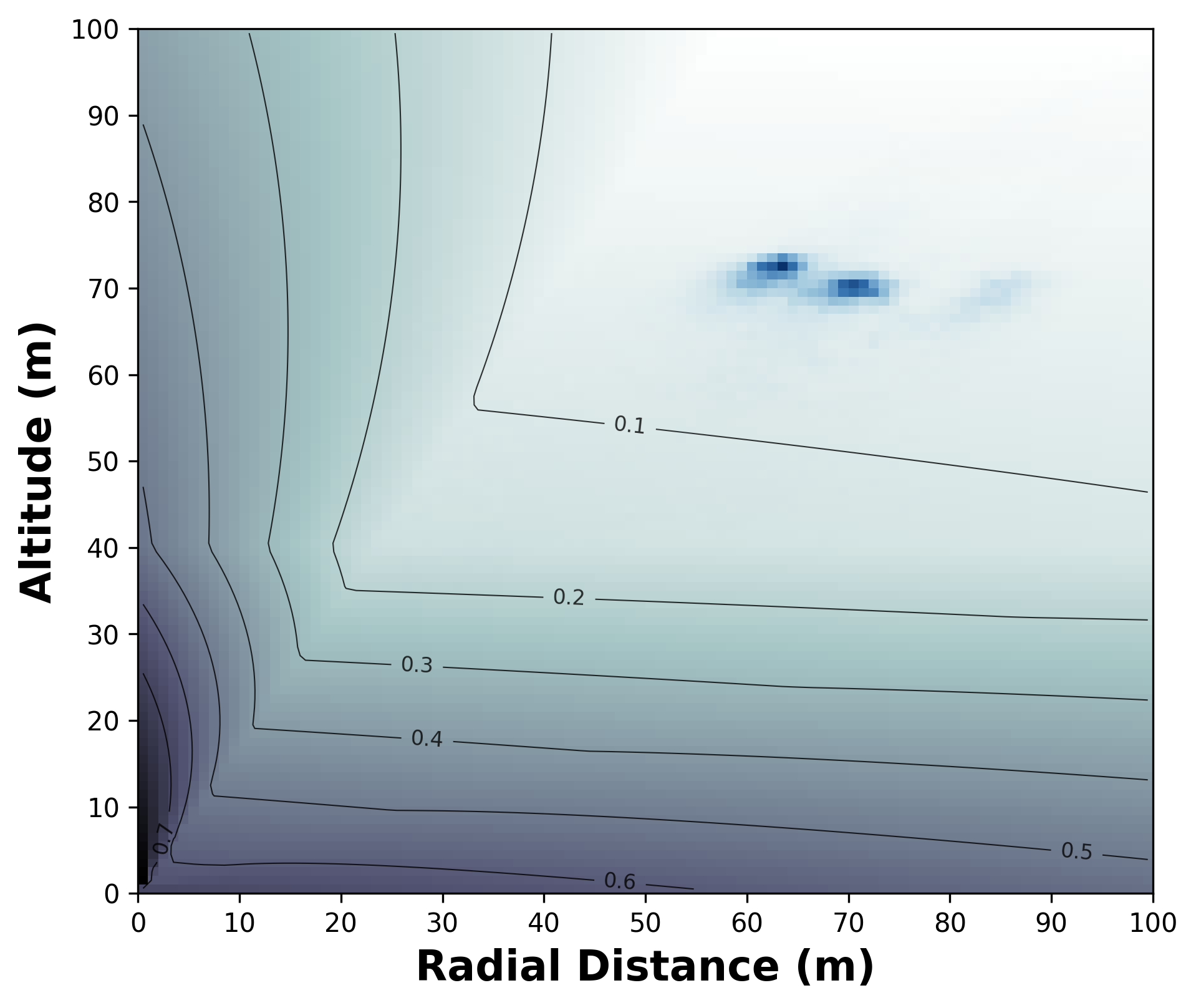}
        \label{fig:synthetic_cont_lpoi_ppo}
    \end{subfigure}

    \vspace{0.4cm}

    \begin{subfigure}[t]{0.24\textwidth}
        \centering
        \includegraphics[width=\linewidth]{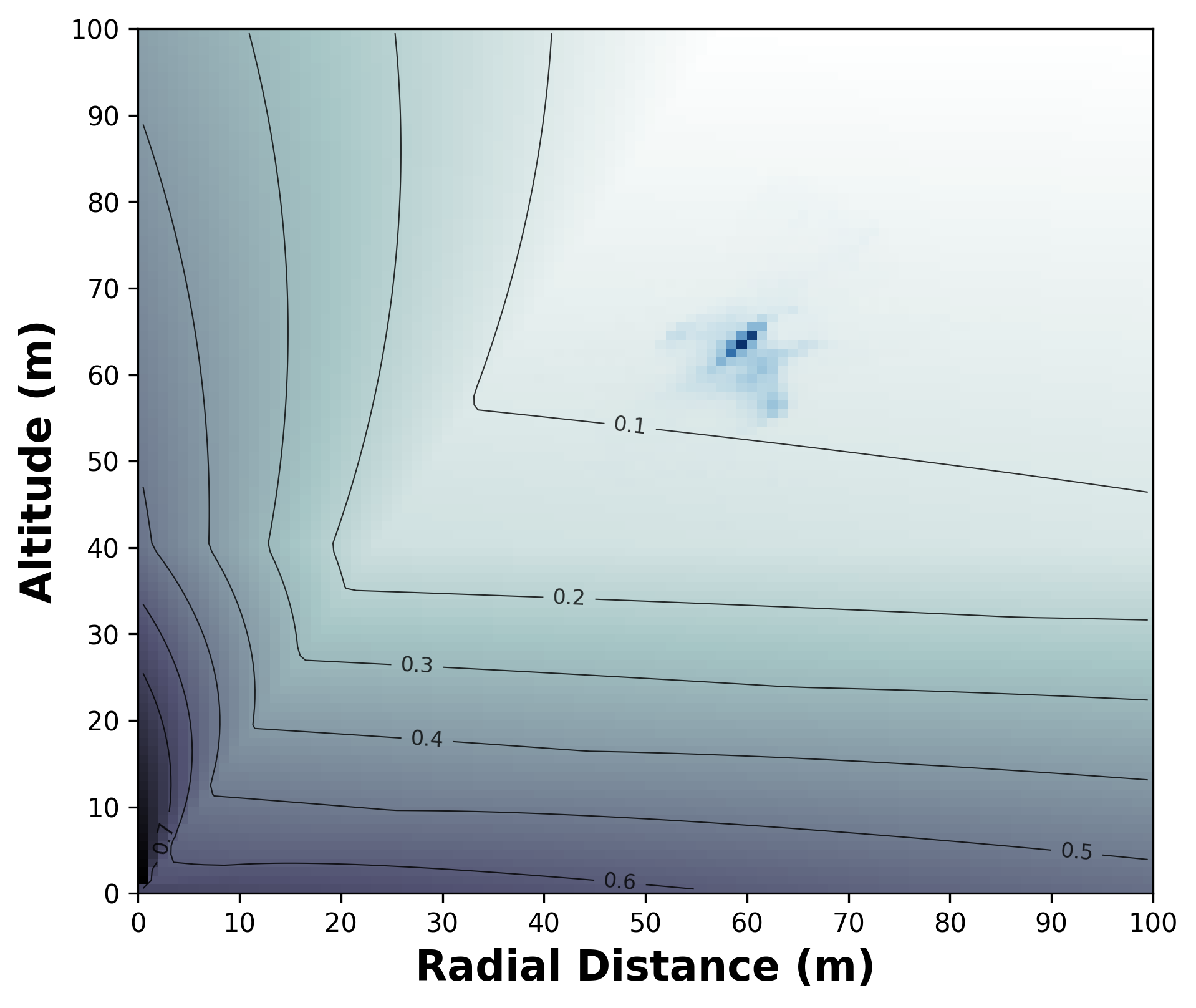}
        \label{fig:synthetic_cont_crw_sac}
    \end{subfigure}
    \hfill
    \begin{subfigure}[t]{0.24\textwidth}
        \centering
        \includegraphics[width=\linewidth]{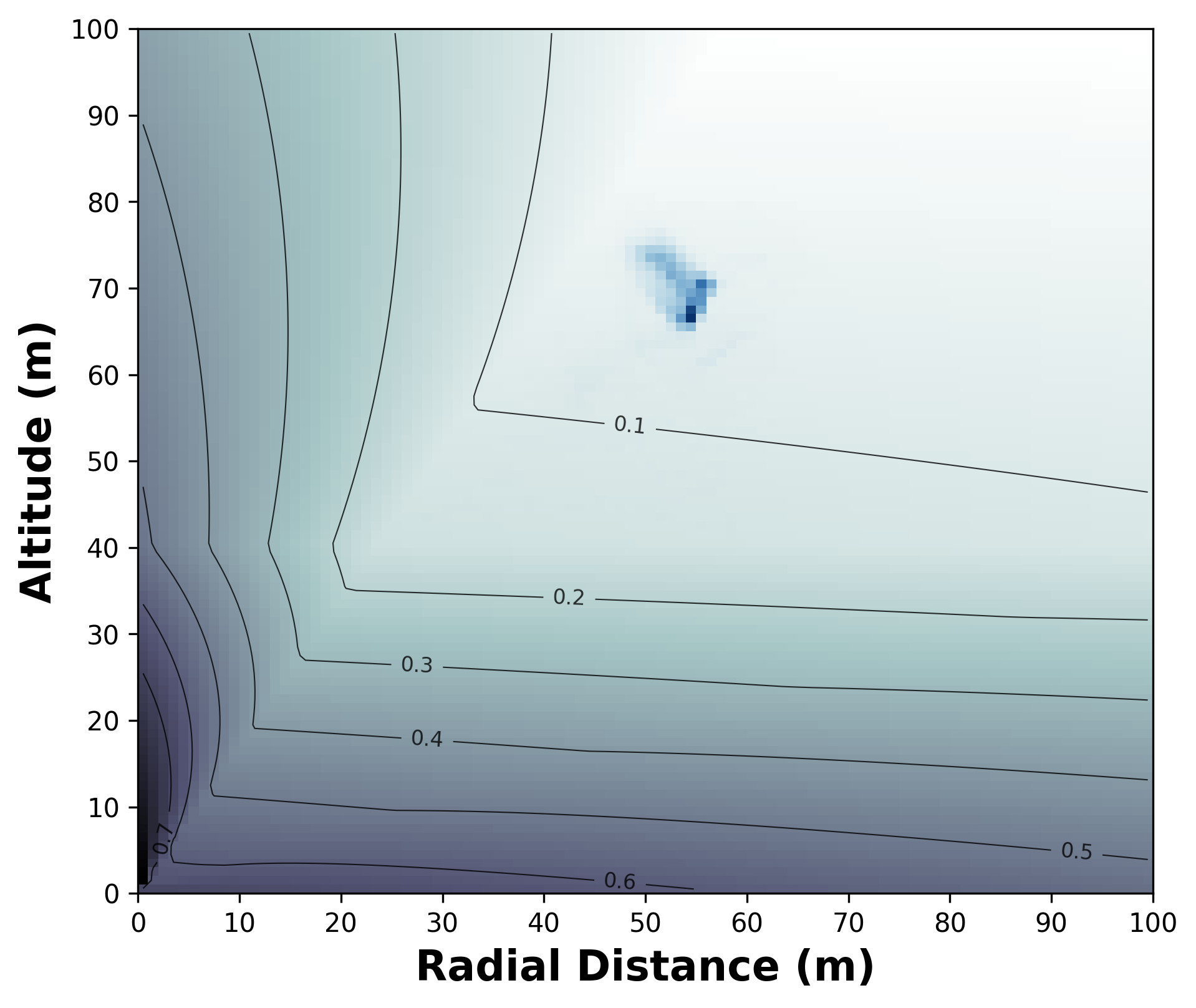}
        \label{fig:synthetic_cont_ee_sac}
    \end{subfigure}
    \hfill
    \begin{subfigure}[t]{0.24\textwidth}
        \centering
        \includegraphics[width=\linewidth]{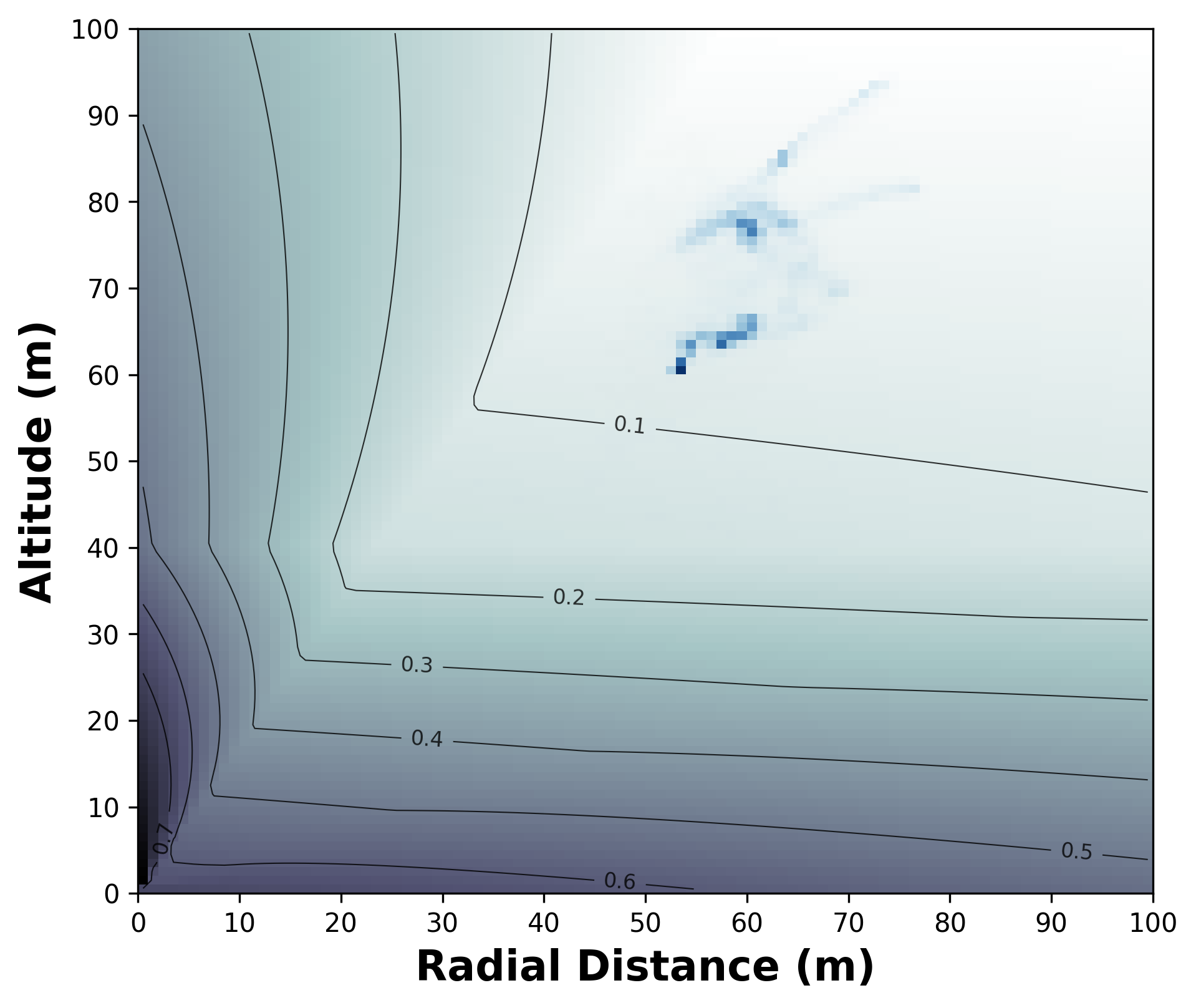}
        \label{fig:synthetic_cont_poi_sac}
    \end{subfigure}
    \hfill
    \begin{subfigure}[t]{0.24\textwidth}
        \centering
        \includegraphics[width=\linewidth]{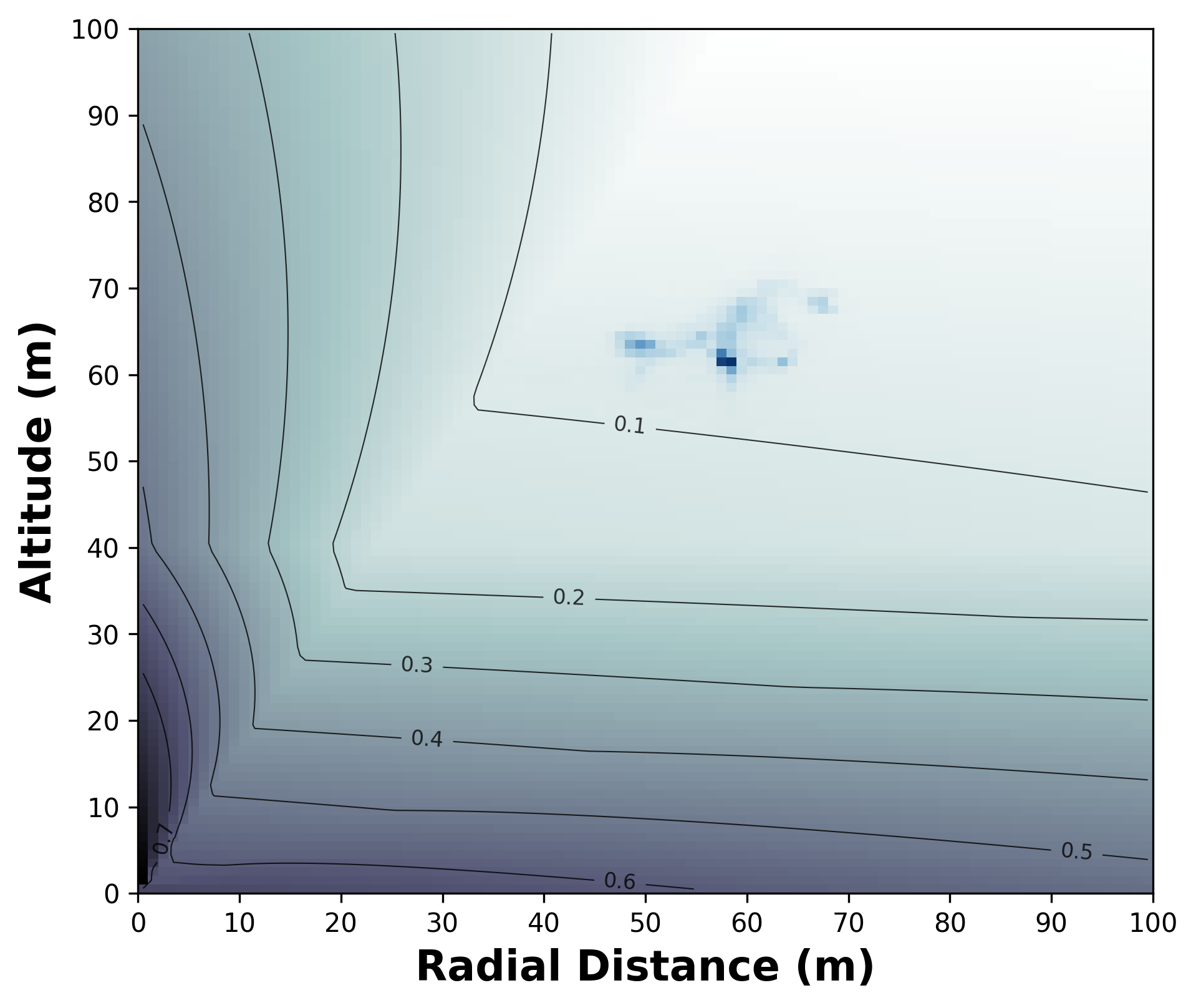}
        \label{fig:synthetic_cont_lpoi_sac}
    \end{subfigure}

    \caption{Spatial distribution of drone positions relative to the animal (located at position $x=0$,  $y=0$) for policies trained on synthetic movement behaviours and evaluated on synthetic data. Policy is evaluated on synthetic data. Columns correspond to the movement behaviour used during training: CRW, EE, POI and LPOI. Rows correspond to the trained agent: DQN in the first row, PPO in the second row, and SAC in the third row. Each panel shows drone radial distance from the target on the horizontal axis and drone altitude on the vertical axis. The blue density indicates where the policy most frequently positioned the drone during evaluation, while the background contour field indicates the corresponding disturbance level. No wind environment, D2 drone sensing capabilities.}
    \label{fig:spatial_distributions_synthetic_cont}
\end{figure*}

The following figures visualize where learned policies position the drone relative to the monitored animal. Figure~\ref{fig:spatial_distributions_synthetic_cont} shows radial distance and altitude distributions for policies trained and evaluated on synthetic movement models. These distributions help interpret whether each policy tends to monitor from close range, maintain larger stand-off distances, or occupy disturbance-sensitive regions of the state space.

\newpage

\begin{figure*}[!ht]
    \centering

    \begin{subfigure}[t]{0.32\textwidth}
        \centering
        \includegraphics[width=\linewidth]{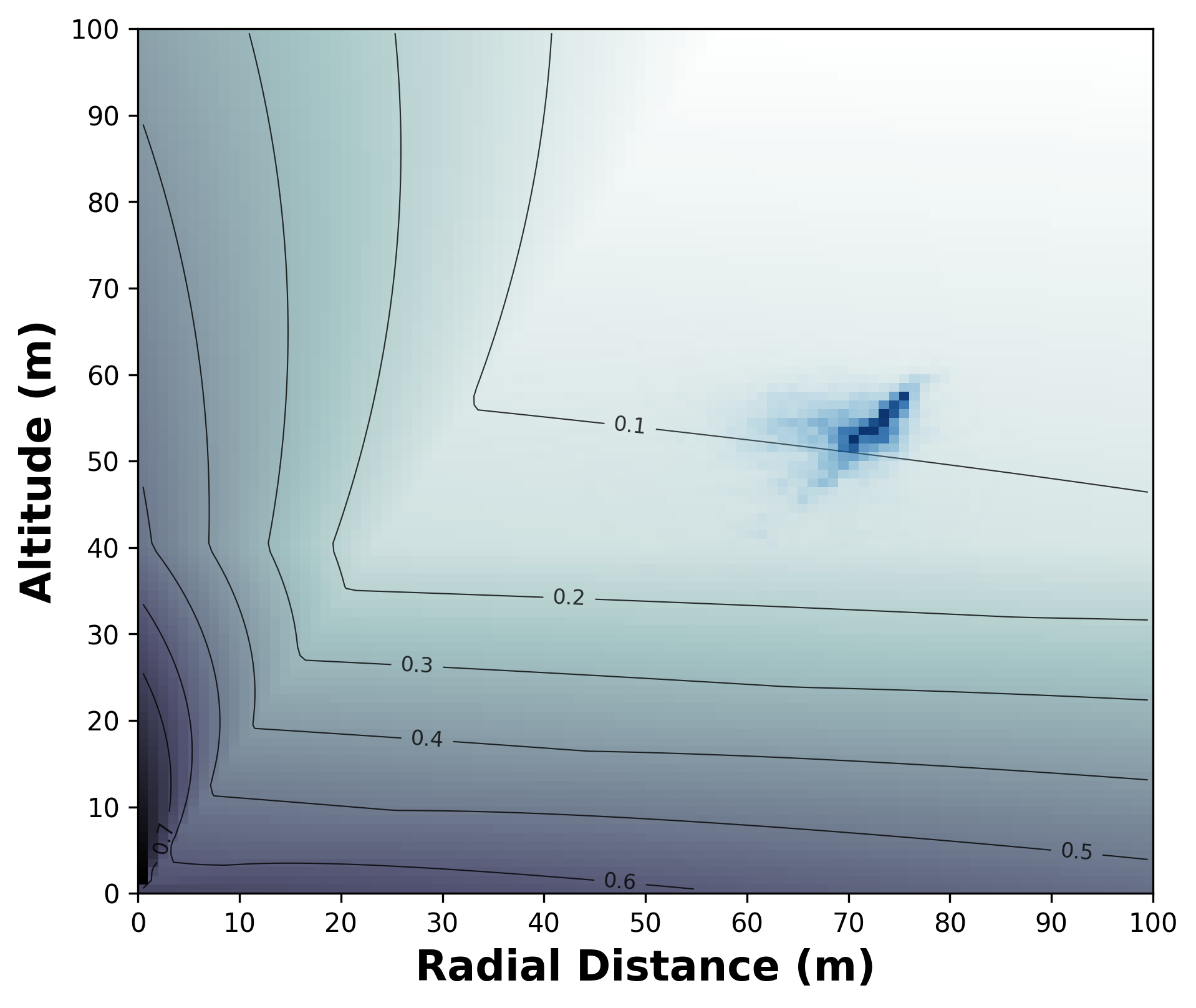}
        \caption{}
        \label{fig:prior_1}
    \end{subfigure}
    \hfill
    \begin{subfigure}[t]{0.32\textwidth}
        \centering
        \includegraphics[width=\linewidth]{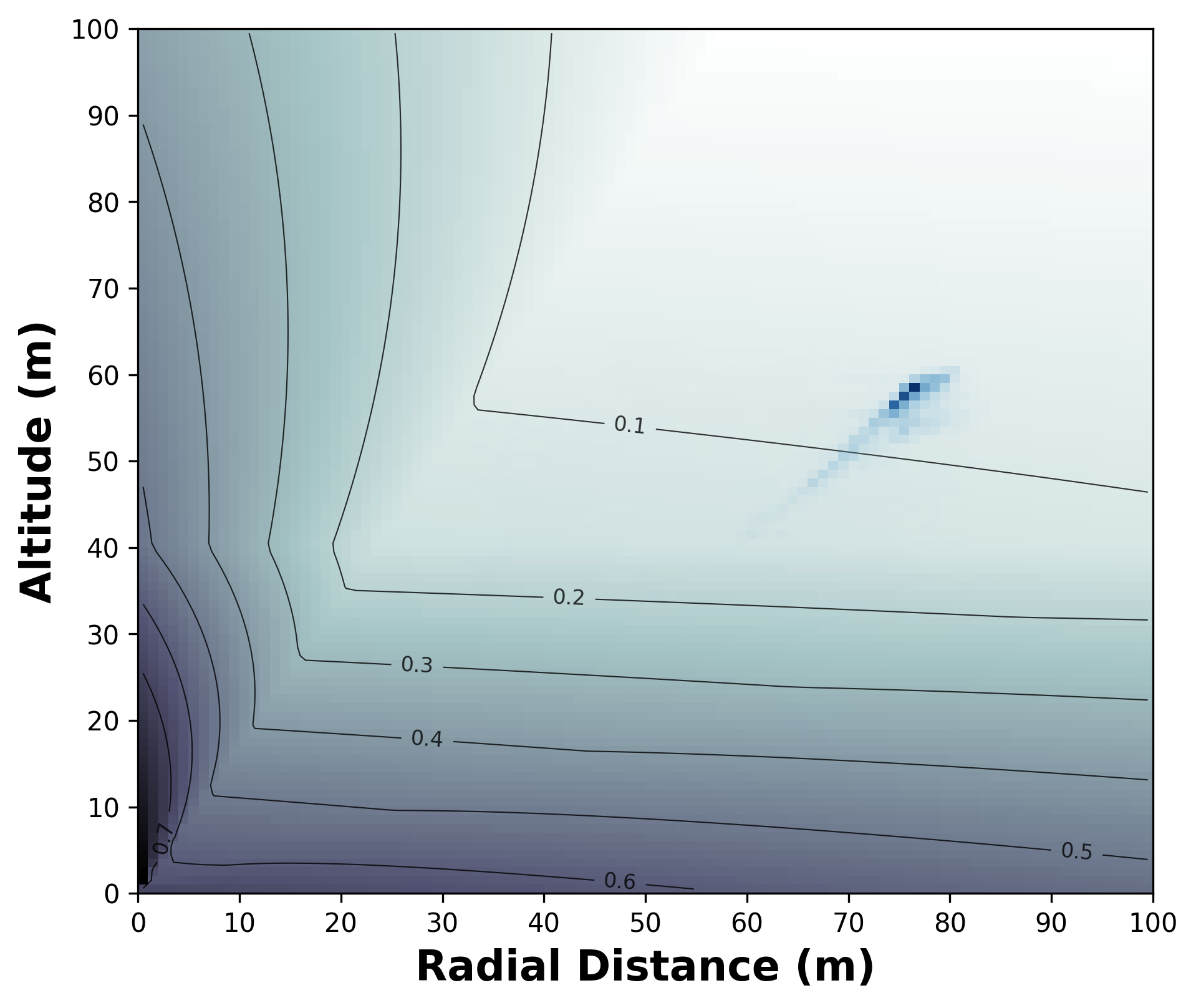}
        \caption{}
        \label{fig:prior_2}
    \end{subfigure}
    \hfill
    \begin{subfigure}[t]{0.32\textwidth}
        \centering
        \includegraphics[width=\linewidth]{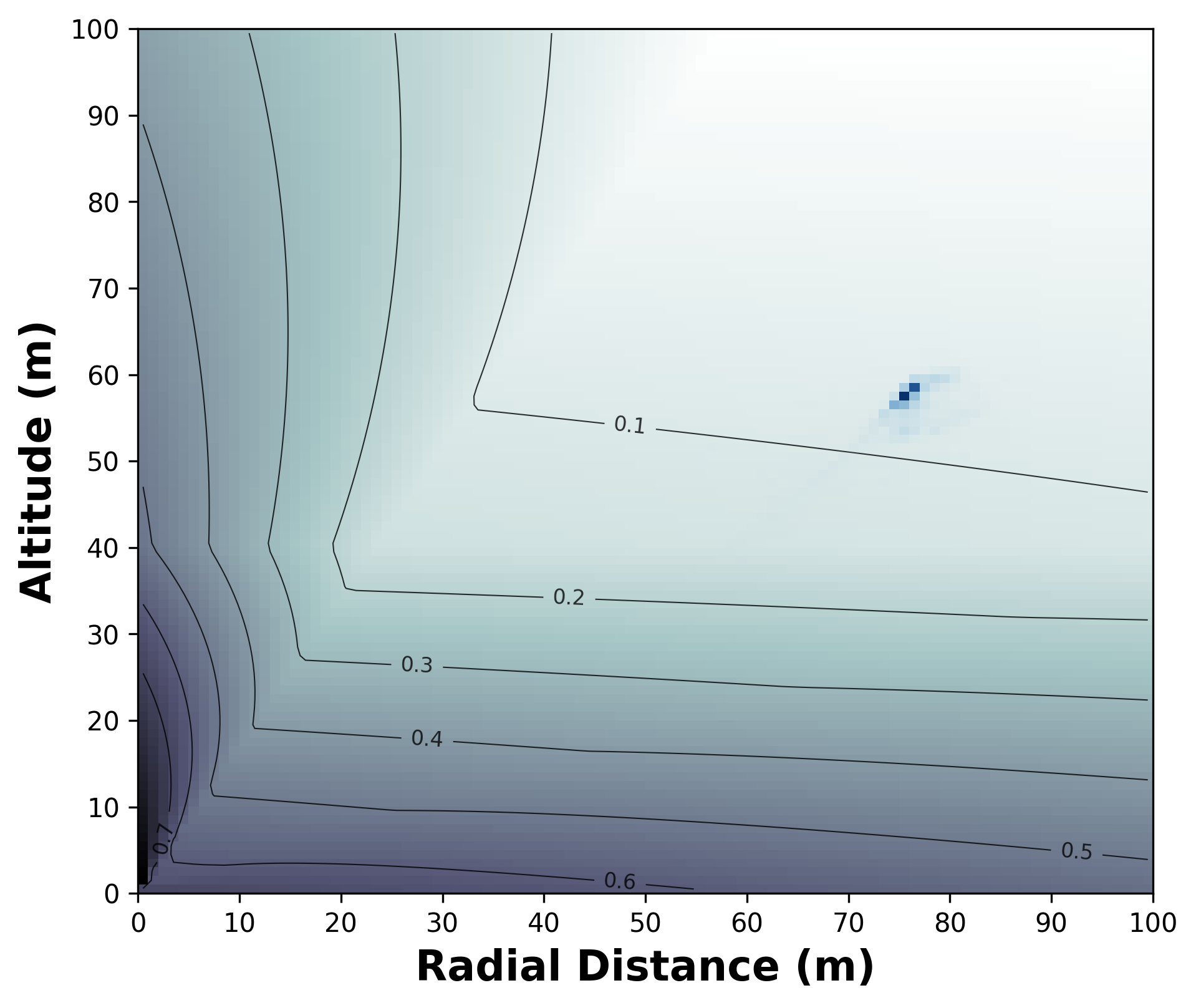}
        \caption{}
        \label{fig:prior_3}
    \end{subfigure}

    \vspace{0.4cm}

    \begin{subfigure}[t]{0.32\textwidth}
        \centering
        \includegraphics[width=\linewidth]{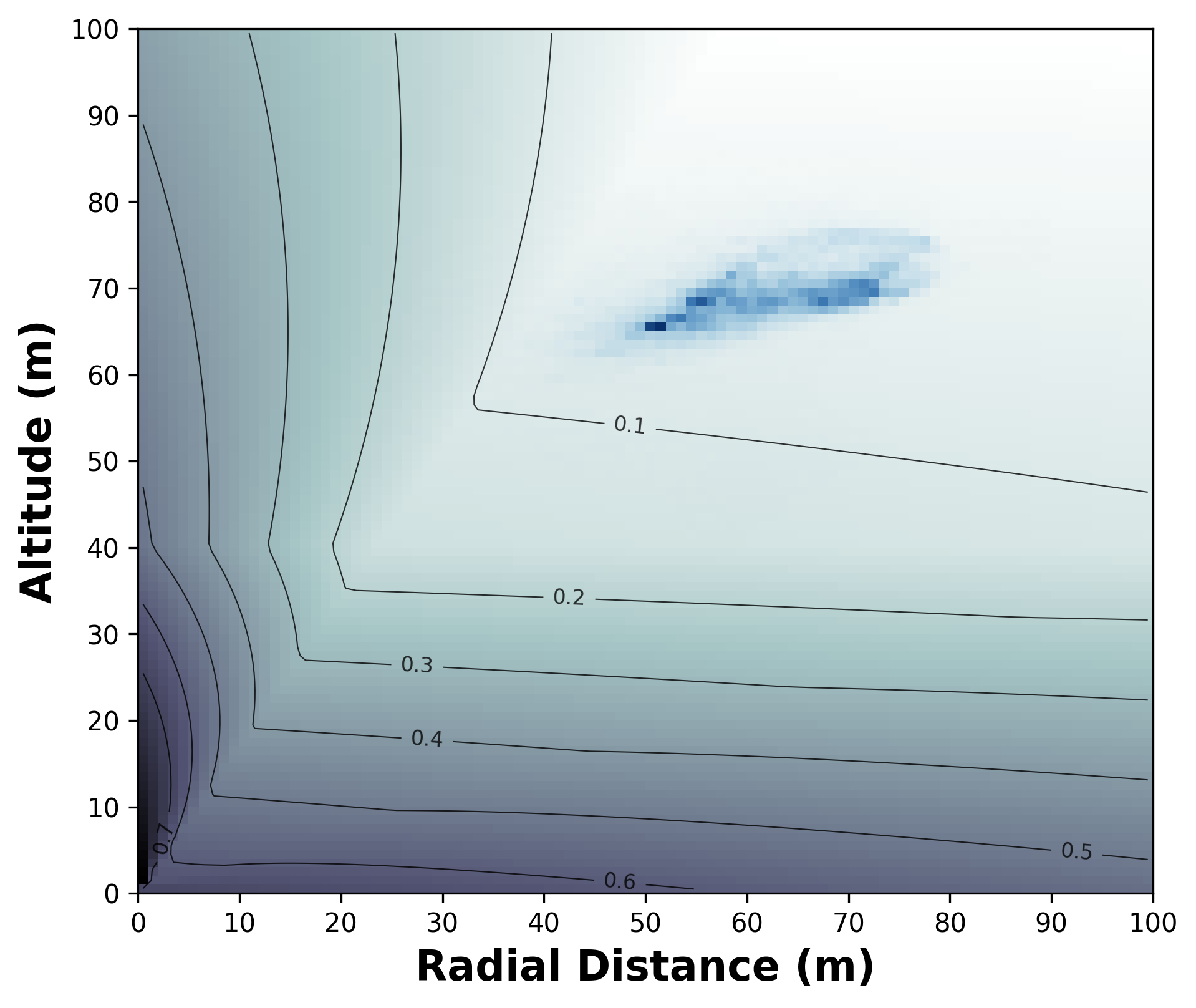}
        \caption{}
        \label{fig:prior_4}
    \end{subfigure}
    \hfill
    \begin{subfigure}[t]{0.32\textwidth}
        \centering
        \includegraphics[width=\linewidth]{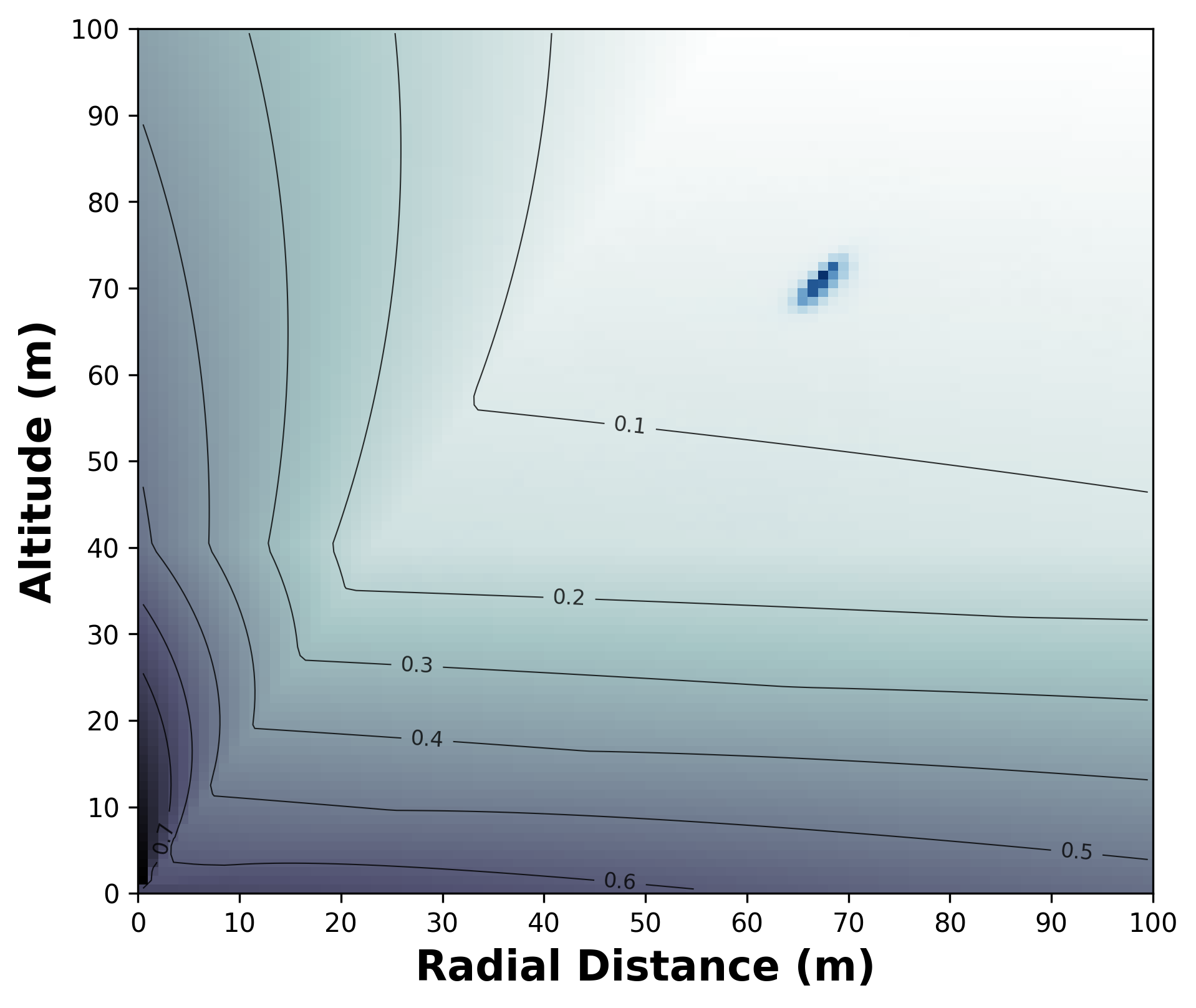}
        \caption{}
        \label{fig:prior_5}
    \end{subfigure}
    \hfill
    \begin{subfigure}[t]{0.32\textwidth}
        \centering
        \includegraphics[width=\linewidth]{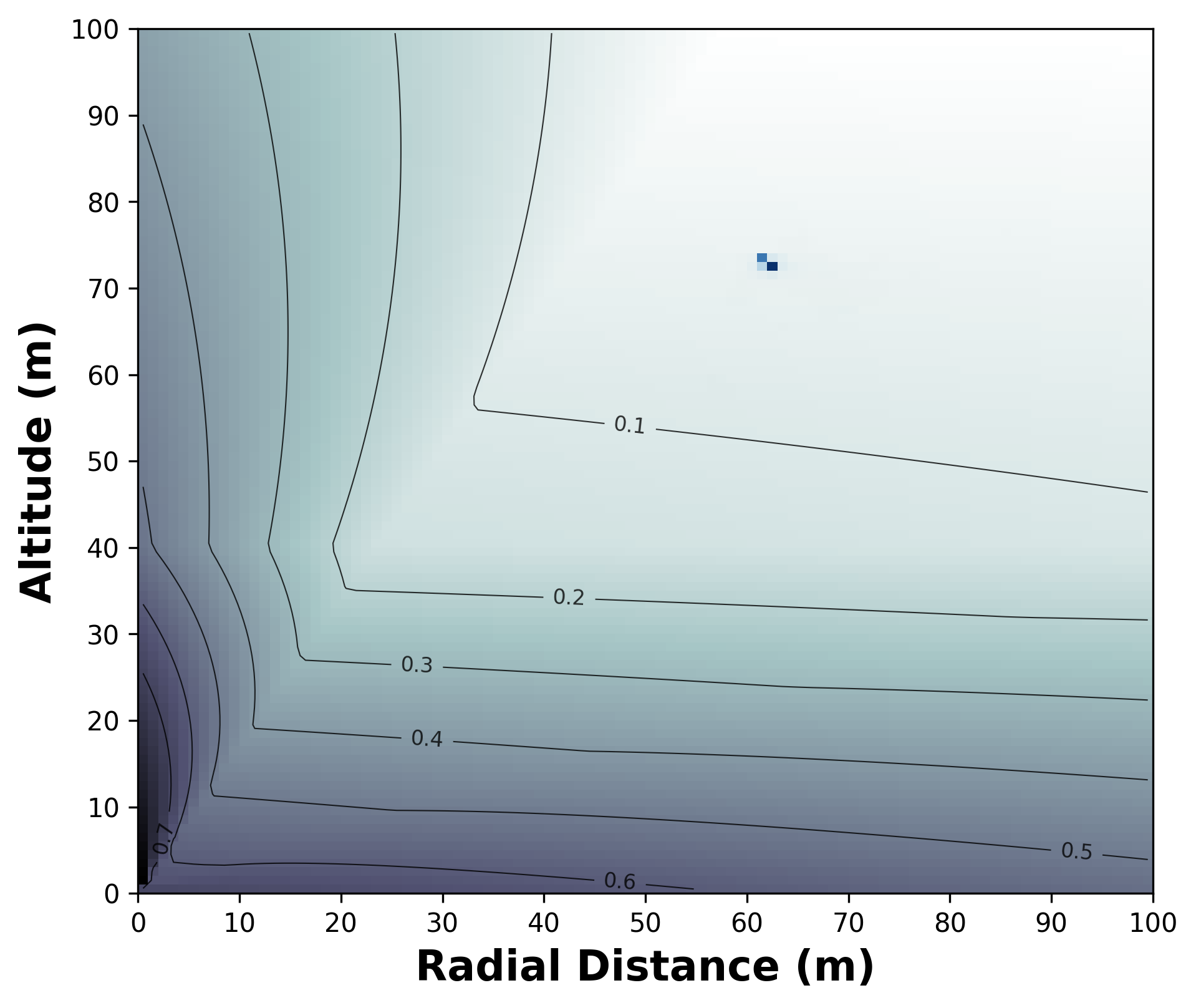}
        \caption{}
        \label{fig:prior_6}
    \end{subfigure}

    \vspace{0.4cm}

    \begin{subfigure}[t]{0.32\textwidth}
        \centering
        \includegraphics[width=\linewidth]{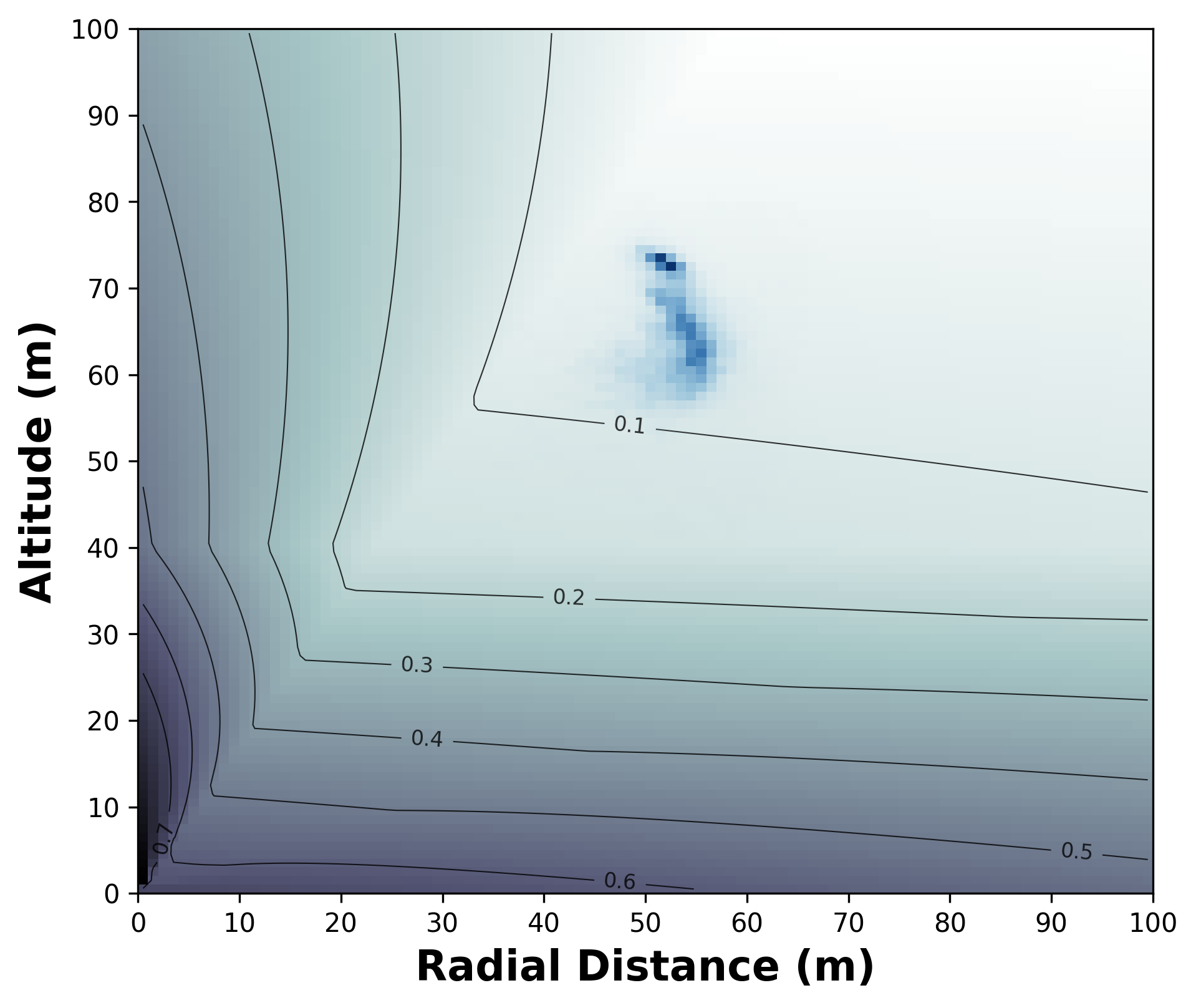}
        \caption{}
        \label{fig:prior_7}
    \end{subfigure}
    \hfill
    \begin{subfigure}[t]{0.32\textwidth}
        \centering
        \includegraphics[width=\linewidth]{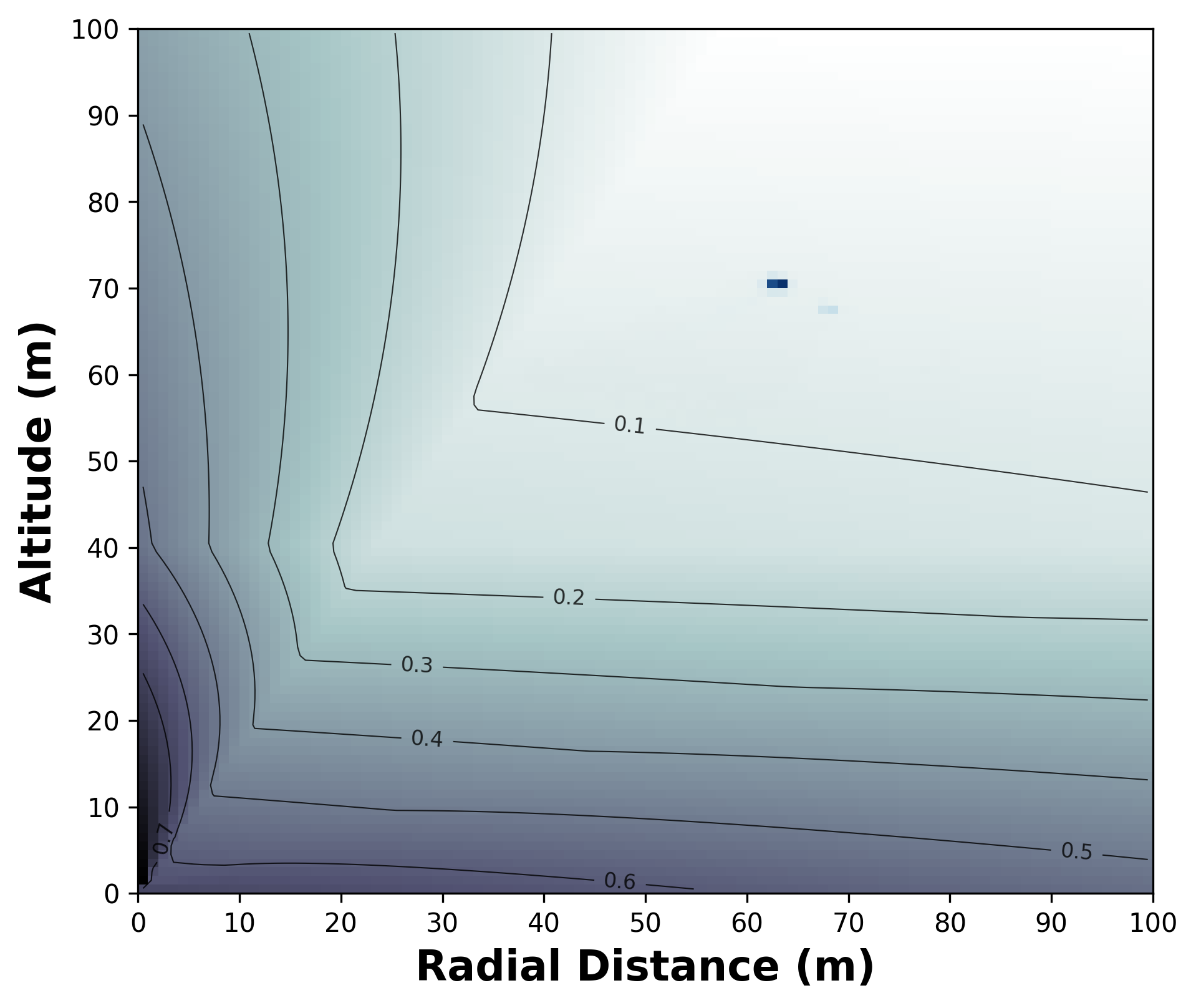}
        \caption{}
        \label{fig:prior_8}
    \end{subfigure}
    \hfill
    \begin{subfigure}[t]{0.32\textwidth}
        \centering
        \includegraphics[width=\linewidth]{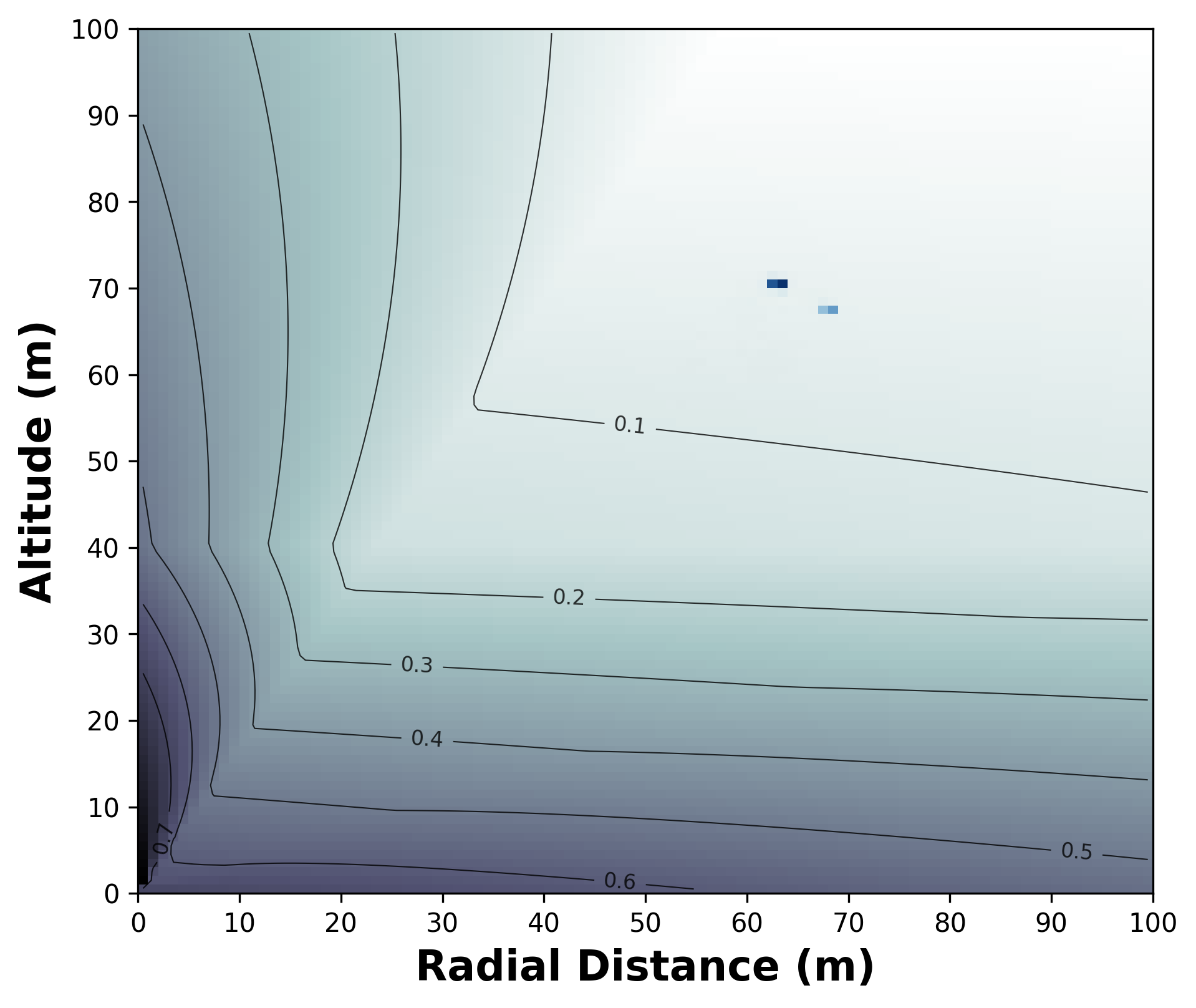}
        \caption{}
        \label{fig:prior_9}
    \end{subfigure}

    \caption{Spatial distribution of drone positions relative to the animal (located at position $x=0$,  $y=0$) for the best movement-prior policy selected for each model and animal. Policy is evaluated on real-sampled GPS animal movement data. Rows correspond to the trained agent: DQN in the first row, PPO in the second row, and SAC in the third row. Columns correspond to the animal species: Jackals in the first column, Pigeons in the second column, and Spur-winged lapwings in the third column. Each panel shows drone radial distance from the animal on the horizontal axis and drone altitude on the vertical axis. The blue density indicates where the policy most frequently positioned the drone during evaluation, while the background contour field indicates the corresponding disturbance level. No wind environment, D2 drone sensing capabilities.}
    \label{fig:spatial_distributions_gps_cont}
\end{figure*}

Figure~\ref{fig:spatial_distributions_gps_cont} shows the corresponding radial-distance and altitude distributions when the selected policies are evaluated on empirical GPS replay trajectories. This comparison illustrates whether the spatial monitoring strategies learned in simulation are preserved when animal movement follows sampled GPS trajectories rather than synthetic movement priors.

\begin{figure*}[!ht]
    \centering

    \begin{subfigure}[t]{0.24\textwidth}
        \centering
        \includegraphics[width=\linewidth]{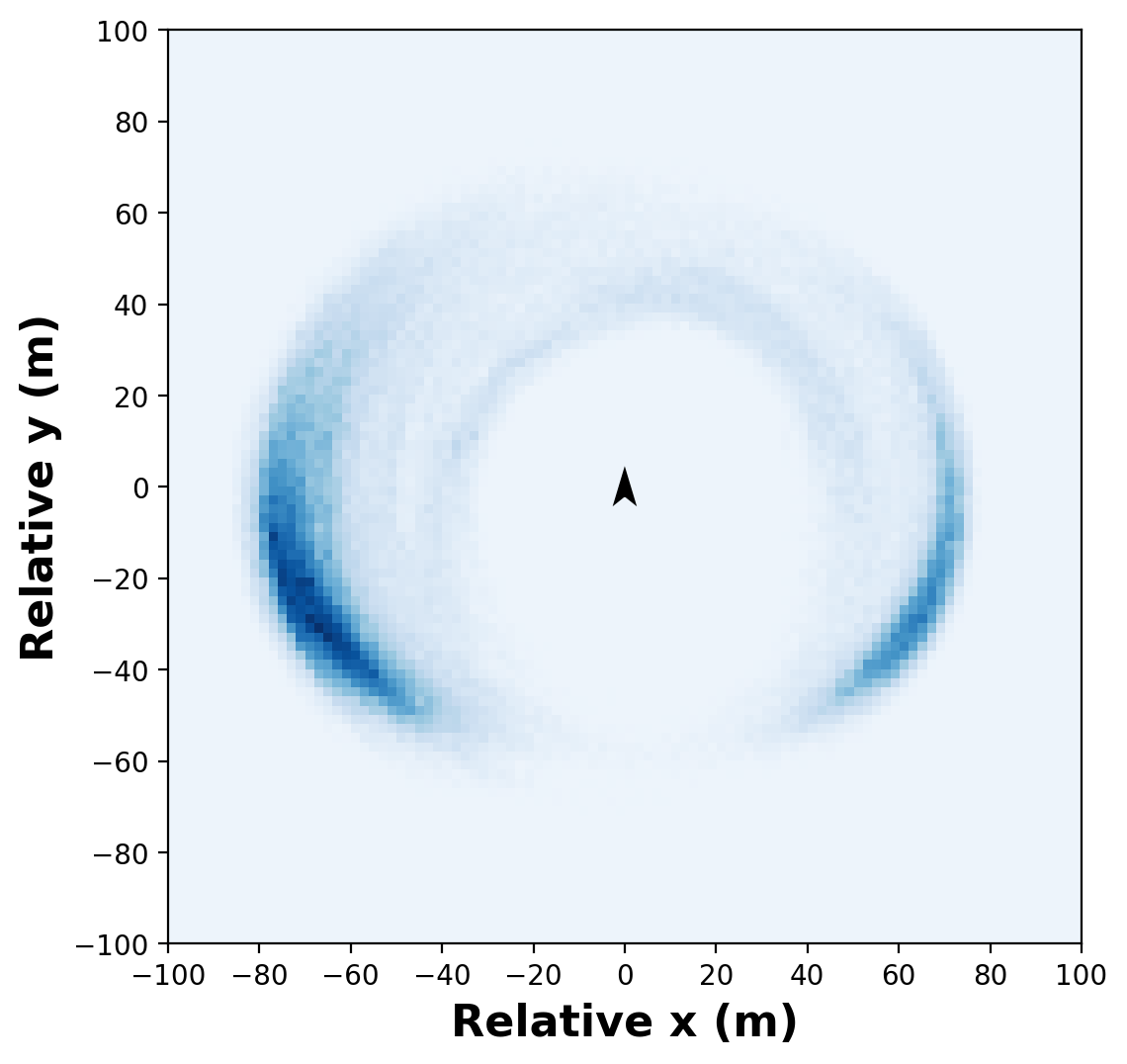}
        \label{fig:synthetic_visit_crw_dqn}
    \end{subfigure}
    \hfill
    \begin{subfigure}[t]{0.24\textwidth}
        \centering
        \includegraphics[width=\linewidth]{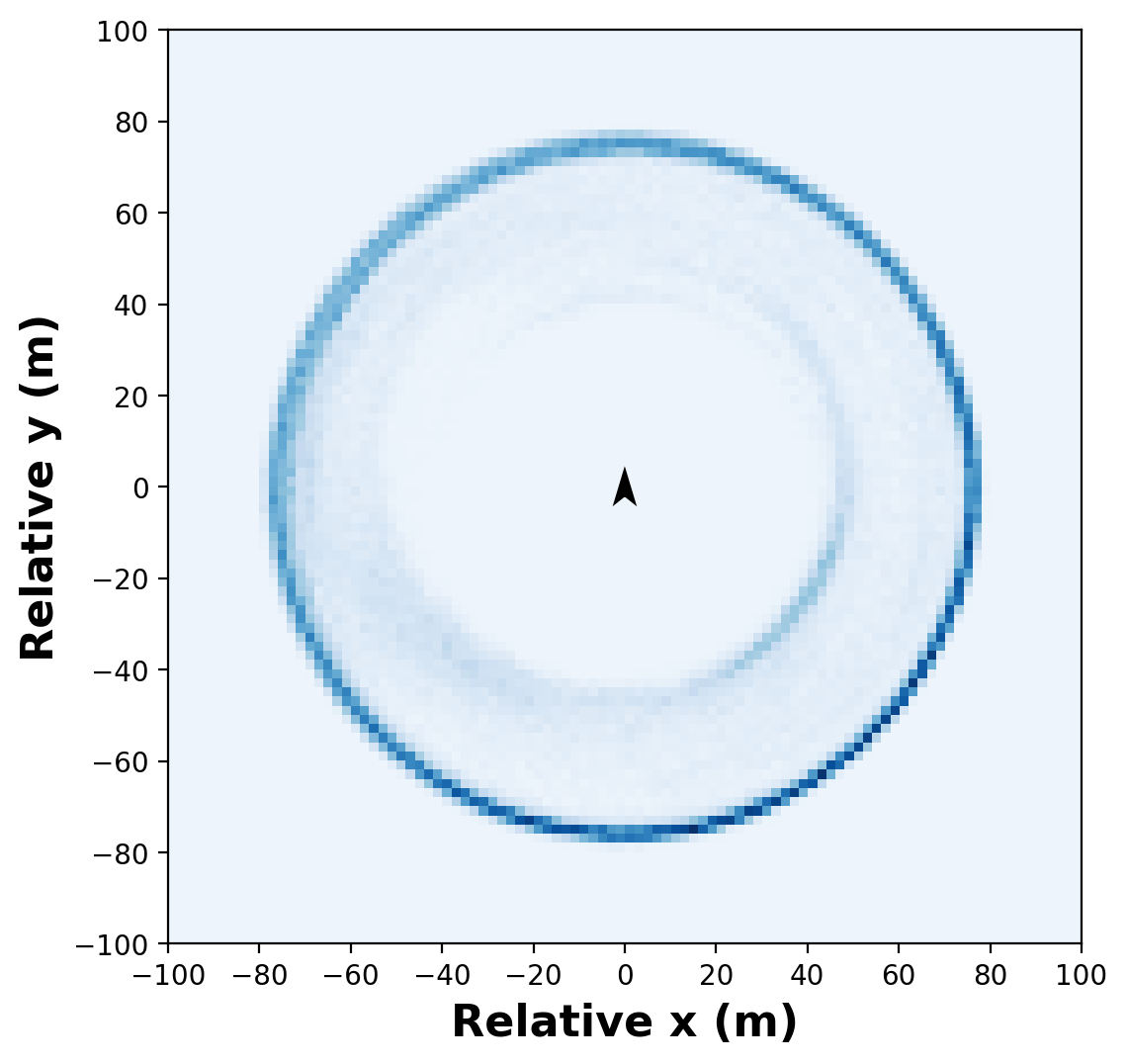}
        \label{fig:synthetic_visit_ee_dqn}
    \end{subfigure}
    \hfill
    \begin{subfigure}[t]{0.24\textwidth}
        \centering
        \includegraphics[width=\linewidth]{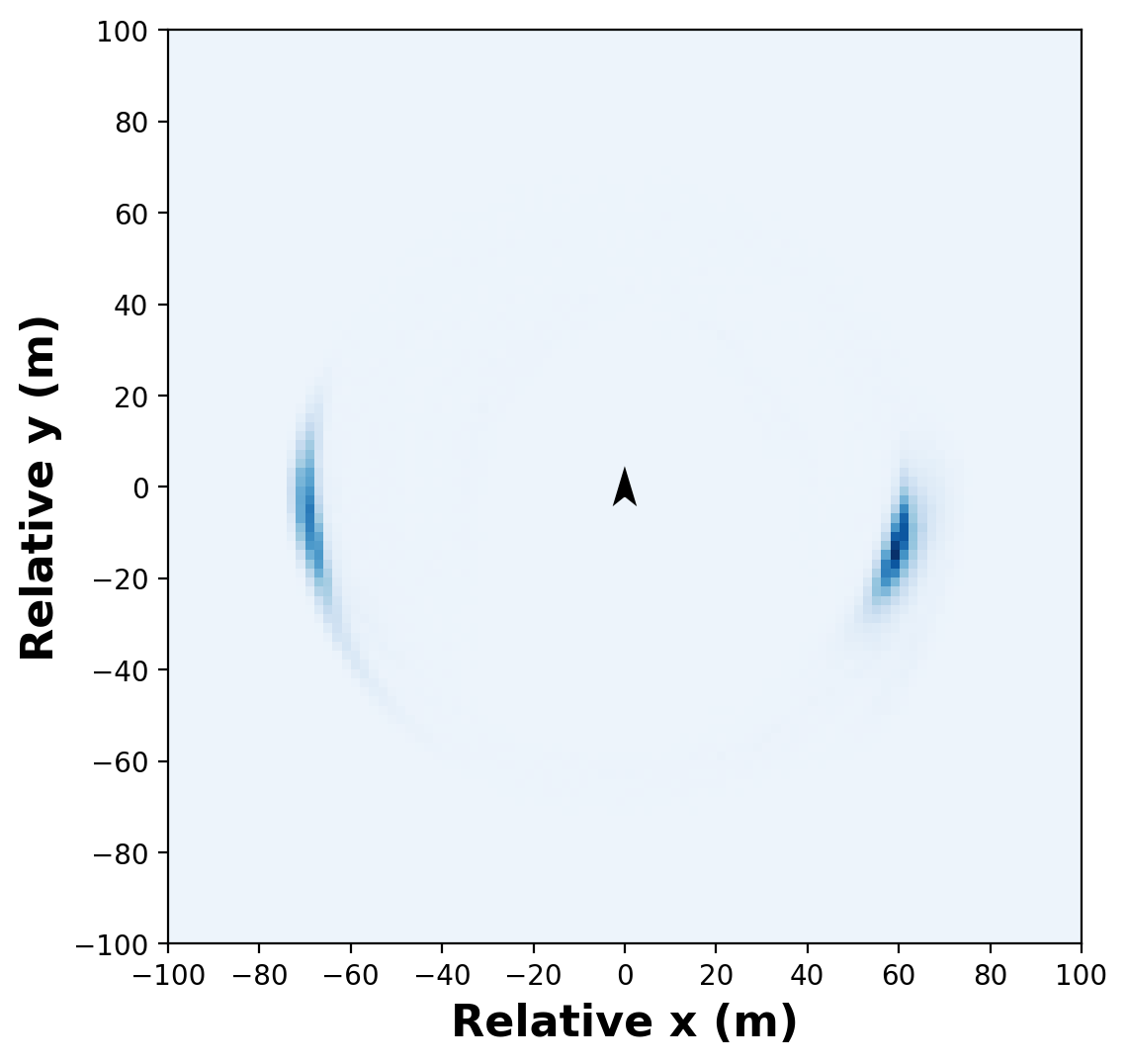}
        \label{fig:synthetic_visit_poi_dqn}
    \end{subfigure}
    \hfill
    \begin{subfigure}[t]{0.24\textwidth}
        \centering
        \includegraphics[width=\linewidth]{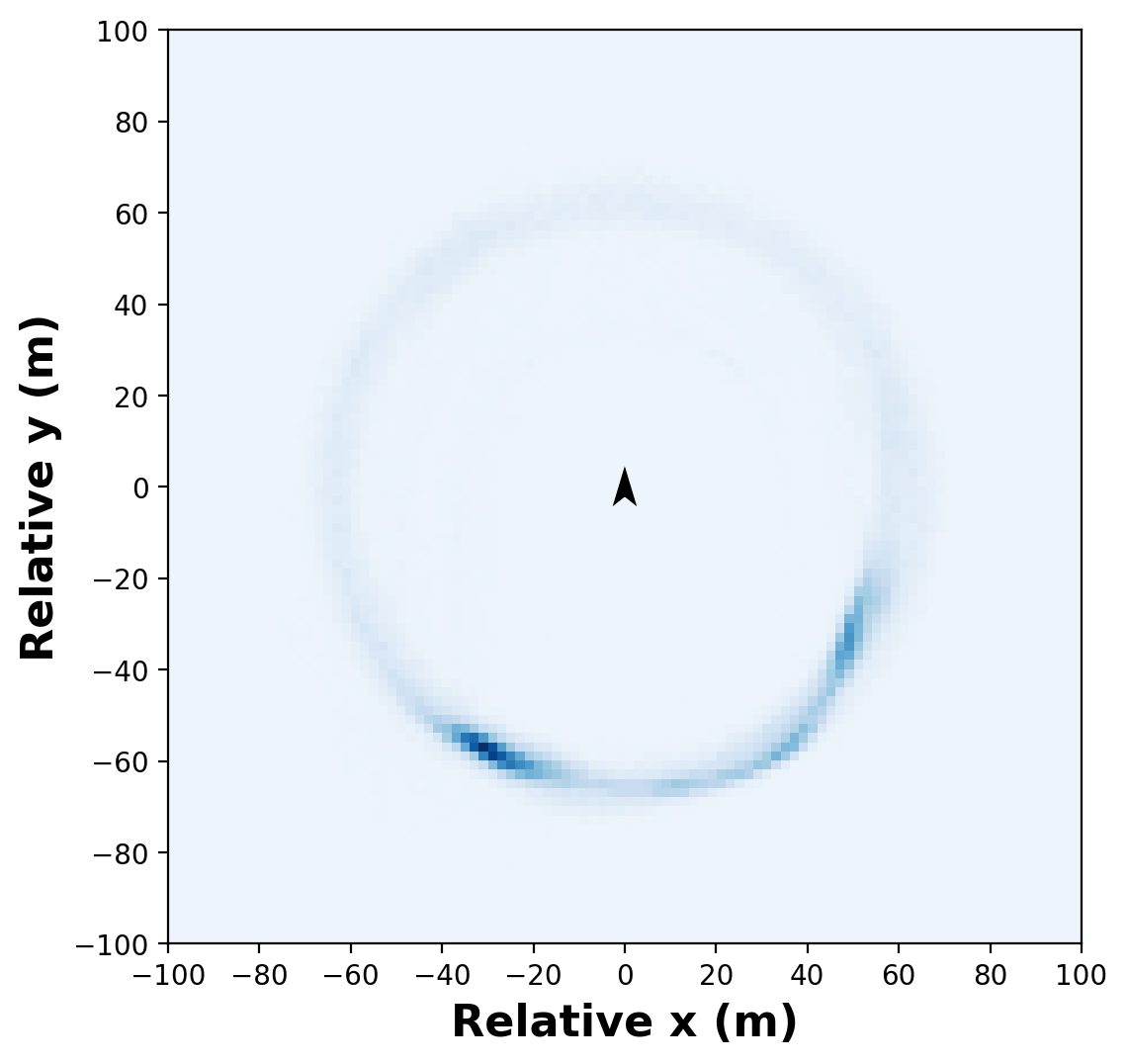}
        \label{fig:synthetic_visit_lpoi_dqn}
    \end{subfigure}

    \vspace{0.4cm}

    \begin{subfigure}[t]{0.24\textwidth}
        \centering
        \includegraphics[width=\linewidth]{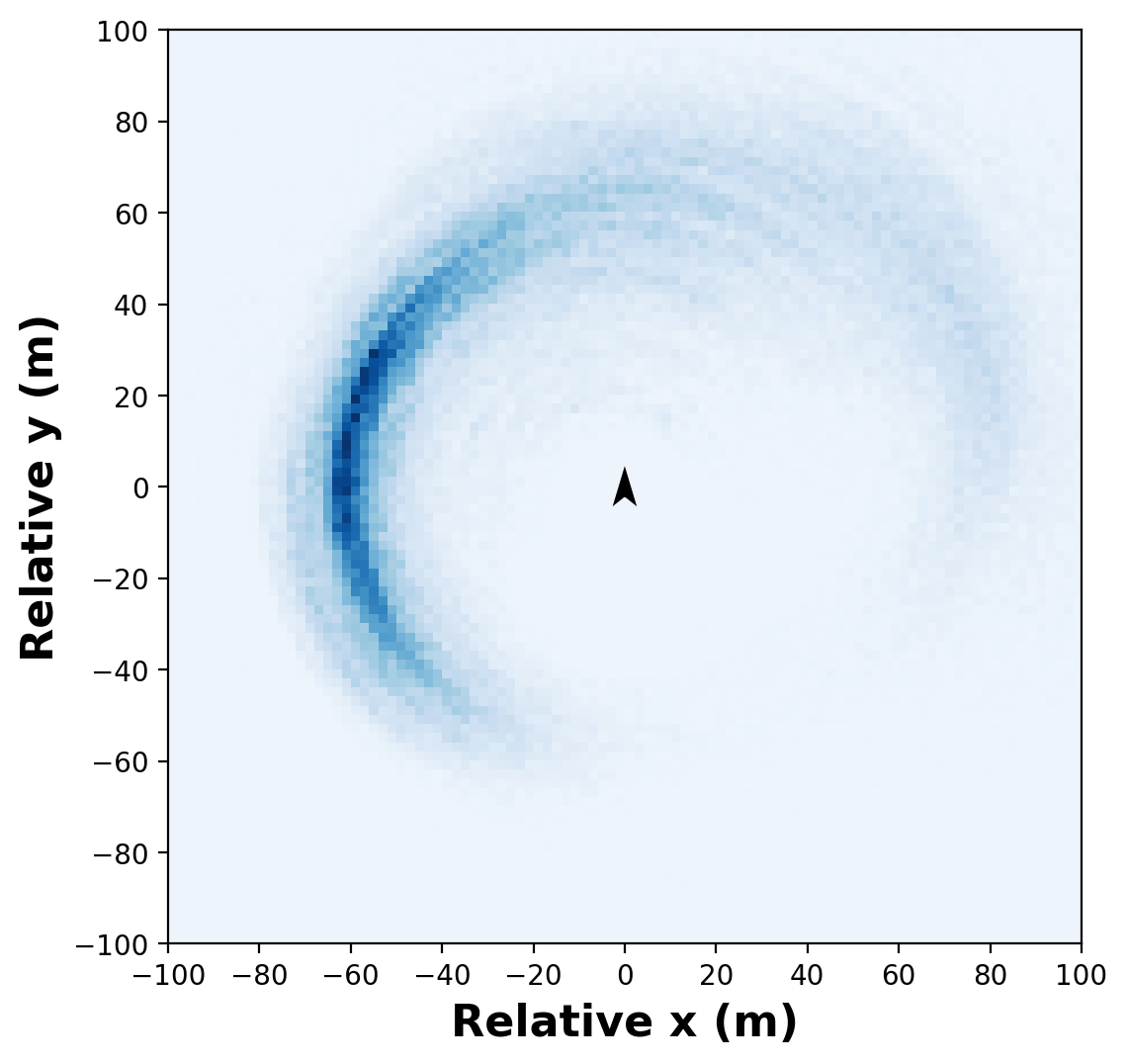}
        \label{fig:synthetic_visit_crw_ppo}
    \end{subfigure}
    \hfill
    \begin{subfigure}[t]{0.24\textwidth}
        \centering
        \includegraphics[width=\linewidth]{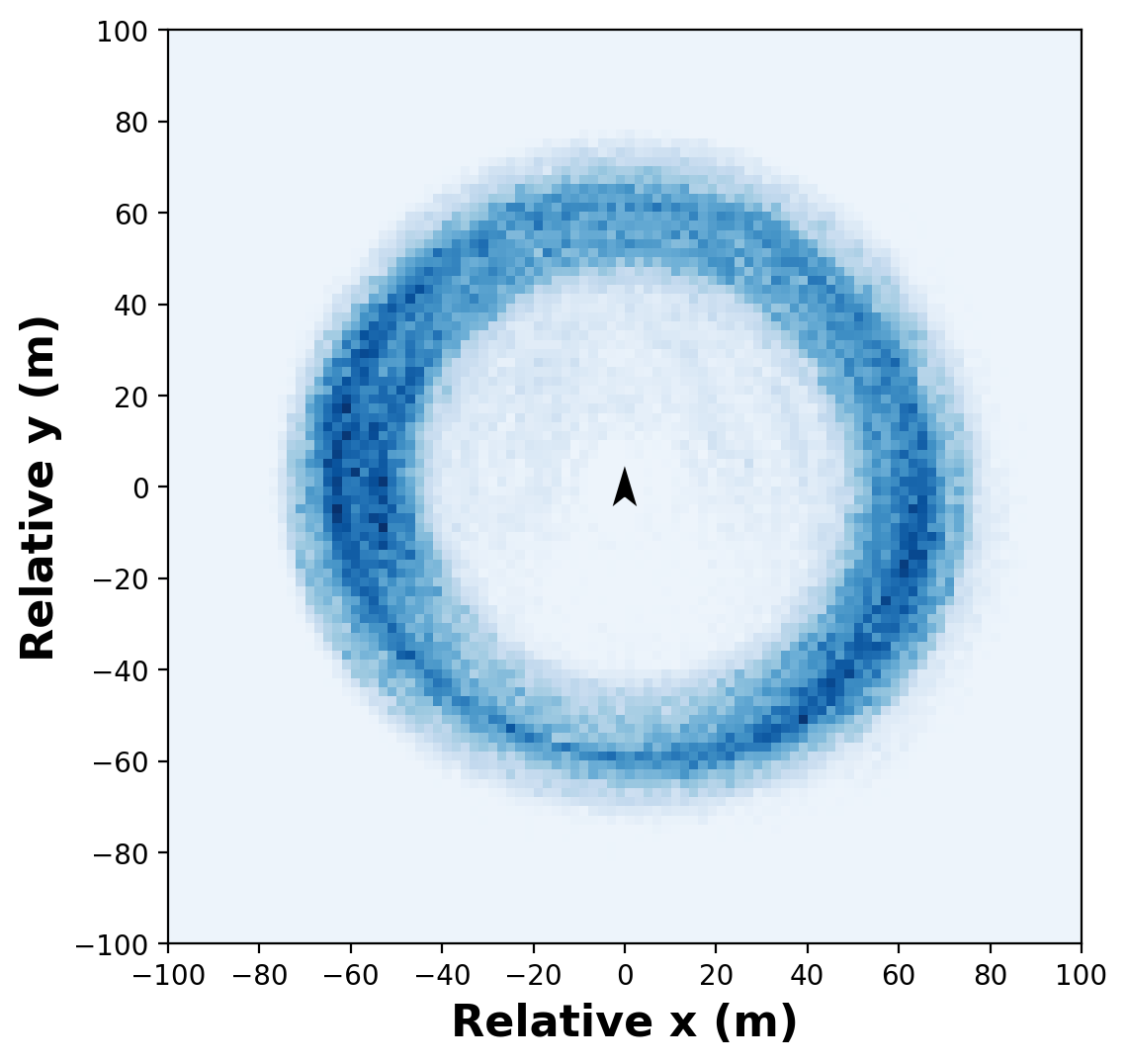}
        \label{fig:synthetic_visit_ee_ppo}
    \end{subfigure}
    \hfill
    \begin{subfigure}[t]{0.24\textwidth}
        \centering
        \includegraphics[width=\linewidth]{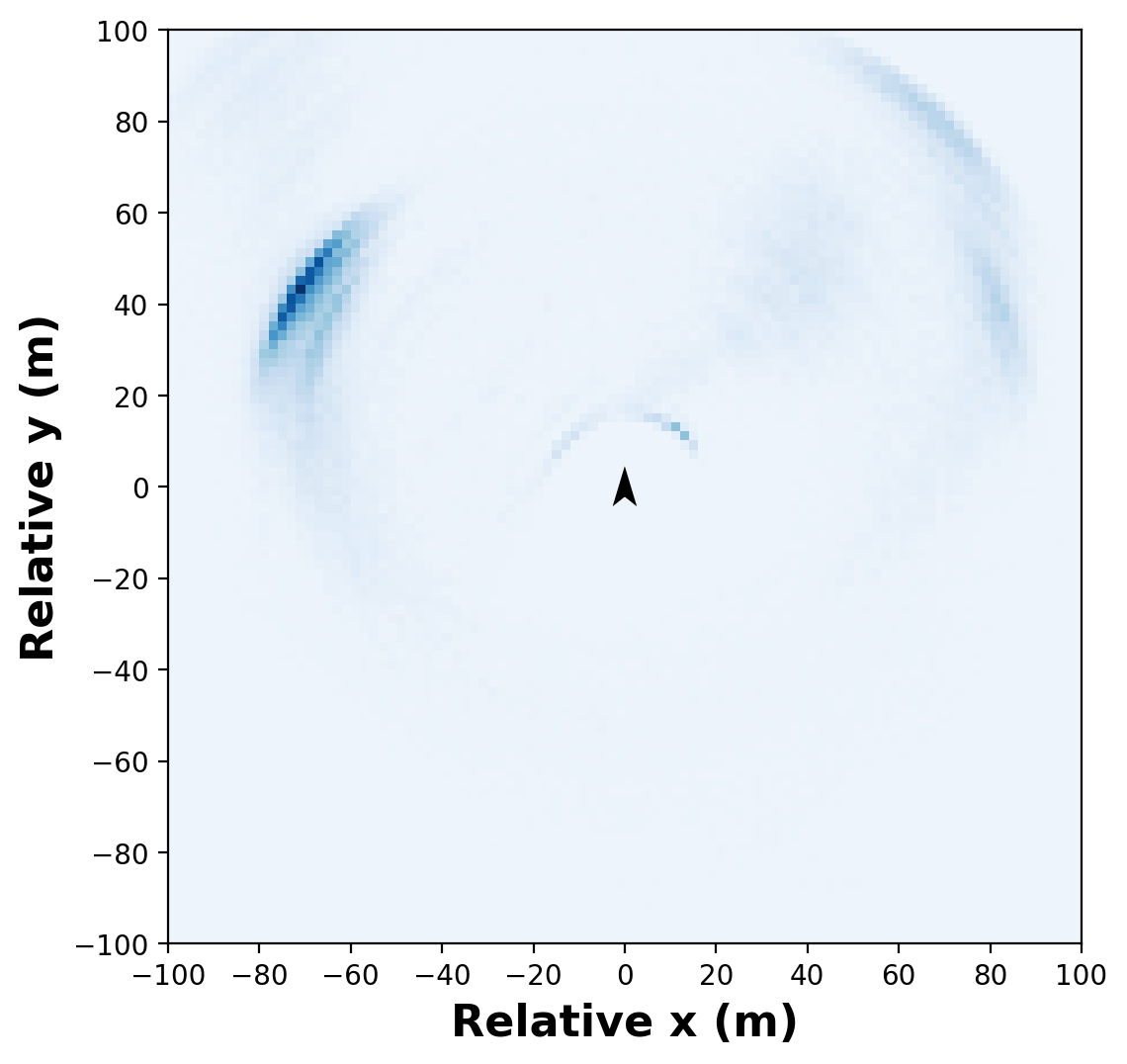}
        \label{fig:synthetic_visit_poi_ppo}
    \end{subfigure}
    \hfill
    \begin{subfigure}[t]{0.24\textwidth}
        \centering
        \includegraphics[width=\linewidth]{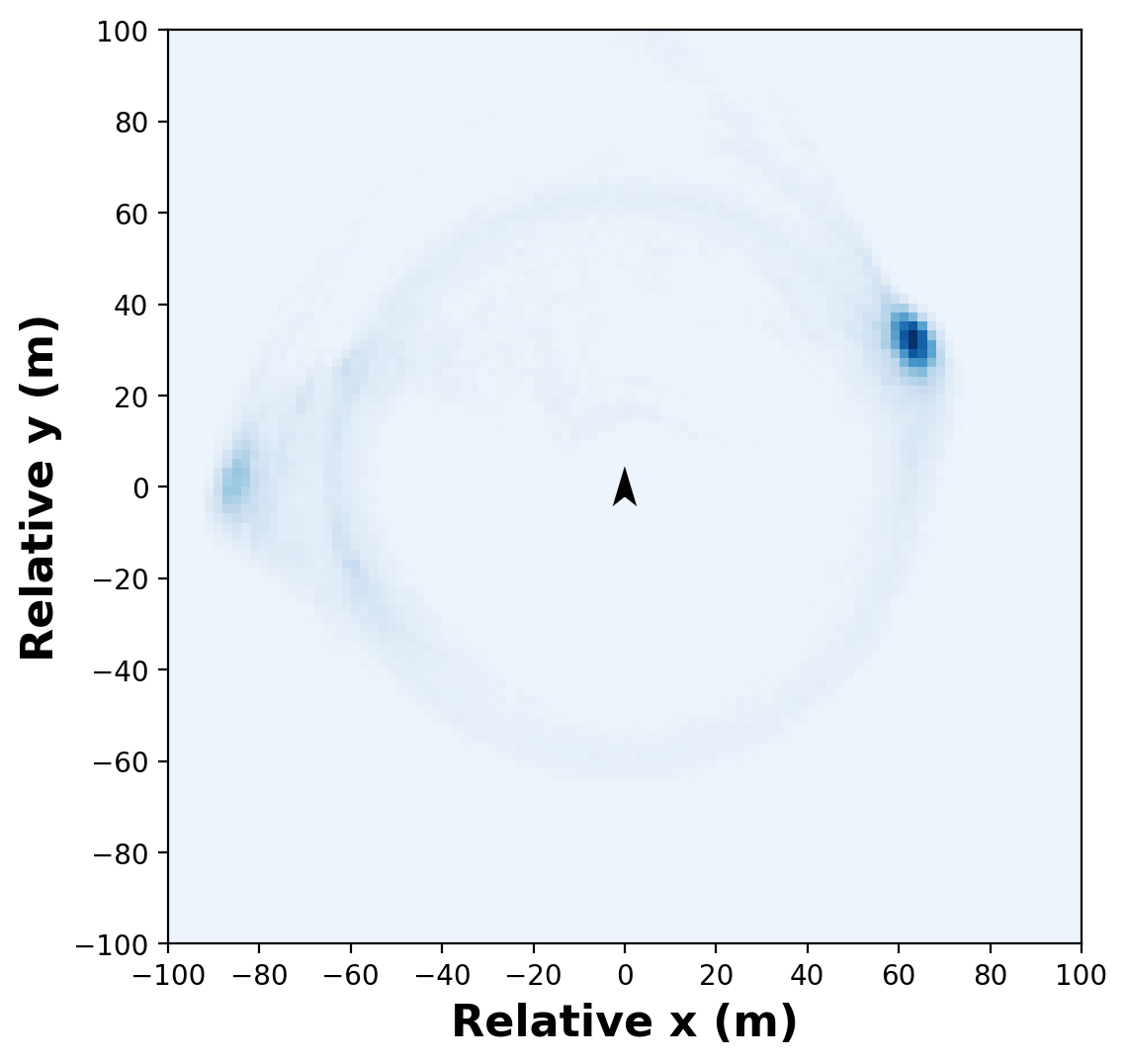}
        \label{fig:synthetic_visit_lpoi_ppo}
    \end{subfigure}

    \vspace{0.4cm}

    \begin{subfigure}[t]{0.24\textwidth}
        \centering
        \includegraphics[width=\linewidth]{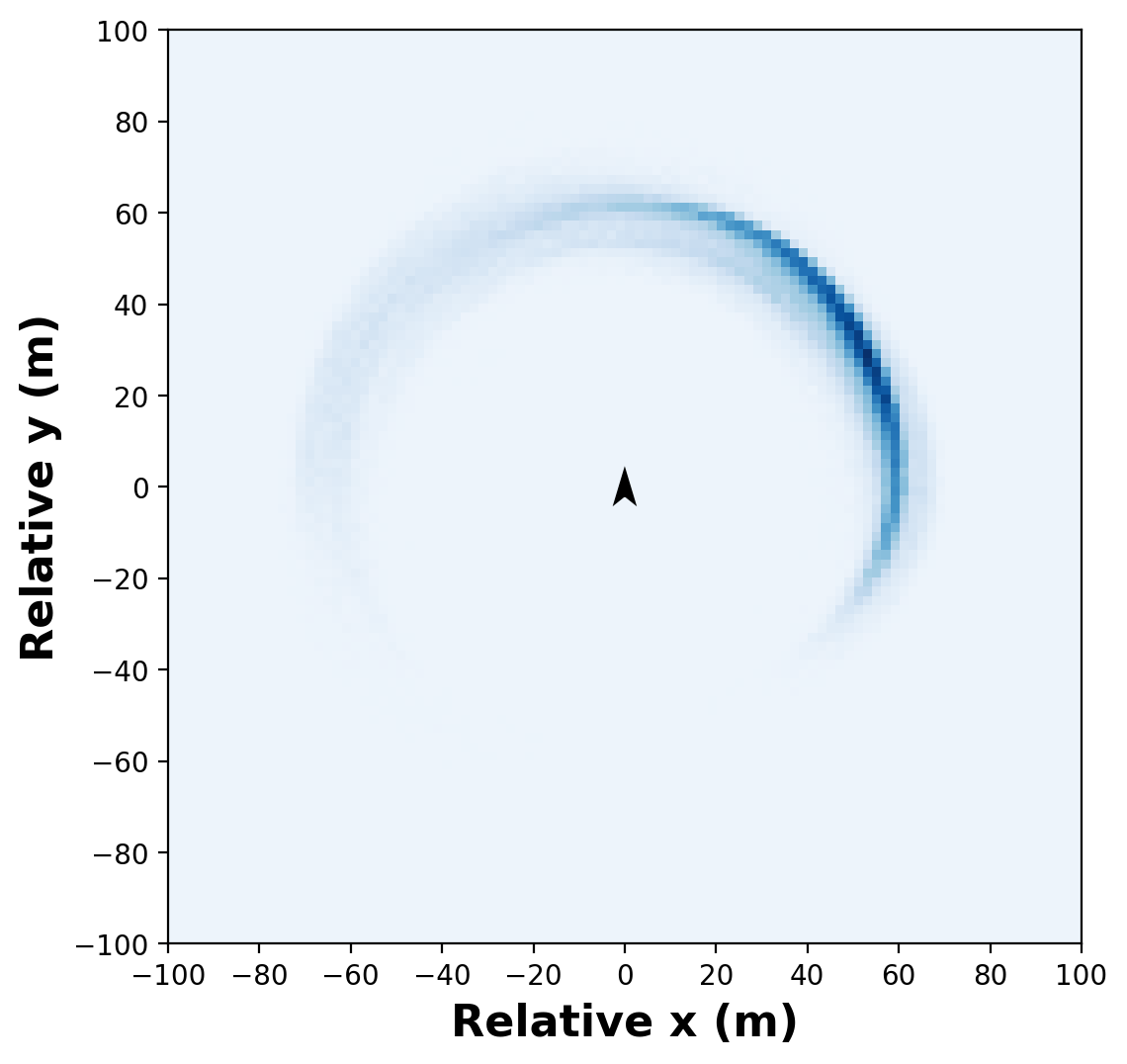}
        \label{fig:synthetic_visit_crw_sac}
    \end{subfigure}
    \hfill
    \begin{subfigure}[t]{0.24\textwidth}
        \centering
        \includegraphics[width=\linewidth]{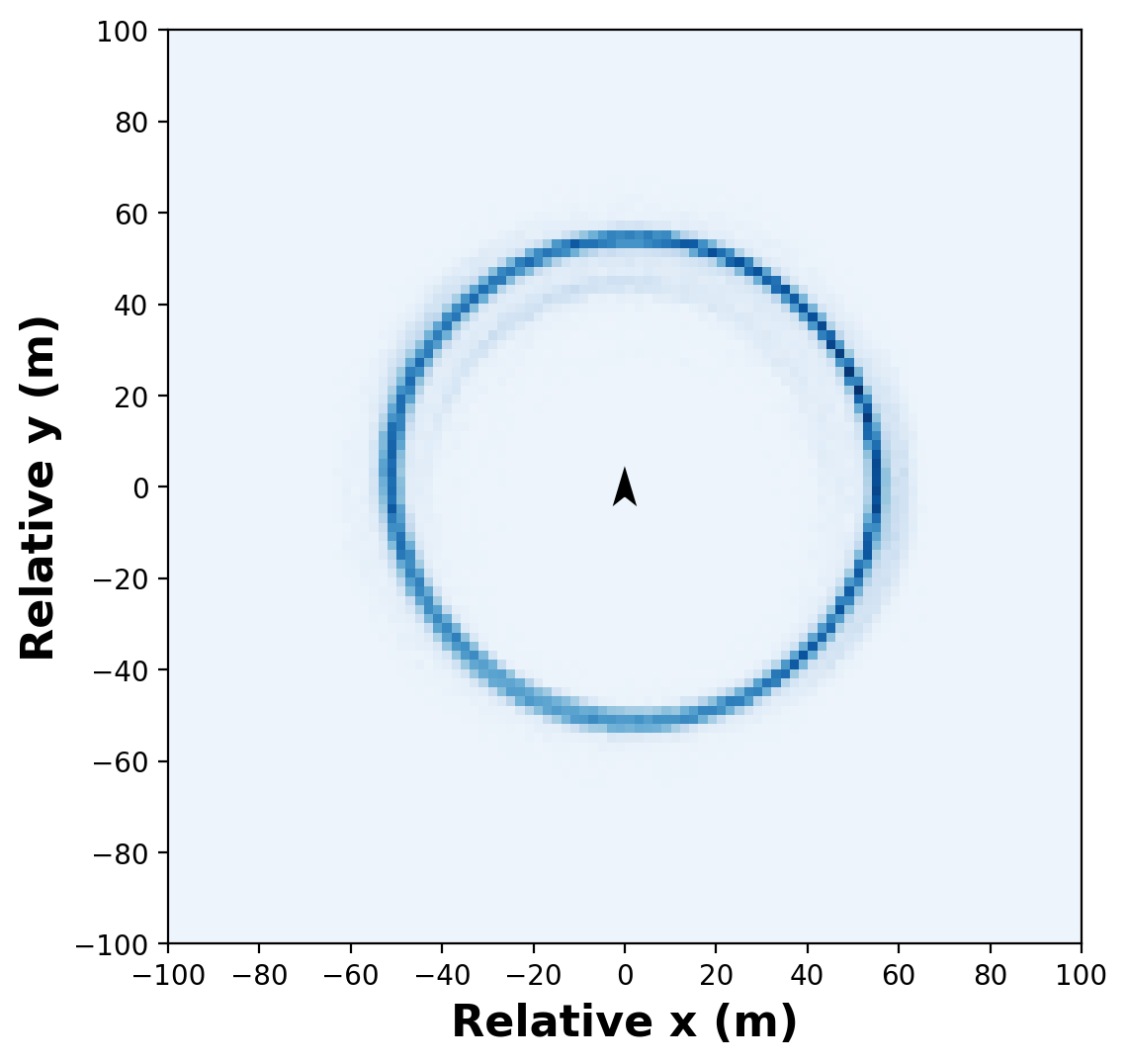}
        \label{fig:synthetic_visit_ee_sac}
    \end{subfigure}
    \hfill
    \begin{subfigure}[t]{0.24\textwidth}
        \centering
        \includegraphics[width=\linewidth]{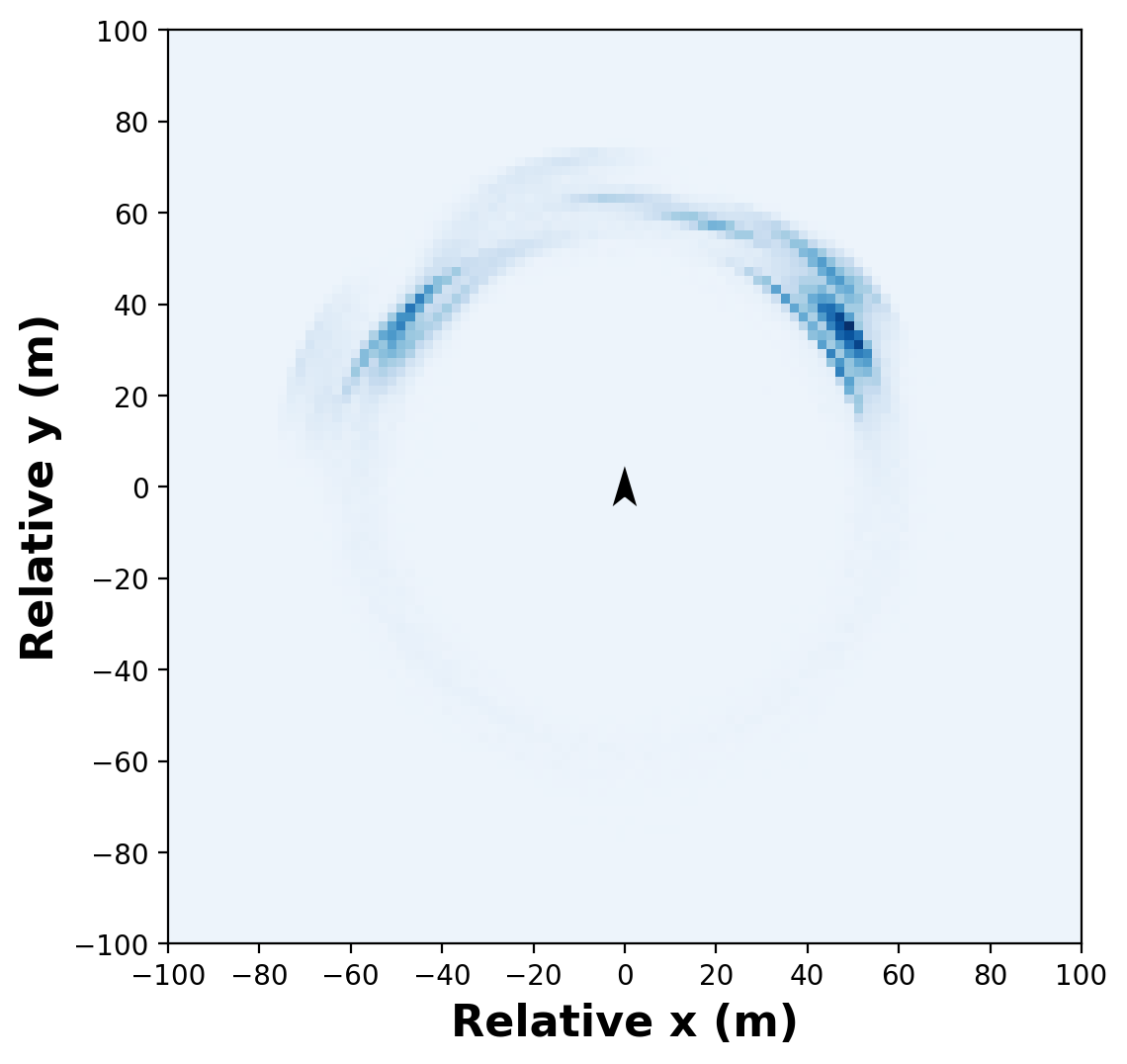}
        \label{fig:synthetic_visit_poi_sac}
    \end{subfigure}
    \hfill
    \begin{subfigure}[t]{0.24\textwidth}
        \centering
        \includegraphics[width=\linewidth]{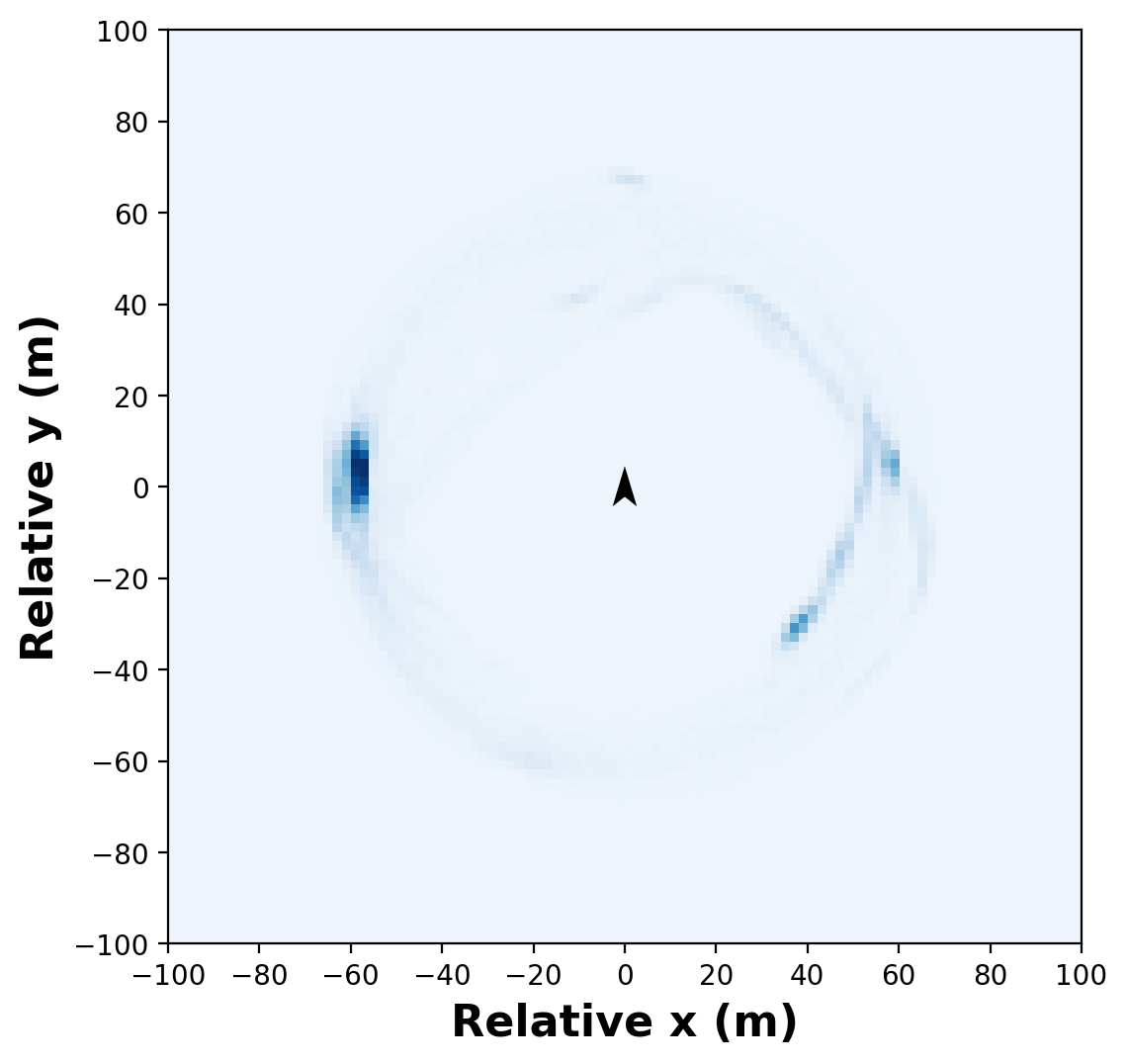}
        \label{fig:synthetic_visit_lpoi_sac}
    \end{subfigure}

    \caption{Top-down spatial distribution of drone positions relative to the target for policies trained on synthetic movement behaviors and evaluated on synthetic data. Policy is evaluated on synthetic data. Animal position and direction are signified by the centered marker. Columns correspond to the movement behavior used during training: CRW, EE, POI and LPOI. Rows correspond to the trained agent: DQN in the first row, PPO in the second row, and SAC in the third row. Each panel shows drone positions in animal-centered coordinates, with the horizontal relative position on the x-axis and the vertical relative position on the y-axis. The blue density indicates where the policy most frequently positioned the drone during evaluation, revealing whether the learned behavior tends to circle the target, maintain a preferred stand-off direction, or concentrate in a small number of approach regions. The black triangle shows the animal position and heading (along y). No wind environment, D2 drone sensing capabilities.}
    \label{fig:spatial_distributions_synthetic}
\end{figure*}

Figure~\ref{fig:spatial_distributions_synthetic} provides a top-down view of drone positions in animal-centered coordinates for policies trained and evaluated on synthetic movement. While the radial-altitude plots show distance and height, these top-down distributions reveal directional preferences, circling behavior, and whether policies concentrate around particular approach angles relative to animal heading.

\begin{figure*}[!ht]
    \centering

    \begin{subfigure}[t]{0.32\textwidth}
        \centering
        \includegraphics[width=\linewidth]{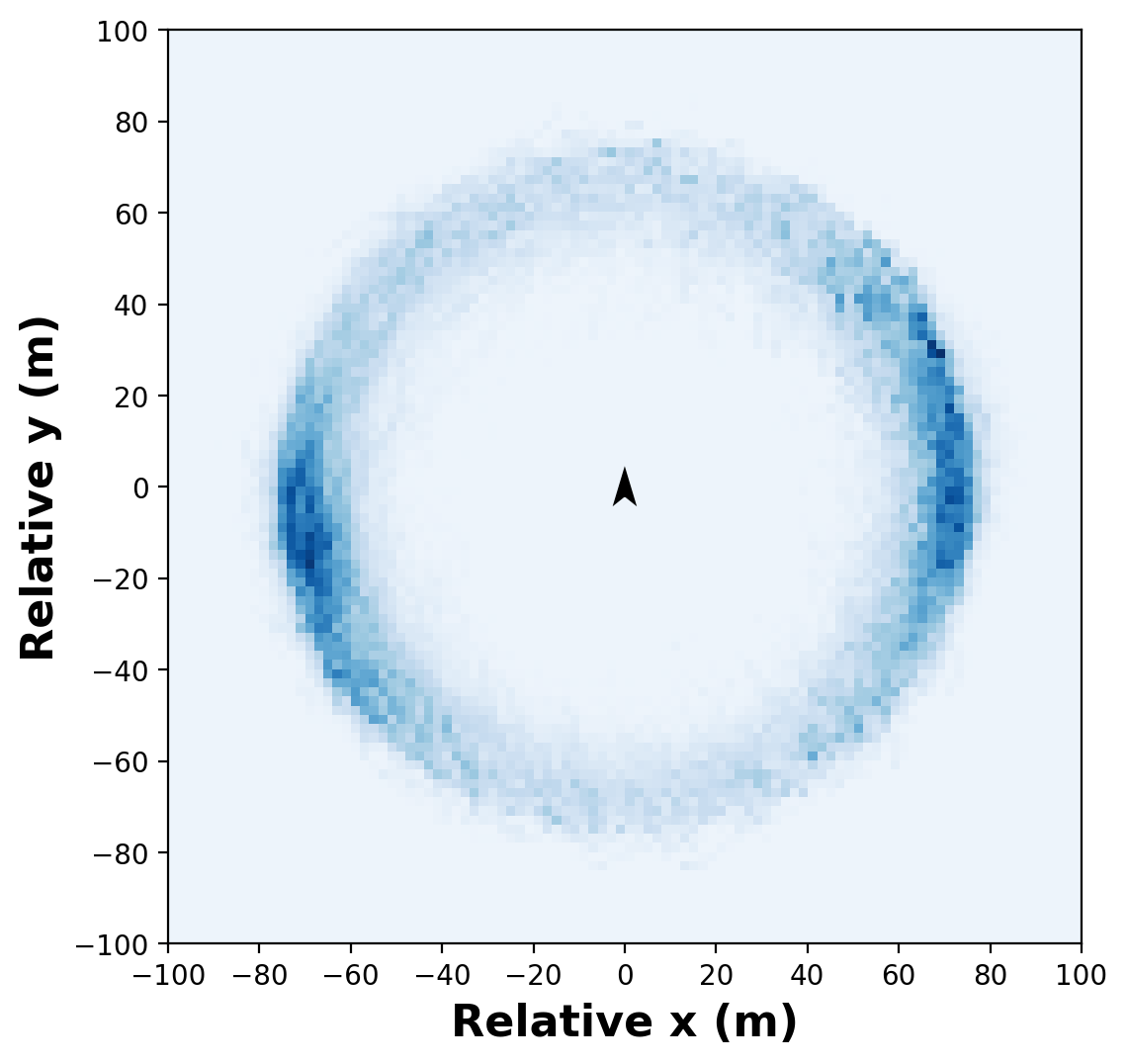}
        \caption{}
        \label{fig:prior_1}
    \end{subfigure}
    \hfill
    \begin{subfigure}[t]{0.32\textwidth}
        \centering
        \includegraphics[width=\linewidth]{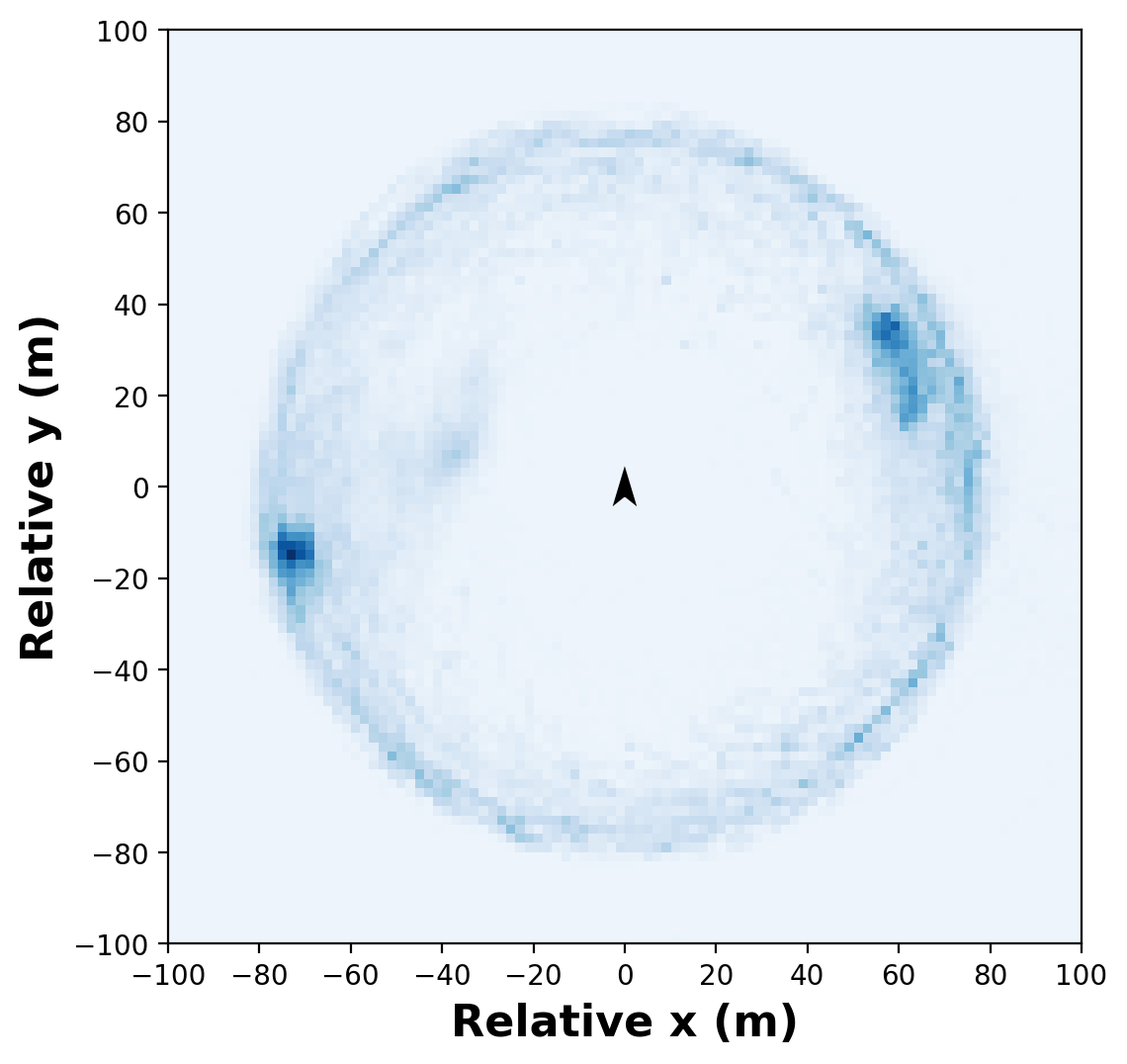}
        \caption{}
        \label{fig:prior_2}
    \end{subfigure}
    \hfill
    \begin{subfigure}[t]{0.32\textwidth}
        \centering
        \includegraphics[width=\linewidth]{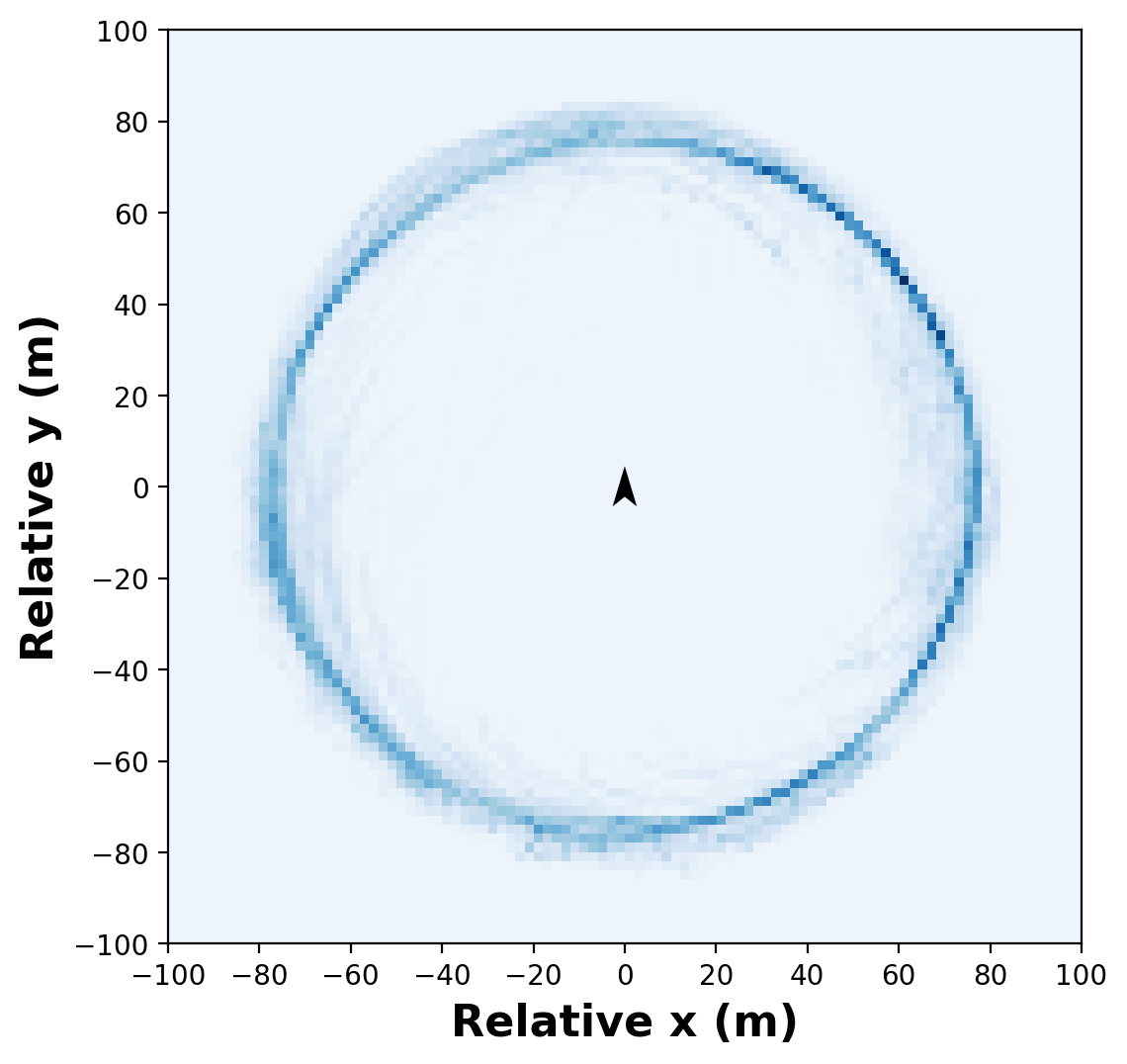}
        \caption{}
        \label{fig:prior_3}
    \end{subfigure}

    \vspace{0.4cm}

    \begin{subfigure}[t]{0.32\textwidth}
        \centering
        \includegraphics[width=\linewidth]{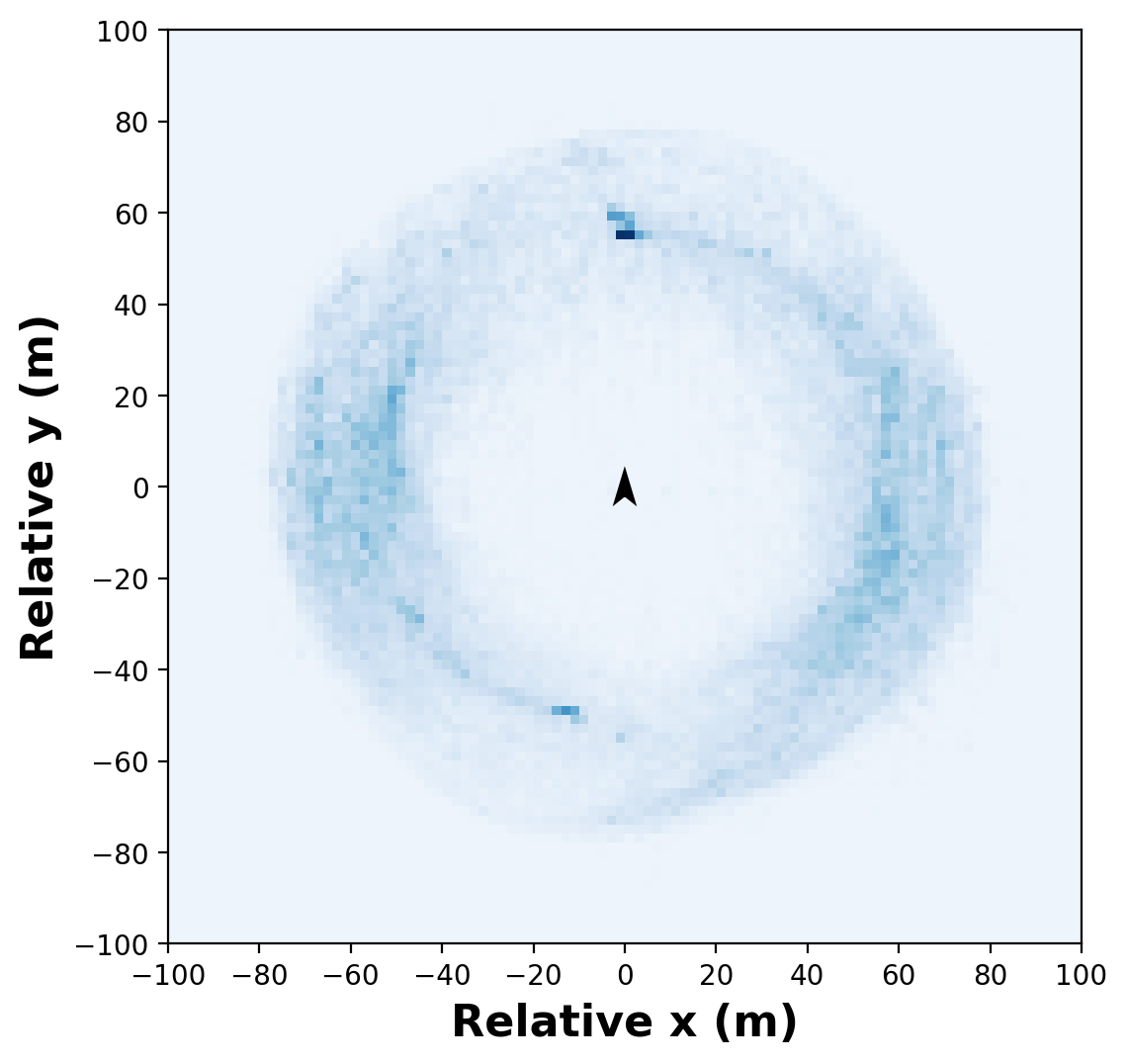}
        \caption{}
        \label{fig:prior_4}
    \end{subfigure}
    \hfill
    \begin{subfigure}[t]{0.32\textwidth}
        \centering
        \includegraphics[width=\linewidth]{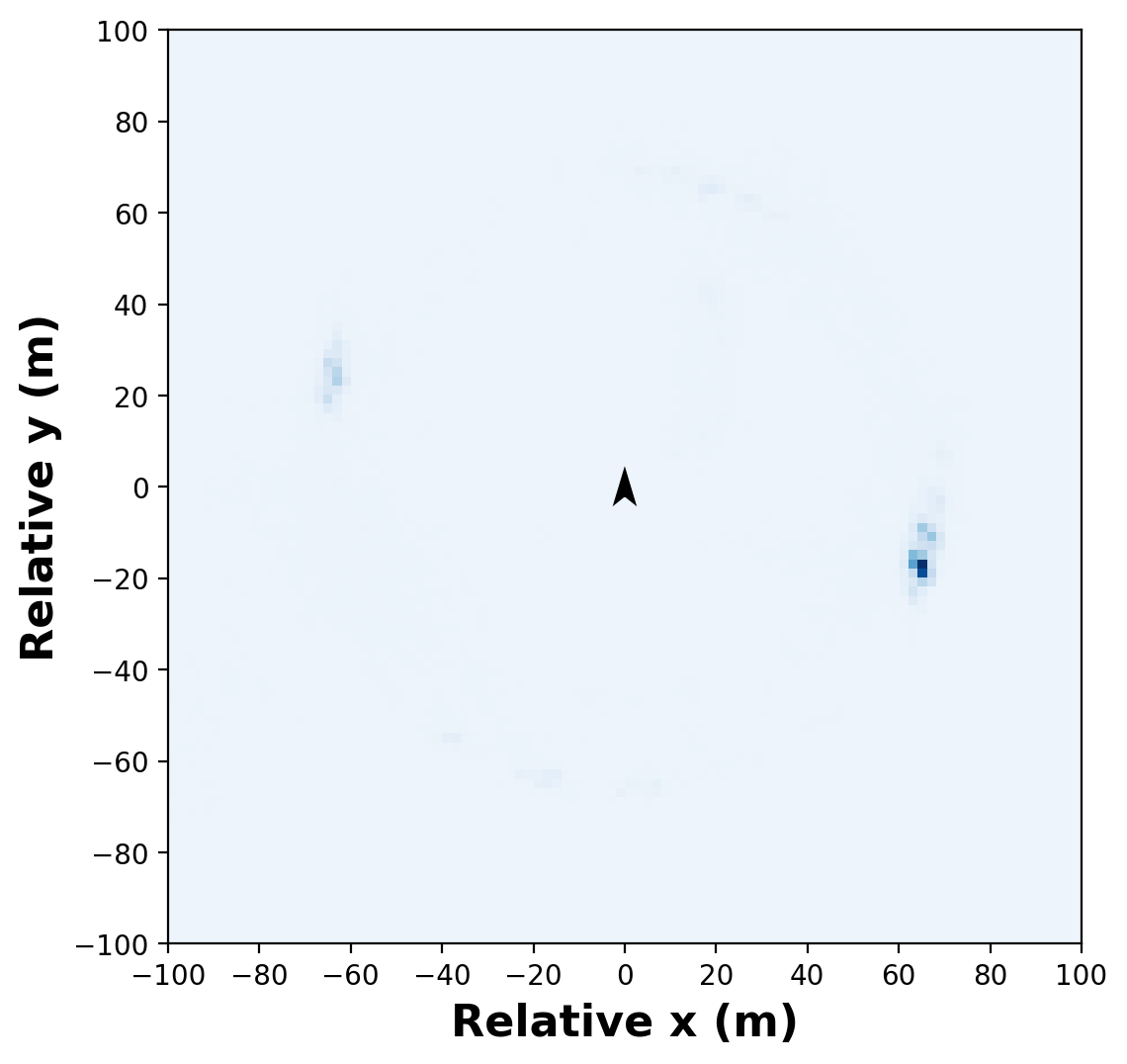}
        \caption{}
        \label{fig:prior_5}
    \end{subfigure}
    \hfill
    \begin{subfigure}[t]{0.32\textwidth}
        \centering
        \includegraphics[width=\linewidth]{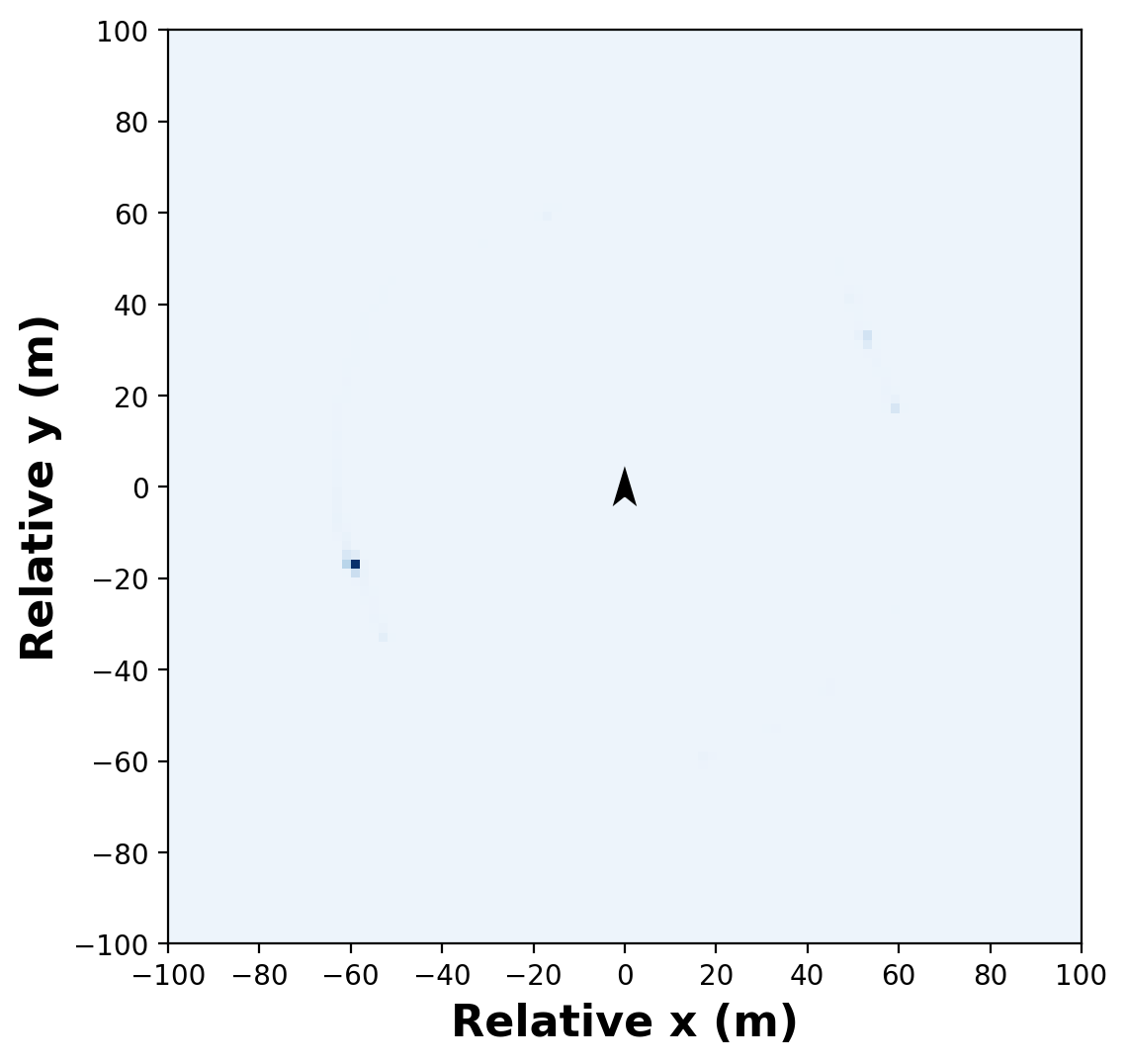}
        \caption{}
        \label{fig:prior_6}
    \end{subfigure}

    \vspace{0.4cm}

    \begin{subfigure}[t]{0.32\textwidth}
        \centering
        \includegraphics[width=\linewidth]{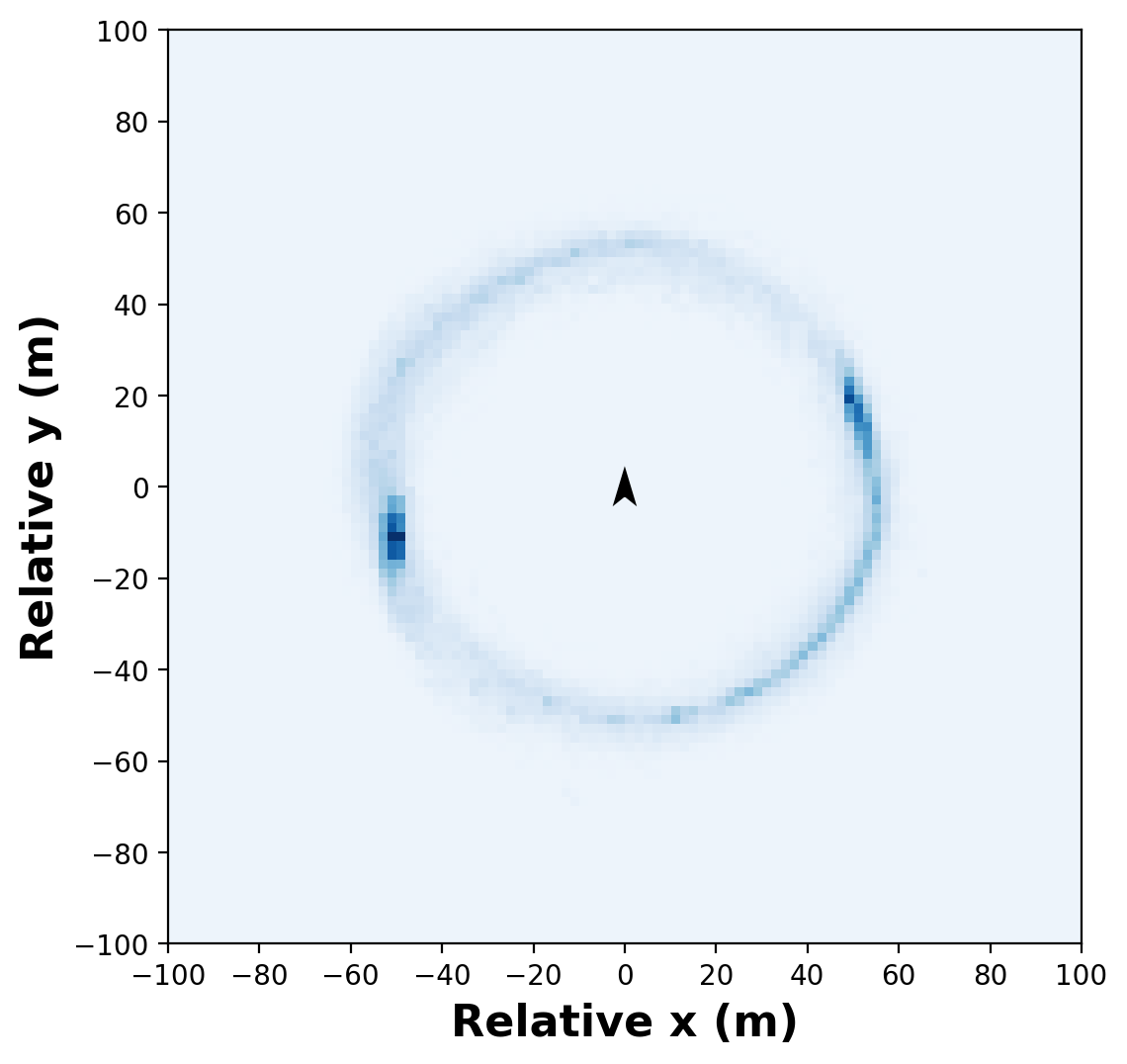}
        \caption{}
        \label{fig:prior_7}
    \end{subfigure}
    \hfill
    \begin{subfigure}[t]{0.32\textwidth}
        \centering
        \includegraphics[width=\linewidth]{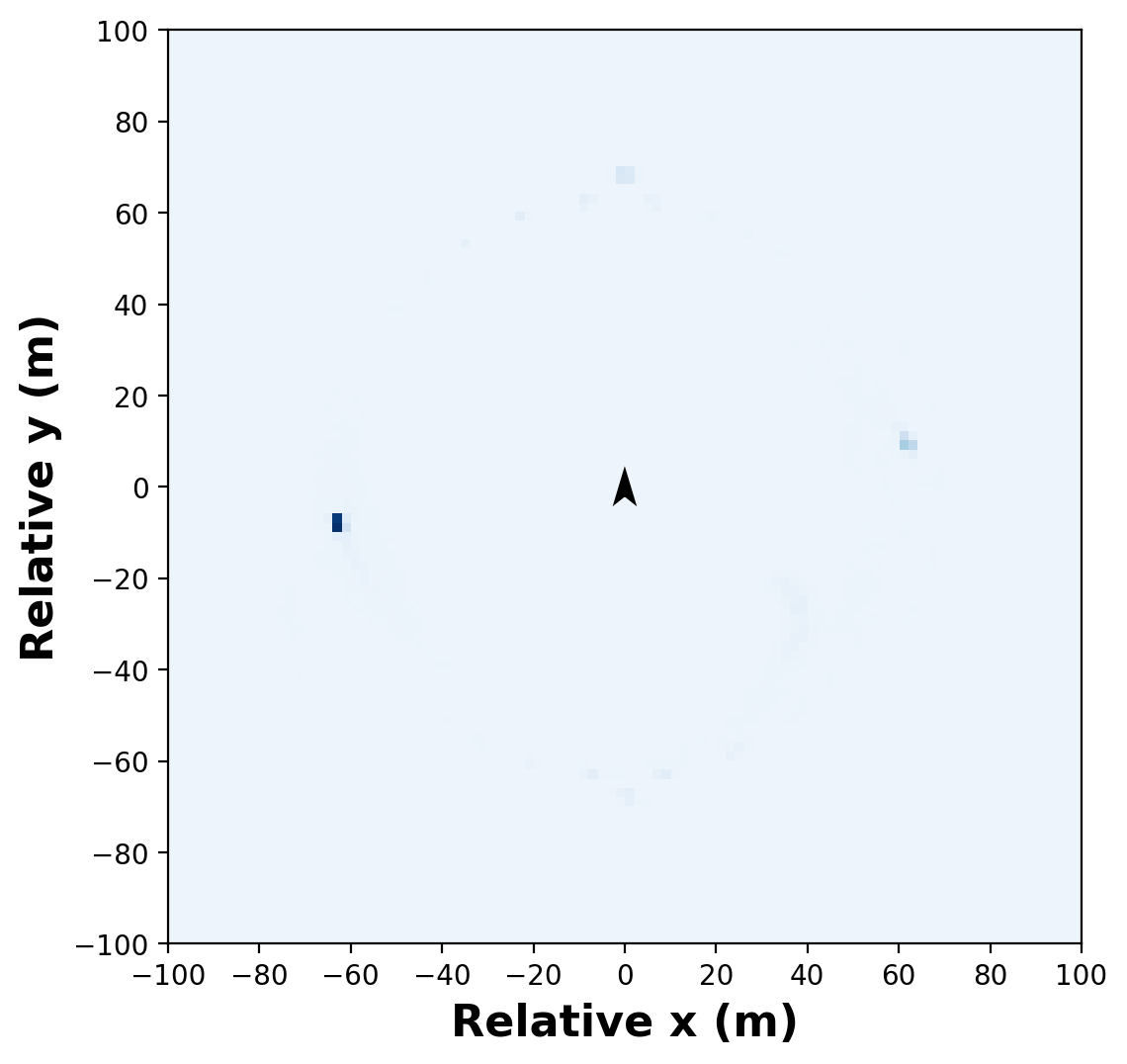}
        \caption{}
        \label{fig:prior_8}
    \end{subfigure}
    \hfill
    \begin{subfigure}[t]{0.32\textwidth}
        \centering
        \includegraphics[width=\linewidth]{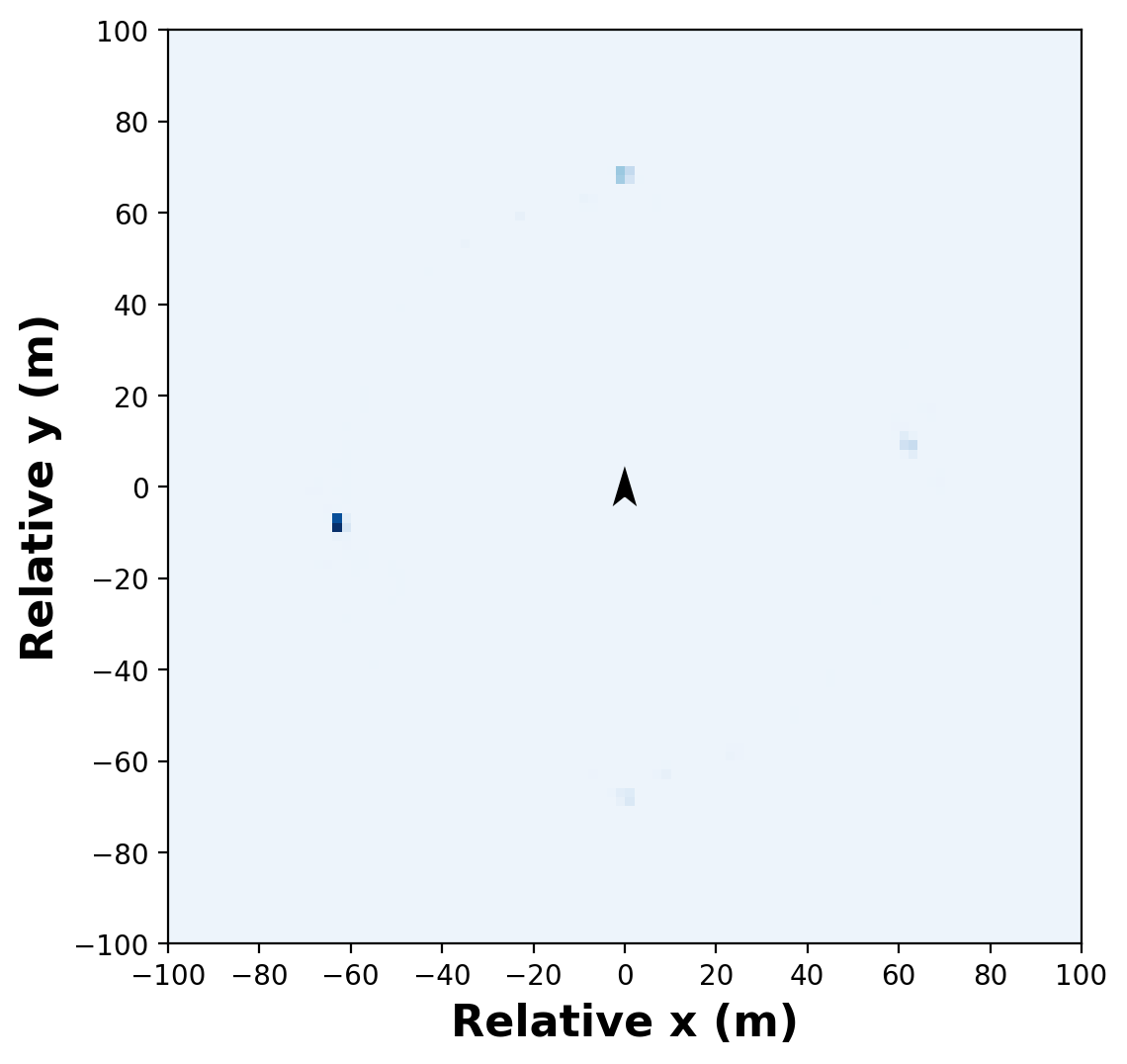}
        \caption{}
        \label{fig:prior_9}
    \end{subfigure}

    \caption{Top-down spatial distribution of drone positions relative to the animal for the best movement-prior policy selected for each model and animal. Policy is evaluated on real-sampled GPS animal movement data. Animal position and direction is signified by the centered marker. Rows correspond to the trained agent: DQN in the first row, PPO in the second row, and SAC in the third row. Columns correspond to the animal species: Jackals in the first column, Pigeons in the second column, and Spur-winged lapwings in the third column. Each panel shows the drone position in animal-centered coordinates, with relative horizontal position on the x-axis and relative vertical position on the y-axis. The blue density indicates where the policy most frequently positioned the drone during evaluation, showing whether the policy tended to circle the animal, remain at a preferred stand-off direction, or concentrate in a small number of approach regions. The black triangle shows the animal position and heading (along y). No wind environment, D2 drone sensing capabilities.}
    \label{fig:spatial_distributions_gps}
\end{figure*}

Finally, Figure~\ref{fig:spatial_distributions_gps} shows top-down animal-centered drone-position distributions during empirical GPS trajectories. These plots provide a qualitative view of how learned policies behave around real animal movement trajectories and whether their preferred monitoring positions differ across species and learning algorithms.

\end{document}